\documentclass{article} 
\usepackage[preprint]{colm2026_conference}

\usepackage{microtype}
\usepackage{hyperref}
\usepackage{url}
\usepackage{booktabs}
\usepackage{soul}
\usepackage{bbm}
\usepackage{wrapfig}
\usepackage{multirow}
\usepackage{makecell}
\usepackage{tcolorbox}
\usepackage{xcolor}
\usepackage{colortbl}
\usepackage{amsmath}
\usepackage{amssymb}
\usepackage{mathtools}
\usepackage{amsthm}
\usepackage{graphicx}
\usepackage{caption}
\usepackage{subcaption}
\usepackage{dashrule}
\usepackage{algorithm}
\usepackage{algcompatible}
\usepackage{lineno}

\definecolor{darkblue}{rgb}{0, 0, 0.5}
\hypersetup{colorlinks=true, citecolor=darkblue, linkcolor=darkblue, urlcolor=darkblue}
\sethlcolor{red!20}

\newcommand\blfootnote[1]{%
  \begingroup
  \renewcommand\thefootnote{}\footnote{#1}%
  \addtocounter{footnote}{-1}%
  \endgroup
}

\title{
Measuring Five-Nines Reliability: Sample-Efficient \\ LLM Evaluation in Saturated Benchmarks
}

\author{Eungyeup Kim \quad Chenchen Gu$^{*}$ \quad Vashisth Tiwari$^{*}$ \quad J. Zico Kolter \\
Carnegie Mellon University
}

\begin{document}

\ifcolmsubmission
\linenumbers
\fi

\maketitle

\begin{abstract}
While existing benchmarks demonstrate the near-perfect performance of large language models (LLMs) on various tasks, this apparent saturation often obscures the need for rigorous evaluation of their reliability.
In real-world deployment, however, achieving extremely high reliability (e.g., ``five-nines'' ($99.999\%$) vs. ``three-nines'' ($99.9\%$)) is fundamentally critical, as this gap results in an order-of-magnitude increase in failures, which is catastrophic in reliability-critical applications.
Still, estimating such a rare failure probability with tight confidence bounds requires prohibitively large LLM inference sizes, making standard Monte Carlo evaluation infeasible under limited compute budgets.
In this paper, we observe that LLM failures exhibit strong systematic patterns: across broad parameterized input spaces, a small subset of inputs disproportionately accounts for the majority of failures.
Leveraging this observation, we propose to learn a sampling distribution concentrated on failure-prone inputs via the cross-entropy method (CEM).
We evaluate our framework on three LLMs, Qwen2.5-Math-7B-Instruct, gpt-oss-20b-low, and Gemini 2.5 Flash Lite, across parameterized GSM8K templates and achieve up to $156.22\times$ reduction in required inferences compared to naive uniform sampling.
Our estimates reveal that models with indistinguishable accuracy on standard benchmarks can differ substantially in estimated failure rates, underscoring that reliability is a distinct and measurable axis of model quality.
Our simple yet practical framework enables the evaluation of extreme reliability in LLMs, a distinct and underexplored dimension of evaluation beyond existing benchmarks, for their growing use in reliability-sensitive applications.
\end{abstract}
\blfootnote{*co-second author,  \href{https://five-nines-reliability.notion.site/Measuring-Five-Nines-Reliability-Sample-Efficient-LLM-Evaluation-in-Saturated-Benchmarks-312b998d4f39802d88c0e9886db1b9cd}{Project page}}

\section{Introduction}
\label{sec:introduction}

\begin{figure}[ht]
    \begin{subfigure}[b]{0.32\linewidth}
        \caption*{\scriptsize Qwen2.5-Math-7B-Instruct,\\ Template 6}
        \vspace{-0.1cm}
    \includegraphics[width=\linewidth]{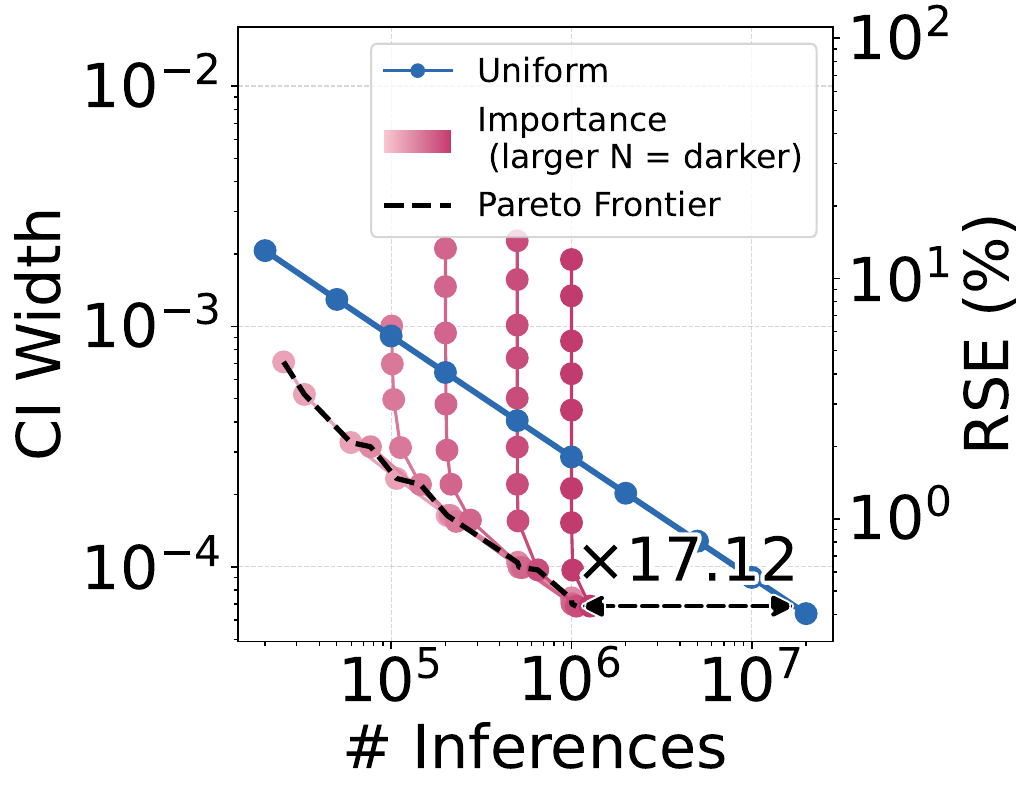}
    \end{subfigure}
    \hfill
    \begin{subfigure}[b]{0.32\linewidth}
        \caption*{\scriptsize gpt-oss-20b-low,\\ Template 8}
        \vspace{-0.1cm}
        \includegraphics[width=\linewidth]{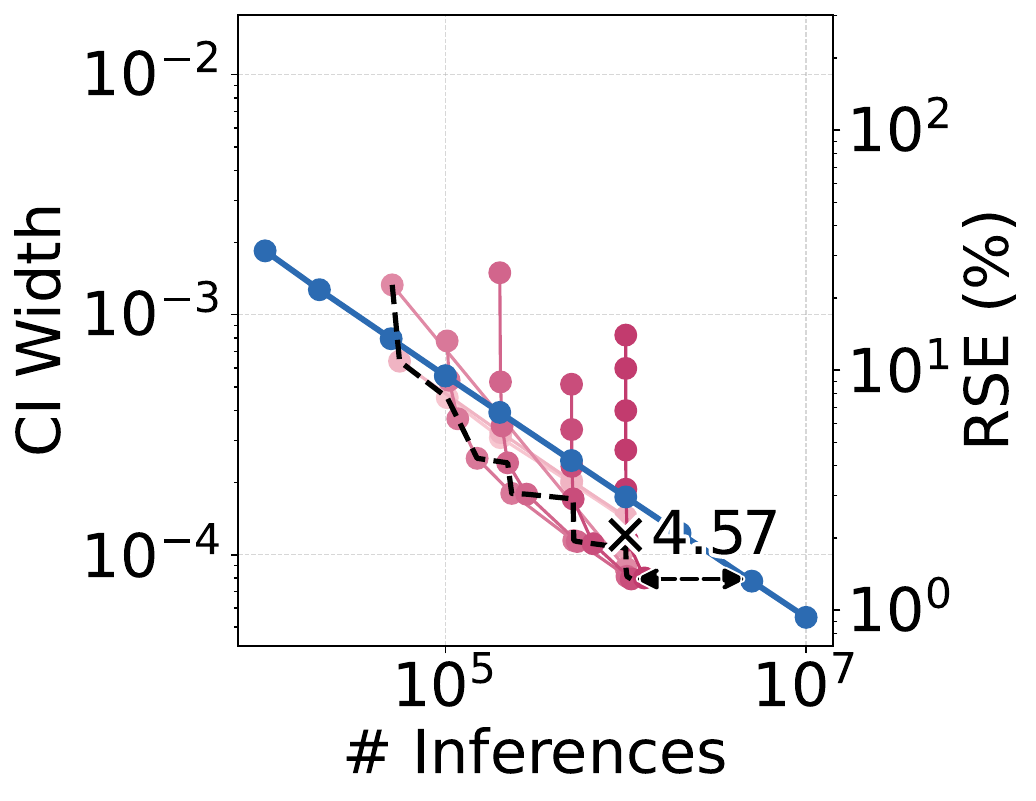}
    \end{subfigure}
    \hfill
    \begin{subfigure}[b]{0.32\linewidth}
        \caption*{\scriptsize Gemini 2.5 Flash Lite,\\ Template 6}
        \vspace{-0.1cm}
        \includegraphics[width=\linewidth]{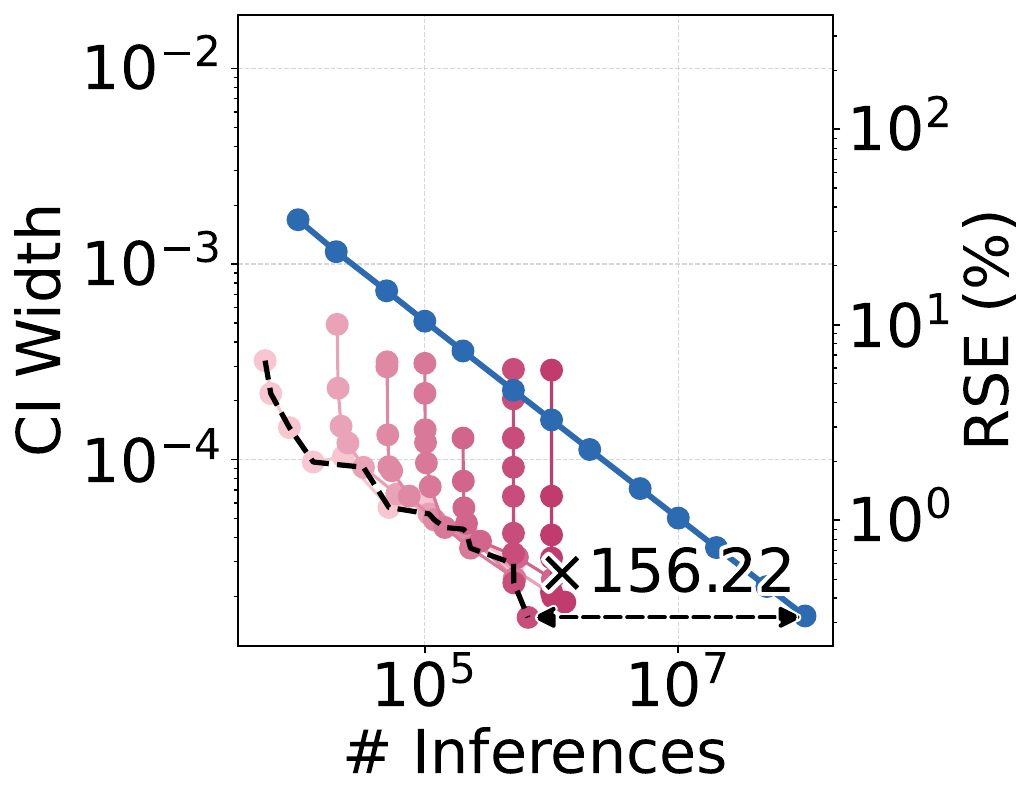}
    \end{subfigure}
    \caption{
    Each solid pink lines connects estimations with varying number of inferences (pink dots) under importance sampling with same number of overhead inferences for CEM ($N$), and darker pink means larger $N$. 
    Similarly, solid blue line for uniform sampling.
    Only estimates with coverage within $1\%$ tolerance of $99\%$ are shown.
    The Pareto-frontier (dotted line) is the lower-left envelope across all $(N, \text{\# evaluation})$ combinations, which achieves a tight CI width, i.e., relative standard error (RSE) of around $1\%$, at a much smaller inference budget compared to uniform sampling (e.g., $156.22\times$ gain for Gemini 2.5 Flash Lite on Template 6).
    }
    \vspace{-0.5cm}
    \label{fig:main_curve}
\end{figure}

Large language models (LLMs) show rapid progress not just in completing tasks with shorter duration~\citep{metr2026horizon}, but in giving saturating performance in numerous benchmarks~\citep{maslej2025artificialintelligenceindexreport}.
For example, in AIME 2025 (American Invitational Mathematics Examination), GPT-5.2 Thinking~\citep{openai2025gpt52} and Gemini-3-Pro~\citep{google_gemini3_2025} achieve a perfect score, while Claude Opus 4.6~\citep{anthropic2026opus4.6} reaches $99.8\%$.
While these near-perfect performances give the impression that models seem reliable at first glance, they may instead obscure how genuinely reliable these models are in real-world deployments where reliability becomes extremely important.
As an example, in safety-critical settings such as refusing harmful requests~\citep{openai2026gpt54benchmarks,anthropic2026opus4.6}, even a vanishingly small gap in accuracy, e.g., $99.999\%$ (``five-nines'') and $99.9\%$ (``three-nines'') in millions of queries, translates into an order-of-magnitude difference in failures, e.g., $1000$ vs. $10$ failures.
Thus, this dimension of reliability evaluation is increasingly important but remains under-explored in current evaluation pipelines of existing benchmarks~\citep{vendrow2024large}.

One major practical bottleneck is that, for such rare failures, achieving the tight confidence bounds requires a prohibitively large sample size via Monte Carlo sampling under a uniform distribution $P$.
For instance, to estimate a failure rate of $p = 10^{-5}$ (equally $99.999\%$ accuracy) with small standard error $\varepsilon$, the binomial estimator $\sqrt{p(1-p)/n}$ requires $n \approx p/\varepsilon^2 = 10^{-5}/(10^{-6})^2=10^7$ when $\varepsilon=10^{-6}$.
Combined with the expense of LLM inference (which grows further with long reasoning traces in recent models), exhaustive uniform sampling for evaluation becomes computationally infeasible.

In this paper, we observe that \textit{LLM failures are not uniformly distributed but exhibit systematic patterns}.
Specifically, in parameterized GSM8K~\citep{mirzadeh2025gsmsymbolic}, across a combinatorially large input space, only a small set of inputs consistently induces failures at disproportionately high rates.
For example, gpt-oss-20b-low~\citep{agarwal2025gpt} shows that approximately $82\%$ of its failures strongly co-occur with the input prompts with a specific parameter value (e.g., \texttt{fraction of blue ball} $=2/7$ for GSM template 7), while other values rarely cause failures (Figure~\ref{fig:param_hist} middle row, Figure~\ref{fig:example_fails} right panel).
Such concentration appears across different models, such as Qwen2.5-Math-7B-Instruct~\citep{yang2024qwen25math} and Gemini 2.5 Flash Lite~\citep{comanici2025gemini25pushingfrontier}, and other templates in GSM.
This phenomenon sharpens under self-consistency decoding with majority voting~\citep{wang2023selfconsistency}, which filters out random errors and reveals inputs that consistently trigger failures across multiple generations.

This systematic failure structure motivates our approach---rather than sampling uniformly, we concentrate \textit{sampling on failure-prone regions}.
We cast this as a rare-event estimation problem and adopt the CEM~\citep{rubinstein1999cross,de2005tutorial} to learn a new sampling distribution $Q$ concentrated on failure-prone regions.
From a small number of inferences, $Q$ iteratively converges to a distribution that assigns higher probability to failure-prone input regions, which enables importance sampling for inference-efficient error estimation.

Our evaluation testbed includes multiple parameterized GSM8K templates~\citep{cobbe2021gsm8k,mirzadeh2025gsmsymbolic}, and three small but capable LLMs: Qwen2.5-Math-7B-Instruct, gpt-oss-20b-low, and Gemini 2.5 Flash Lite.
Across models and templates, our method gains substantial inference efficiency, e.g., $156.22\times$ reduction in the required number of inferences for Gemini 2.5 Flash Lite on Template 6, for equivalent confidence bounds compared to uniform sampling (Figure~\ref{fig:main_curve} third plot).
Notably, these gains grow as LLM failures become more concentrated, i.e., via increasing majority-vote ensemble size $K$.
Specifically, efficiency improves from $3.68\times$ at $K=4$ to $16.41\times$ at $K=16$ for Qwen2.5-Math-7B-Instruct on Template 0 (Table~\ref{tab:gain_error}).
Our estimation results give $\sim1\%$ relative standard errors for models with nearly $0.1$\%--$0.01\%$ error rates, indicating that our method provides a practical evaluation pipeline for models' reliability on extremely saturated tasks.
Applying this pipeline across models, we find that models all exceeding $99.9\%$ accuracy still differ by up to $2.4\times$ in estimated failure rates (e.g., gpt-oss-20b-low at $0.0254\%$ vs.\ Qwen2.5-Math-7B-Instruct at $0.0614\%$ on Template 0, $K{=}16$), confirming that saturated benchmarks retain discriminative power at the reliability level (Figure~\ref{fig:eval}).

\section{Related Work}
\label{sec:related_work}

\paragraph{Evaluating reliability in saturated benchmarks.}
As LLMs push the frontier, they are evaluated on new, challenging benchmarks~\citep{jimenez2024swebench,merrill2026terminalbenchbenchmarkingagentshard,rein2024gpqa,yue2023mmmu,yue2025mmmu-pro,phan2025lastexam,foundation2026arcagi3newchallengefrontier}, leaving behind older, saturated ones~\citep{cobbe2021gsm8k,zellers2019hellaswag,wang-etal-2018-glue,sakaguchi2019winogrande} on which frontier models are now considered to perform nearly perfectly.
Yet near-perfect is not perfect: \citet{vendrow2024large} showed that frontier LLMs, such as o1-2024-12-17 (high)~\citep{openai2024o1}, retain rare failures, around $0.1\%$, on GSM8K~\citep{cobbe2021gsm8k} even after correcting noisy label errors.
Relatedly, GSM-Symbolic~\citep{mirzadeh2025gsmsymbolic} demonstrates that frontier models' near-perfect performance degrades on parameterized variants of the same problem templates.
We focus on rigorous estimation of failure rates for models on highly saturated benchmarks, targeting regimes as extreme as ``five-nines'' (99.999\%) accuracy, where the rarity of failures makes accurate estimation computationally costly compared to less saturated settings.

\paragraph{Efficient LLM evaluation.}
The autoregressive nature of LLMs makes inference expensive, and this cost grows as test-time compute scales~\citep{openai2024o1,guo2025deepseek}.
Several approaches reduce evaluation cost by selecting informative subsets of benchmark items~\citep{polo2024tinybenchmarks,vivek2024anchor,perlitz2024efficient,hofmann2025fluid}, leveraging historical information~\citep{wu2026efficientevaluationllmperformance}, or using difficulty-adaptive testing~\citep{truong2025reliable}.
These studies target general capability evaluation on non-saturated benchmarks, whereas our work specifically addresses the rare-failure regime of highly saturated benchmarks.

\paragraph{Rare-event estimation with the CEM.}
The CEM~\citep{rubinstein1999cross, rubinstein2004cross, de2005tutorial} is a method for estimating rare-event probabilities by iteratively learning a proposal distribution concentrated on the rare-event region, enabling importance sampling with substantially reduced variance.
A few works have adopted the CEM for rare-event estimation, including estimating failure probabilities of black-box physical systems in safety-critical applications~\citep{arief2021deepprob} and estimating rare predictions of linear predictors~\citep{bai2022randomforest}.
Our work is the first to apply the CEM to LLM evaluation, where the rare event is a majority-vote failure, the parameter space is a high-dimensional categorical distribution over problem instances, and each LLM inference is expensive.

\section{Problem and Experimental Setup}
\label{sec:setup}

\paragraph{Problem parameterization.}
To evaluate a model over a broad problem distribution, we construct a parameterized representation of task instances (e.g., mathematical reasoning prompts) as a mapping from the parameters $Z$ to natural language prompts.
Specifically, we define $Z = (Z_1, \dots, Z_d)$ as a vector of $d$ independent categorical variables, where each $Z_i$ takes values in a finite domain $\mathcal{Z}_i$. 
Following \citet{mirzadeh2025gsmsymbolic}, we parameterize all types of variables --- including nouns and numerical values.

To illustrate, consider a template where specific nouns and values are treated as parameters:

\begin{figure}[h]
\begin{minipage}{\linewidth}
\fbox{
\parbox{0.97\linewidth}{
\footnotesize
\textbf{Prompt (Template 0)}\\
\{$Z_{\text{Name}}$\} saw a \{$Z_{\text{x}}$\}-foot \{$Z_{\text{Fish}}$\} with \{$Z_{\text{y}}$\} remoras, each \{$Z_{\text{z}}$\}-inches long, attached to it. What percentage of the 
\{$Z_{\text{Fish}}$\}'s body length is the combined length of the remoras?
}
}
\end{minipage}
\end{figure}

Each placeholder corresponds to a variable $Z_i$ sampled from its domain $\mathcal{Z}_i$ (e.g., $Z_{\text{Name}} \in \mathcal{Z}_{\text{Name}}$, $Z_x \in \mathcal{Z}_x$).
This formulation provides a structured, factorized representation of the problem input space.
Let $\mathcal{Z} = \prod_{i=1}^d \mathcal{Z}_i$ denote the full parameter space.
Details for templates are given in Appendix~\ref{appendix:templates}.

\paragraph{Sampling distribution and failure.}
Given the parameter space $\mathcal{Z}$, we define a uniform sampling distribution:
\begin{equation}
P(Z)=\frac{1}{\lvert \mathcal{Z}\rvert}.
\label{eq:targetP}
\end{equation}

Our goal is to estimate the failure probability under $P$,
\begin{align}
    \mu \coloneqq \mathbb{E}_{Z\sim P}\left[\mathbb{P}(f(Z) = 1)\right],
    \label{eq:target_mu}
\end{align}
where for each input $Z$, $f(Z) \in \{0,1\}$ is a Bernoulli random variable indicating whether majority voting over $K$ independent stochastic generations yields an incorrect prediction.

Modern LLMs often produce non-deterministic outputs (e.g., under non-zero temperature), especially for reasoning tasks.
To obtain more consistent predictions, we adopt self-consistency decoding~\citep{wang2023selfconsistency}: for each input $Z$, we perform $K$ independent inference runs and aggregate the outputs via majority voting.
This reduces output variance by aggregating over diverse reasoning paths, approximating the mode of the model's output distribution.

Formally, we define the failure indicator as
\begin{equation}
f(Z) \coloneqq \mathbbm{1}\{\mathrm{MajorityVote}\Big(\hat{y}_1(Z),\ldots,\hat{y}_K(Z)\Big)\neq y\},
\label{eq:majvote}
\end{equation}
where $\hat{y}_k(Z)$ denotes the model output from the $k$-th independent inference on input $Z$, and $y$ is the ground-truth label.
Ties are counted as failures.

\paragraph{Datasets and models.}
Building on GSM-Symbolic~\citep{mirzadeh2025gsmsymbolic}, we generate approximately $100$K instances per template by varying parameter ranges, yielding nine templates in total.
Details of all templates used throughout our paper can be found in Appendix~\ref{appendix:templates}.
We notice that for some template-model pairs, failures are extremely rare (i.e., the template is too easy for some model) which is potentially non-zero but undetectable within our feasible evaluation budget.
Therefore, we only report results for pairs with a sufficient number of observed failures to support reliable estimation.
Details of such excluded pairs are provided in Section~\ref{appendix:too_rare}.

Due to compute budget limits, we focus on relatively efficient yet capable LLMs, including Qwen2.5-Math-7B-Instruct~\citep{yang2024qwen25math}, gpt-oss-20b-low~\citep{agarwal2025gpt} with low reasoning efforts, and Gemini 2.5 Flash Lite~\citep{comanici2025gemini25pushingfrontier}.
All inferences follow the models' recommended decoding configurations (e.g., temperature, top-$p$, and max number of tokens), and details such as system prompts and parsing final answers are provided in Appendix~\ref{appendix:implementation_details}.

\paragraph{Bootstrapping Inferences.}
Ideally, evaluating an LLM requires fresh inferences for each generation, since LLM outputs are inherently stochastic (i.e., non-zero temperature), particularly during reasoning.
However, this becomes computationally infeasible for evaluating with large sample sizes for rare failures, across multiple models, templates, and random seeds (for coverage).
Therefore, we employ \textit{bootstrap resampling}~\citep{efron1979Bootstrap} to approximate the stochastic variation of LLM outputs without fresh inference for each input prompt.
Specifically, we pre-compute and cache a pool of independent generations for each input, $16$ generations for Qwen2.5-Math-7B-Instruct, $40$ for gpt-oss-20b-low, and $32$ for Gemini 2.5 Flash Lite, which is deemed large enough to approximate the true sampling.
We randomly sample $K$ outputs with replacement from this cached pool to simulate independent generations for majority voting.
This resampling approximates the distribution of majority-vote outcomes given fresh inferences without additional inference costs. 

\section{LLM Failures Are Systematic, Not Random}
\label{sec:observation}

\begin{figure}[t]
\centering
\begin{minipage}[t]{0.52\linewidth}
    \vspace{0pt}
    \centering

    \begin{subfigure}[t]{\linewidth}
        \centering
        \caption*{\scriptsize Qwen2.5-Math-7B-Instruct, Template 5, $K=16$}
        \vspace{-0.2cm}
        \includegraphics[width=\linewidth]{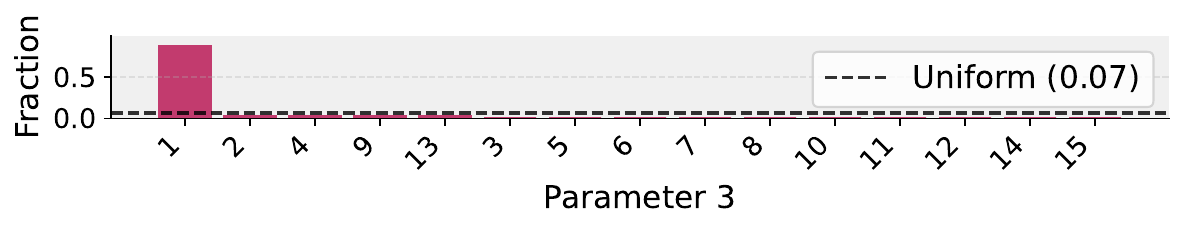}
    \end{subfigure}

    \begin{subfigure}[t]{\linewidth}
        \centering
        \caption*{\scriptsize gpt-oss-20b-low, Template 7, $K=24$}
        \vspace{-0.2cm}
        \includegraphics[width=\linewidth]{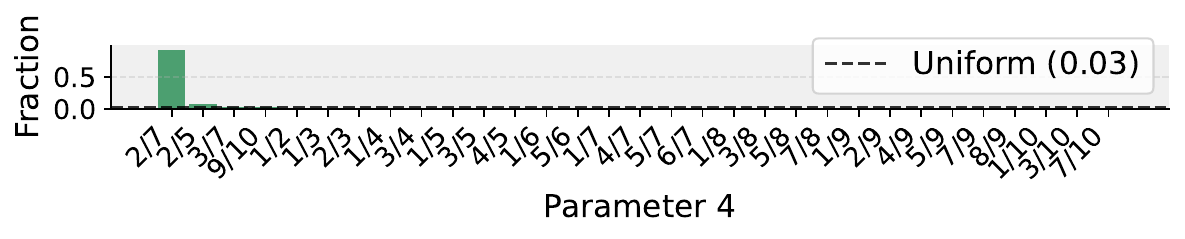}
    \end{subfigure}

    \begin{subfigure}[t]{\linewidth}
        \centering
        \caption*{\scriptsize Gemini 2.5 Flash Lite, Template 6, $K=16$}
        \vspace{-0.2cm}
        \includegraphics[width=\linewidth]{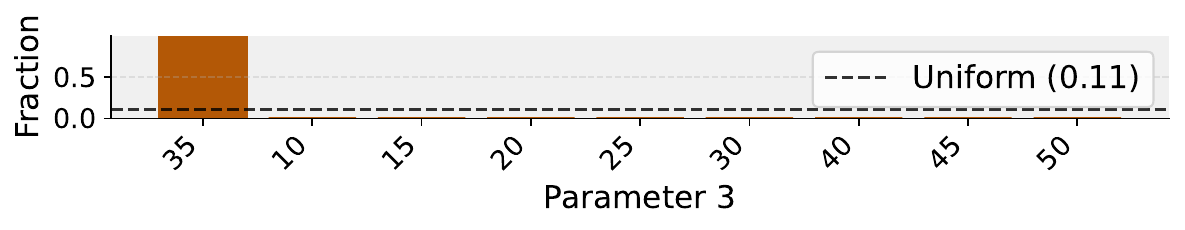}
    \end{subfigure}
    \vspace{-0.1cm}
    \caption{Distribution of failures across parameter values for model–template pairs. Each bar represents the proportion of failures for each parameter value, and the dotted line shows the expected uniform error rate.}
    \label{fig:param_hist}
\end{minipage}
\hfill
\begin{minipage}[t]{0.45\linewidth}
    \vspace{0pt}
    \centering
    \scalebox{0.8}{
    \begin{tabular}{c c ccc}
    \toprule
    \multirow{2}{*}{Model} & \multirow{2}{*}{ID} & \multicolumn{3}{c}{Distance}\\
    \cmidrule(lr){3-5}
    && Top1 & Top2 & Top3\\
    \midrule
    \multirow{8}{*}{\makecell{Qwen2.5-Math\\-7B-Instruct\\($K=16$)}}
    & 0 & 0.900 & 0.750 & 0.509\\
    & 1 & 0.914 & 0.833 & 0.750\\
    & 2 & 0.538 & 0.249 & 0.019\\
    & 3 & 0.472 & 0.362 & 0.310\\
    & 4 & 0.339 & 0.264 & 0.185\\
    & 5 & 0.940 & 0.867 & 0.800\\
    & 6 & 0.879 & 0.752 & 0.247\\
    & 7 & 0.798 & 0.748 & 0.557\\
    \midrule
    \multirow{5}{*}{\makecell{gpt-oss\\-20b-low\\($K=24$)}}
    & 0 & 0.933 & 0.850 & 0.729\\
    & 2 & 0.699 & 0.541 & 0.052\\
    & 4 & 0.631 & 0.508 & 0.385\\
    & 7 & 0.895 & 0.633 & 0.537\\
    & 8 & 0.760 & 0.599 & 0.487\\
    \midrule
    \multirow{4}{*}{\makecell{Gemini 2.5\\Flash Lite\\($K=16$)}}
    & 0 & 0.867 & 0.850 & 0.267\\
    & 2 & 0.907 & 0.750 & 0.080\\
    & 4 & 0.208 & 0.133 & 0.102\\
    & 6 & 0.983 & 0.889 & 0.189\\
    \bottomrule
    \end{tabular}}
    \vspace{-0.1cm}
    \captionof{table}{Total variation distance between failure distributions and uniform baseline. Parameters with the three largest distances are reported for each template.}
    \label{tab:tv_table}
\end{minipage}

\begin{minipage}[t]{0.55\linewidth}
\fbox{
\parbox{0.96\linewidth}{
\footnotesize
\textbf{Prompt (Template 5)}\\
\{Katy\} makes \{coffee\} using teaspoons of sugar and cups of water in the ratio of \{15\}:
\{\hl{1}\}. 
If she used a total of \{736\} teaspoons of sugar and cups of water, calculate the number of teaspoonfuls of sugar she used.\\
\hdashrule{0.96\linewidth}{0.5pt}{3pt 2pt}
\textbf{LLM Output (Qwen2.5-Math-7B-Instruct)}\\
To determine the number of teaspoonfuls of sugar Katy used, we start by\\
$\dots$\\
\hl{Since 1 cup of water is equivalent to 16 teaspoons (assuming standard US measurements}\\
$\dots$\\
Therefore, the number of teaspoonfuls of sugar Katy used is:
\boxed{360}
\\
Answer: \textbf{690}
}
}
\end{minipage}%
\hfill
\begin{minipage}[t]{0.43\linewidth}
\fbox{
\parbox{0.94\linewidth}{
\footnotesize
\textbf{Prompt (Template 7)}\\
\{Jamie\} is juggling at a \{circus\}. \{Jamie\} can juggle \{168\} balls. \{1/2\} of the balls are \{golf\} balls, and \{\hl{2/7}\} of the \{golf\} balls are \{blue\}. How many \{blue\} \{golf\} balls are there?\\
\hdashrule{0.94\linewidth}{0.5pt}{3pt 2pt}

\textbf{LLM Output (gpt-oss-20b-low)}\\
analysisCompute: total balls 168. Half are golf balls: 168/2=84 golf balls. \hl{Of those, 2/7 are blue: 84*(2/7)=12}.assistantfinalJamie has \\(\boxed{12}) blue golf balls.\\
Answer: \textbf{24}
}
}
\end{minipage}
\vspace{-0.1cm}
\caption{Example pairs of prompt-generations that illustrate repetitive failures by LLMs. Failure-inducing parameter values and output traces are highlighted in \hl{red}.
}
\label{fig:example_fails}
\end{figure}

\paragraph{Failures concentrate on a small subset of parameters.}
We observe that \textit{LLM failures are not random, but exhibit systematic structure}. Within the high-dimensional parameter space $\mathcal{Z}$, a small subset of parameter values disproportionately accounts for a large fraction of failures.
Across models and templates, we observe striking concentration of failures on individual parameter values (Figure~\ref{fig:param_hist}).
We first iterate over all possible inputs for testing models, and aggregate all failures observed.
We observe that a single parameter value accounts for the majority of errors: parameter \texttt{term of water cups} $=1$ in Template 5 for Qwen2.5-Math-7B-Instruct ($86\%$ of failures), parameter \texttt{fraction of blue ball} $=2/7$ in Template 7 for gpt-oss-20b-low~\citep{agarwal2025gpt} ($82\%$), and parameter \texttt{area of fruits} $=35$ in Template 6 for Gemini 2.5 Flash Lite~\citep{comanici2025gemini25pushingfrontier} ($81\%$).
Bar plots for all parameters are given in Appendix Figure~\ref{fig:full_param_hist}.

To quantify this concentration, we compute the total variation (TV) distance between the normalized histogram of the failure counts over each parameter dimensions ($Z_i$ such as $Z_\text{Name}$) and the uniform distribution (Table~\ref{tab:tv_table}).
The uniform distribution represents a baseline where the failures occur evenly across parameters. 
Thus, larger distances indicate more pronounced failure concentration, while a value of zero indicates that failures are spread evenly.
For each setup, we report the top-3 parameter dimensions with the largest divergence.
Across models and templates, we observe consistently large deviations from the uniform distribution, i.e., most Top-1 distances reach $0.9$.
Furthermore, as we increase $K$, we observe that the TV distance increases (Table~\ref{appendix:tv_table}), which implies that with larger $K$, the majority of failures becomes more pronounced, with fewer inconsistent failures.
Specifically, larger $K$ filters out random errors---failures caused by incidental factors in individual generations---while amplifying systematic failure patterns that persist across multiple samples.
These results suggest that LLM failures are not spread randomly across the parameter space, but are instead driven by a small subset of failure-inducing inputs.

\paragraph{Qualitative analysis of repetitive failures.}
To understand how the failures concentrate in certain regions of the parameter space, we conduct a qualitative analysis of LLM-generated outputs that exhibit repetitive failures, as in Figure~\ref{fig:example_fails}.
We mark in red the input parameter values that frequently induce failures as well as the erroneous reasoning trace.
For instance, Qwen2.5-Math-7B-Instruct consistently applies an unjustified unit conversion heuristic when the parameter value \texttt{term for cup of water} $=1$ is given in Template 5. 
gpt-oss-20b-low fails at simple multiplication when $2/7$ is given as \texttt{fraction of blue ball} in Template 7.
Similarly, Gemini 2.5 Flash Lite fails in $35\times 43$ in Template 6 (Figure~\ref{fig:appendix_example_fails_gemini6}).
These models exhibit such errors repeatedly across multiple generations, as illustrated in Figures~\ref{fig:appendix_example_fails_qwen5},\ref{fig:appendix_example_fails_low7}, and~\ref{fig:appendix_example_fails_gemini6}.
We provide additional qualitative analysis for other template-model pairs in Section~\ref{appendix:example_fails}, which show various error types including wrong arithmetic operations, unnecessary approximation, problem misinterpretation, and others.
These examples reveal that LLM errors in GSM problems stem from consistent, input-dependent reasoning behaviors, providing insight into the conditions under which models are likely to produce incorrect reasoning trajectories. 
Despite there being no a priori reason to expect systematic failure patterns to persist across template-model pairs, we find compelling empirical evidence that such patterns arise consistently across most pairs we examine.
\section{Learning Failure-Prone Sampling Distributions}
\label{sec:method}

Our key finding of the concentration of failures in Section~\ref{sec:observation} naturally leads to a failure-prone sampling strategy---rather than drawing samples from the full parameter space under $P$, we concentrate sampling on failure-prone regions.
To this end, we leverage the CEM to iteratively refine a proposal distribution $Q$ toward regions of the parameter space where failures are more likely, which we can use to estimate true error probability via importance sampling.

\begin{algorithm}[ht]
\caption{Failure Probability Estimation via CEM}
\begin{algorithmic}[1]
\STATE \textbf{Input:} Failure indicator $f$, original distribution $P$, sample size $N$, smoothing coefficient $\zeta$, iterations $T$, confidence level $\alpha$, defensive sampling coefficient $\lambda$
\STATE \textbf{Output:} Failure probability estimate $\hat{\mu}$, confidence interval $[\hat{\mu}_{\text{low}}, \hat{\mu}_{\text{high}}]$
\STATE \textbf{// CEM for updating $Q$}
\STATE Initialize $Q_0(Z_i)$ as uniform over $\mathcal{Z}_i$ for all $i=1,\ldots,d$
\FOR{$t = 1$ to $T$}
    \STATE \textbf{// Sampling}
    \STATE Sample $Z_1,\ldots,Z_N \sim Q_t(Z) = \prod_{i=1}^d Q_t(Z_i)$
    \STATE \textbf{// Elite set selection}
    \STATE Define elite set $\mathcal{E}_t \leftarrow \{n : f(Z_n) =1\}$
    \IF{$\mathcal{E}_t = \emptyset$} 
        \STATE Set $Q_{t+1} \leftarrow Q_t$ and \textbf{continue}
    \ENDIF
    \STATE \textbf{// Weighted update}
    \FOR{each dimension $i$ and value $j \in \mathcal{Z}_i$}
        \STATE Compute importance weights $w(Z_n) \leftarrow P(Z_n)/Q_t(Z_n)$
        \STATE $Q'_{t+1}(Z_i)_j \leftarrow \frac{\sum_{n\in\mathcal{E}_t} w(Z_n)\,\mathbb{1}\{z_{n,i}=j\}}{\sum_{n\in\mathcal{E}_t} w(Z_n)}$
    \ENDFOR
    \STATE \textbf{// Smoothing}
    \STATE $Q_{t+1} \leftarrow \zeta Q'_{t+1} + (1-\zeta) Q_t$
\ENDFOR
\STATE Apply defensive sampling $\tilde{Q}_T(Z) \leftarrow (1-\lambda) Q_T(Z) + \lambda P(Z)$
\STATE \textbf{// Estimation with confidence bounds}
\STATE $\hat{\mu}, [\hat{\mu}_{\text{low}}, \hat{\mu}_{\text{high}}] \leftarrow$ \textsc{ComputeEstimate}$(f, P, \tilde{Q}_T, N, \alpha)$
\STATE \textbf{Return} $\hat{\mu}$, $[\hat{\mu}_{\text{low}}, \hat{\mu}_{\text{high}}]$
\end{algorithmic}
\label{algorithm}
\end{algorithm}

\paragraph{Importance Sampling with $Q$.}
\label{subsec:ce}
To estimate $\mu$ defined in Eq.~\ref{eq:target_mu},
a standard Monte Carlo estimator under $P$ is
\[
\hat{\mu}_P = \frac{1}{N}\sum_{i=1}^{N} f(Z_i), \quad Z_i \sim P.
\]
However, when failures are rare, the estimator suffers from high variance relative to $\mu$, requiring large numbers of samples to obtain tight confidence bounds.
To improve efficiency, we introduce a proposal distribution $Q$ and estimate $\mu$ via importance sampling:
\[
\hat{\mu}_Q = \frac{1}{N}\sum_{i=1}^{N}  f(Z_i)\,\frac{P(Z_i)}{Q(Z_i)}, \quad Z_i \sim Q,
\]
which yields an unbiased estimate under the support condition that $Q(Z)>0$ whenever $\mathbb{P}\left(f(Z)=1\right)P(Z)>0$.
To ensure that importance weights are bounded, we employ defensive sampling~\citep{hesterberg1995weighted}, which mixes $Q$ with $P$ with mixing coefficient $\lambda$.

\begin{figure}[t]
\centering
\begin{minipage}{\linewidth}
\centering
\begin{minipage}[t]{0.32\linewidth}
    \centering
    \textbf{\scriptsize Qwen2.5-Math-7B-Instruct,\\ Template 7}
    \includegraphics[width=\linewidth]{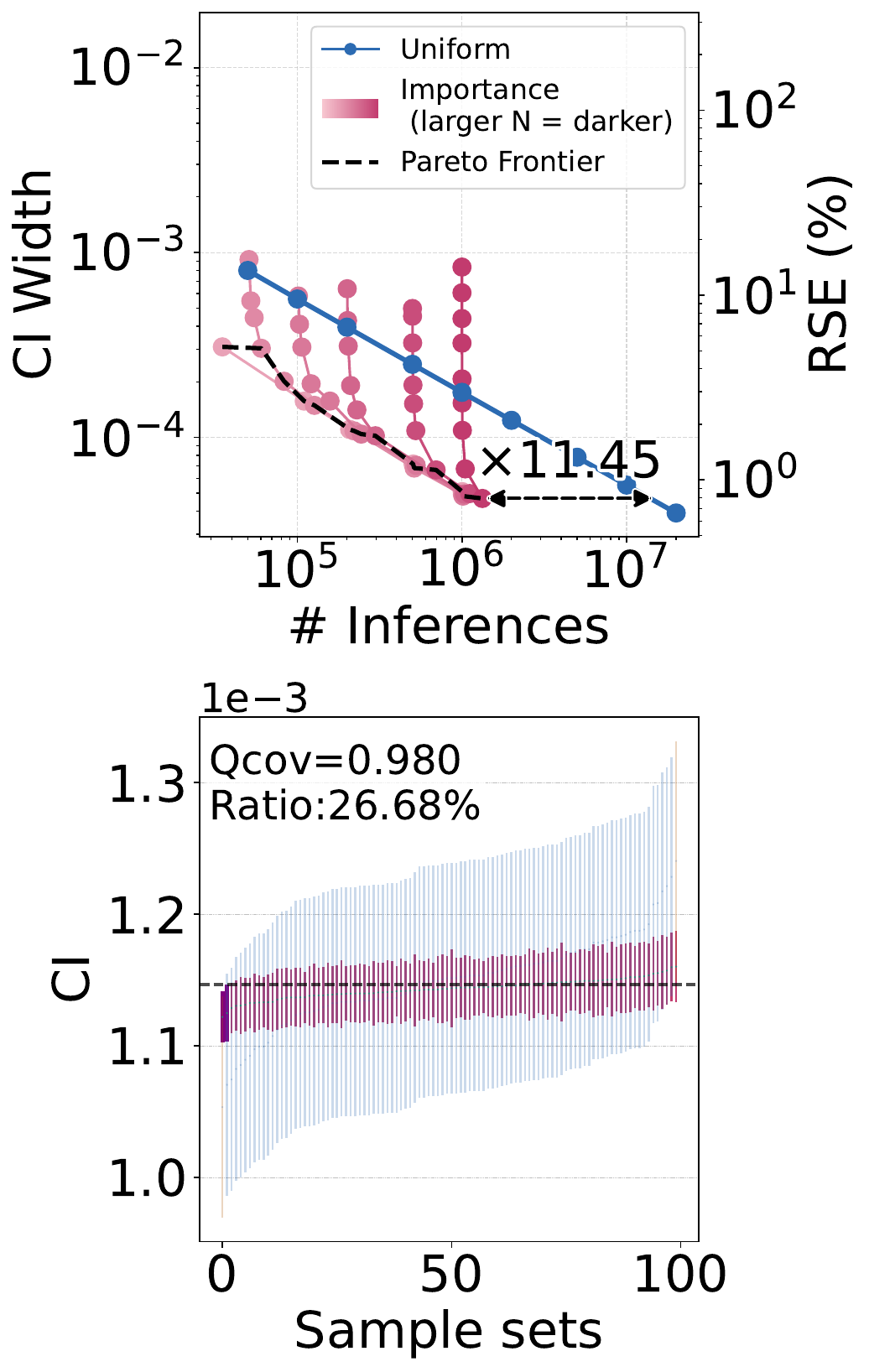}
\end{minipage}
\hfill
\begin{minipage}[t]{0.32\linewidth}
    \centering
    \textbf{\scriptsize gpt-oss-20b-low,\\ Template 2}
    \includegraphics[width=\linewidth]{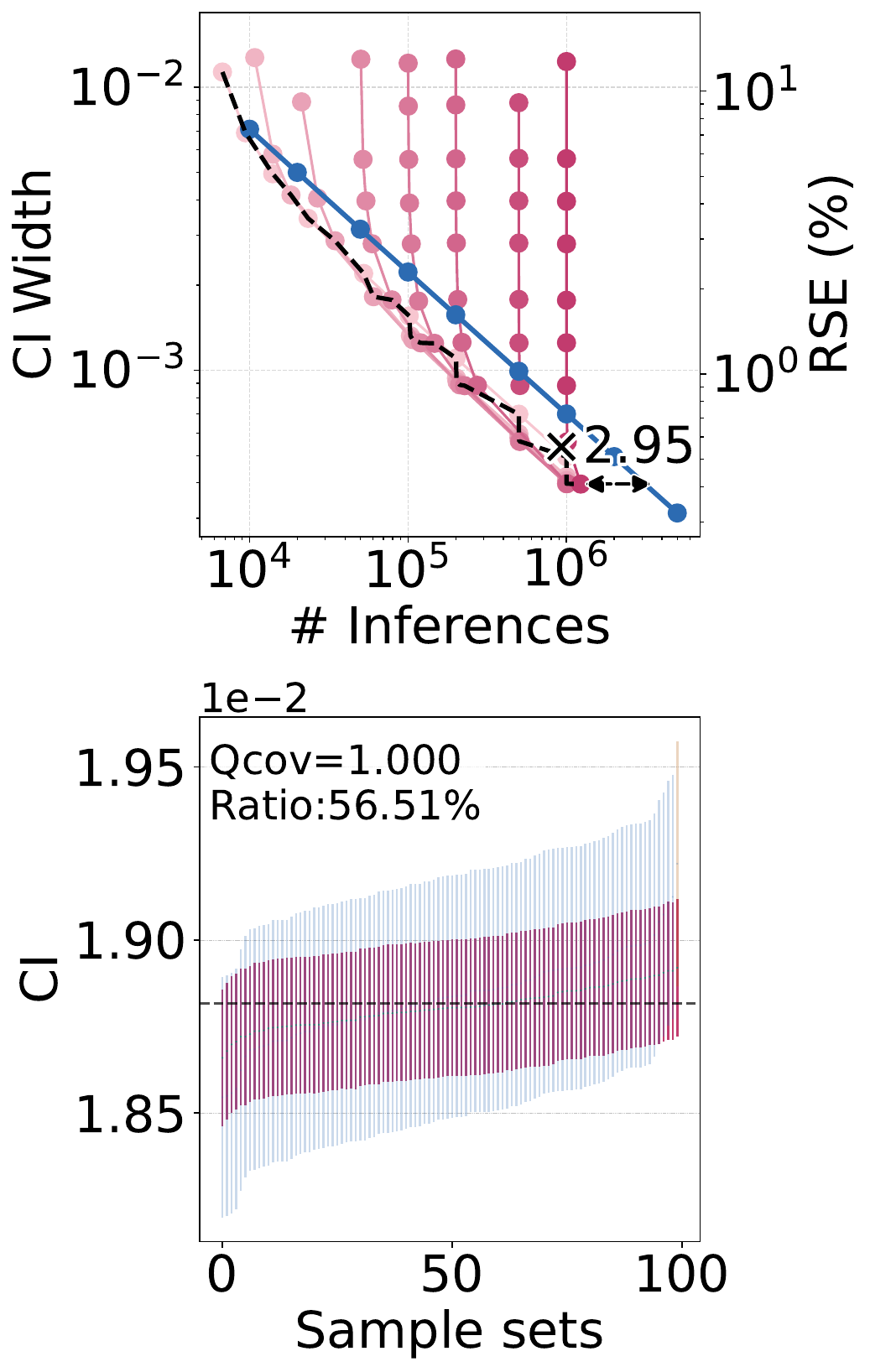}
\end{minipage}
\hfill
\begin{minipage}[t]{0.32\linewidth}
    \centering
    \textbf{\scriptsize Gemini 2.5 Flash Lite,\\ Template 2}
    \includegraphics[width=\linewidth]{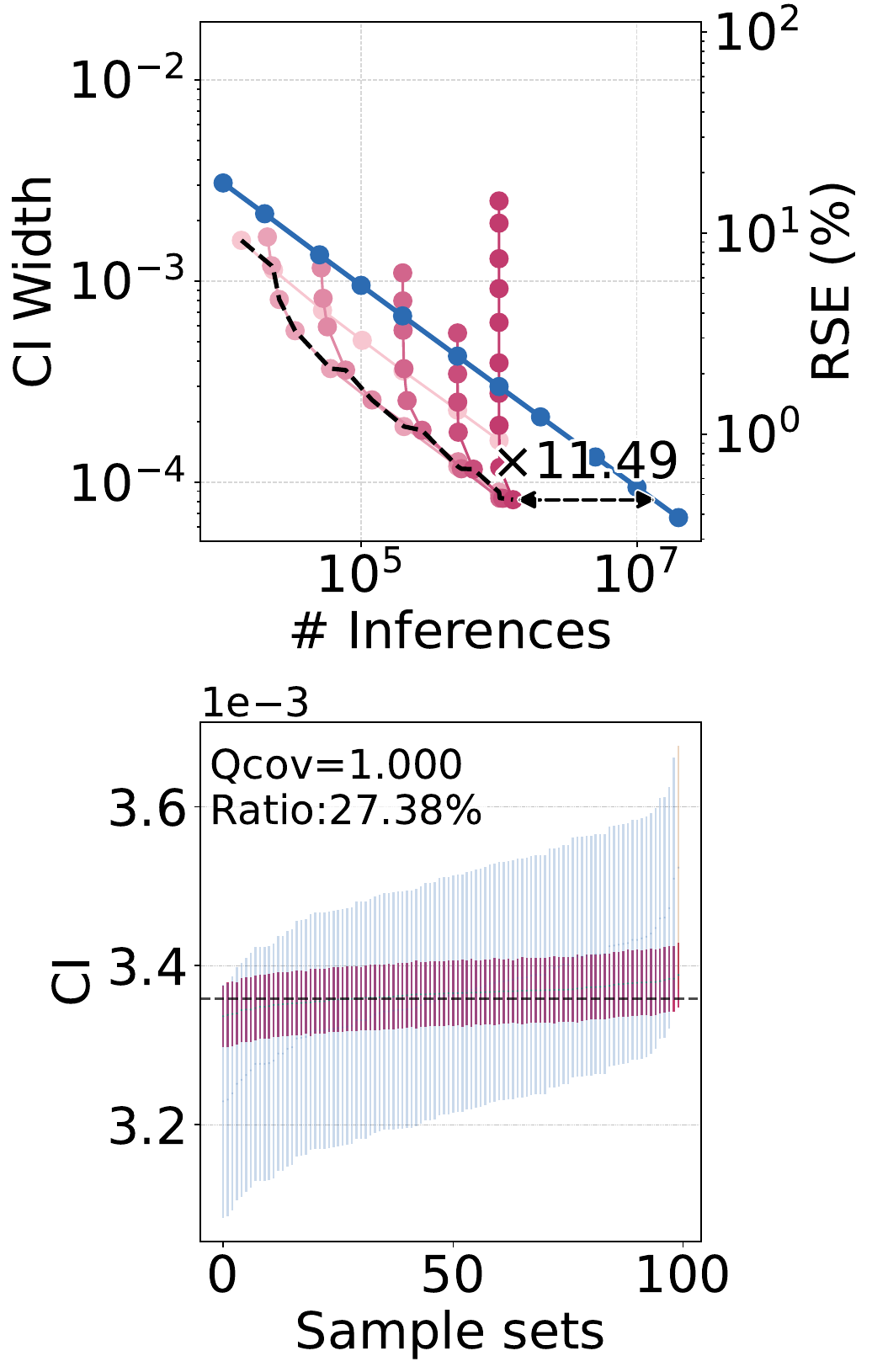}
\end{minipage}
\end{minipage}
\vspace{-0.3cm}
\caption{
\textbf{Upper row: CI width vs. number of inferences.} 
Each solid pink lines connects estimations with varying number of inferences (pink dots) under importance sampling with same number of overhead inferences for CEM ($N$), and darker pink means larger $N$. 
Similarly, solid blue line for uniform sampling.
The Pareto-frontier (dotted line) is the lower-left envelope across all $(N, \text{\# evaluation})$ combinations, which achieves a tight CI width, i.e., relative standard error (RSE) of around $1\%$, at a much smaller inference budget compared to uniform sampling
\textbf{Lower row: sorted CIs over $100$ sample sets.} At the evaluation size of the bottom-right Pareto-frontier dot, $Q$ (pink) yields much narrower confidence bounds than $P$ (blue) for the same sample size, while maintaining coverage close to $99\%$.
We highlight the intervals that fail to cover the population mean in dark pink.
}
\label{fig:curve_and_plots}
\end{figure}

\paragraph{CEM for learning $Q$.}
The variance-minimizing proposal distribution is given by $Q^*(Z) = \frac{\mathbb{P}\left(f(Z)=1\right)P(Z)}{\mu}$, which concentrates probability mass on failure regions. 
The CEM seeks a proposal distribution $Q$ that approximates this optimal distribution by minimizing the Kullback-Leibler divergence, $\text{KL}\left(Q^*\parallel Q\right)$.
Since $Q^*$ depends on the unknown $\mu$, the CEM iteratively refines $Q$ by sampling from the current proposal and updating toward the empirical failure-weighted distribution proportional to $\mathbb{P}\left(f(Z)=1\right)P(Z)$.

\paragraph{Failure probability estimation via $Q$ learned from CEM.}
To begin with, we model $Q$ as a fully factorized distribution:
\begin{align}
    Q(Z)=\prod_{i=1}^{d}Q(Z_i),
\end{align}
where each $Q(Z_i)$ is a categorical distribution over $\mathcal{Z}_i$.
This independence assumption provides a tractable approximation to model the joint parameter space.
We initialize $Q_0(Z)$ as a uniform distribution over each parameter dimension $Z_i\in \mathcal{Z}_i$.

We then apply the CEM followed by importance sampling for estimating true failure probability with $Q$, as described in 
Algorithms~\ref{algorithm} and~\ref{algorithm:compute_estimate}.
\section{Order-of-Magnitude Inference Efficiency Gains}

\paragraph{Setup for CEM}
For CEM, we use fixed hyperparameters across all models and templates: $T{=}10$ iterations, smoothing coefficient $\zeta{=}0.1$, and defensive sampling coefficient $\lambda \in \{0.1, 0.3, 0.5\}$.
The CEM sample size $N$, the number of data for updating $Q$ at each CEM iteration $t$, is set as a hyperparameter, ranging between $0.5$K and $100$K.
\paragraph{Setup for estimation.}
For the final estimation, we vary the evaluation sample size $M$ between $1$K and $1$M (million).
We define the \textit{number of inferences} as the number of samples from $P$ or $Q$ (one inference per $Z$, independent of $K$).
Because the $T{\times}N$ samples drawn during CEM can be reused directly as importance sampling estimation samples, the total number of inferences under $Q$ is $\max(T{\times}N,\,M)$: if the CEM draws already reach or exceed $M$, no additional inferences are needed; otherwise, $M - T{\times}N$ samples are drawn from $\tilde{Q}_T$.
For $P$, the cost is simply $M$ inferences with no CEM phase.
\paragraph{Setup for evaluation.}
For all evaluations, we construct $99\%$ confidence intervals (significance level $\alpha=0.01$) using the normal approximation applied to the importance-weighted estimator $\hat{\mu}_Q$.
To verify the unbiasedness of our estimator, we draw $100$ independent sample sets for each sample size and report the estimates whose coverage falls within $99\pm1\%$ (i.e., $1\%$ tolerance of the nominal coverage level).
For calculating coverage, we use $\hat{\mu}_P$ computed using $1$B bootstrap samples from $P$, which converges to the unknown true mean.

\begin{table}[t]
\centering
\scalebox{0.7}{
\setlength{\tabcolsep}{3pt}
\begin{tabular}{c|ccc|ccc|ccc|ccc|ccc|ccc}
\toprule
& \multicolumn{6}{c|}{$\lambda = 0.1$} & \multicolumn{6}{c|}{$\lambda = 0.3$} & \multicolumn{6}{c}{$\lambda = 0.5$}\\
\cmidrule(lr){2-7}\cmidrule(lr){8-13}\cmidrule(lr){14-19}
& \multicolumn{3}{c|}{\textbf{Gains}} & \multicolumn{3}{c|}{\textbf{RSE} (\%)} & \multicolumn{3}{c|}{\textbf{Gains}} & \multicolumn{3}{c|}{\textbf{RSE} (\%)} & \multicolumn{3}{c|}{\textbf{Gains}} & \multicolumn{3}{c}{\textbf{RSE} (\%)}\\
\midrule
\multicolumn{19}{c}{\textbf{Qwen2.5-Math-7B-Instruct}}\\
\midrule
ID & $K{=}4$ & $K{=}8$ & $K{=}16$ & $K{=}4$ & $K{=}8$ & $K{=}16$ & $K{=}4$ & $K{=}8$ & $K{=}16$ & $K{=}4$ & $K{=}8$ & $K{=}16$ & $K{=}4$ & $K{=}8$ & $K{=}16$ & $K{=}4$ & $K{=}8$ & $K{=}16$\\
\midrule
0 & \cellcolor{blue!18}3.68 & \cellcolor{blue!33}6.50 & \cellcolor{blue!50}16.41 & 1.24 & 1.22 & 1.02 & \cellcolor{blue!23}4.51 & \cellcolor{blue!40}8.03 & \cellcolor{blue!50}15.84 & 1.27 & 1.21 & 1.05 & \cellcolor{blue!19}3.82 & \cellcolor{blue!33}6.59 & \cellcolor{blue!50}13.65 & 1.27 & 1.26 & 1.13 \\
1 & \cellcolor{blue!7}1.31 & \cellcolor{blue!11}2.29 & \cellcolor{blue!27}5.33 & 1.97 & 2.99 & 2.95 & \cellcolor{blue!8}1.58 & \cellcolor{blue!12}2.34 & \cellcolor{blue!35}7.02 & 1.94 & 3.00 & 2.46 & \cellcolor{blue!8}1.57 & \cellcolor{blue!12}2.30 & \cellcolor{blue!30}6.02 & 1.95 & 3.21 & 2.69 \\
2 & \cellcolor{blue!8}1.63 & \cellcolor{blue!7}1.48 & \cellcolor{blue!9}1.87 & 0.22 & 0.24 & 0.27 & \cellcolor{blue!7}1.37 & \cellcolor{blue!7}1.48 & \cellcolor{blue!9}1.80 & 0.22 & 0.25 & 0.27 & \cellcolor{blue!8}1.51 & \cellcolor{blue!8}1.61 & \cellcolor{blue!8}1.68 & 0.22 & 0.26 & 0.29 \\
3 & \cellcolor{blue!5}1.07 & \cellcolor{blue!7}1.46 & \cellcolor{blue!9}1.75 & 0.43 & 0.63 & 0.80 & \cellcolor{blue!6}1.25 & \cellcolor{blue!8}1.50 & \cellcolor{blue!7}1.49 & 0.43 & 0.63 & 0.81 & \cellcolor{blue!6}1.24 & \cellcolor{blue!6}1.25 & \cellcolor{blue!8}1.64 & 0.43 & 0.64 & 0.83 \\
4 & \cellcolor{blue!7}1.34 & \cellcolor{blue!7}1.38 & \cellcolor{blue!6}1.22 & 0.31 & 0.36 & 0.40 & \cellcolor{blue!7}1.30 & \cellcolor{blue!7}1.38 & \cellcolor{blue!6}1.23 & 0.31 & 0.36 & 0.40 & \cellcolor{blue!6}1.29 & \cellcolor{blue!7}1.33 & \cellcolor{blue!7}1.41 & 0.31 & 0.37 & 0.41 \\
5 & \cellcolor{blue!11}2.15 & \cellcolor{blue!18}3.67 & \cellcolor{blue!35}6.93 & 1.20 & 1.77 & 5.05 & \cellcolor{blue!13}2.51 & \cellcolor{blue!21}4.28 & \cellcolor{blue!27}5.40 & 1.14 & 1.87 & 2.81 & \cellcolor{blue!12}2.32 & \cellcolor{blue!18}3.69 & \cellcolor{blue!28}5.62 & 1.19 & 1.85 & 2.78 \\
6 & \cellcolor{blue!27}5.36 & \cellcolor{blue!50}15.84 & \cellcolor{blue!50}17.12 & 0.44 & 0.44 & 0.43 & \cellcolor{blue!32}6.40 & \cellcolor{blue!50}12.87 & \cellcolor{blue!50}15.92 & 0.49 & 0.49 & 0.47 & \cellcolor{blue!29}5.80 & \cellcolor{blue!47}9.46 & \cellcolor{blue!50}11.74 & 0.56 & 0.56 & 0.54 \\
7 & \cellcolor{blue!16}3.22 & \cellcolor{blue!40}8.02 & \cellcolor{blue!50}11.45 & 1.08 & 0.95 & 0.79 & \cellcolor{blue!18}3.53 & \cellcolor{blue!42}8.45 & \cellcolor{blue!50}14.66 & 1.03 & 0.89 & 0.80 & \cellcolor{blue!22}4.44 & \cellcolor{blue!32}6.42 & \cellcolor{blue!50}11.59 & 1.03 & 0.96 & 0.87 \\
8 & \cellcolor{blue!9}1.91 & --- & --- & 4.97 & --- & --- & \cellcolor{blue!14}2.73 & --- & --- & 4.09 & --- & --- & \cellcolor{blue!13}2.45 & --- & --- & 4.57 & --- & --- \\
\midrule
\multicolumn{19}{c}{\textbf{gpt-oss-20b-low}}\\
\midrule
ID & $K{=}8$ & $K{=}16$ & $K{=}24$ & $K{=}8$ & $K{=}16$ & $K{=}24$ & $K{=}8$ & $K{=}16$ & $K{=}24$ & $K{=}8$ & $K{=}16$ & $K{=}24$ & $K{=}8$ & $K{=}16$ & $K{=}24$ & $K{=}8$ & $K{=}16$ & $K{=}24$\\
\midrule
0 & \cellcolor{blue!17}3.34 & \cellcolor{blue!19}3.81 & \cellcolor{blue!18}3.63 & 2.56 & 2.95 & 3.92 & \cellcolor{blue!15}2.95 & \cellcolor{blue!19}3.79 & \cellcolor{blue!34}6.77 & 2.52 & 3.10 & 2.85 & \cellcolor{blue!14}2.71 & \cellcolor{blue!22}4.45 & \cellcolor{blue!18}3.54 & 2.63 & 3.11 & 4.15 \\
2 & \cellcolor{blue!11}2.27 & \cellcolor{blue!14}2.70 & \cellcolor{blue!15}2.95 & 0.28 & 0.37 & 0.41 & \cellcolor{blue!11}2.28 & \cellcolor{blue!14}2.86 & \cellcolor{blue!16}3.28 & 0.30 & 0.39 & 0.43 & \cellcolor{blue!10}1.93 & \cellcolor{blue!12}2.35 & \cellcolor{blue!14}2.81 & 0.32 & 0.42 & 0.46 \\
4 & \cellcolor{blue!7}1.30 & \cellcolor{blue!10}1.93 & \cellcolor{blue!13}2.65 & 0.79 & 1.19 & 1.36 & \cellcolor{blue!8}1.55 & \cellcolor{blue!8}1.69 & \cellcolor{blue!12}2.30 & 0.79 & 1.16 & 1.34 & \cellcolor{blue!8}1.53 & \cellcolor{blue!10}1.91 & \cellcolor{blue!10}2.09 & 0.80 & 1.19 & 1.40 \\
7 & \cellcolor{blue!15}2.89 & \cellcolor{blue!14}2.80 & \cellcolor{blue!23}4.45 & 1.61 & 2.76 & 2.29 & \cellcolor{blue!15}3.07 & \cellcolor{blue!17}3.50 & \cellcolor{blue!27}5.22 & 1.62 & 2.14 & 2.18 & \cellcolor{blue!12}2.39 & \cellcolor{blue!21}4.29 & \cellcolor{blue!27}5.39 & 1.65 & 2.11 & 2.31 \\
8 & \cellcolor{blue!22}4.36 & \cellcolor{blue!22}4.41 & \cellcolor{blue!23}4.57 & 0.84 & 1.18 & 1.36 & \cellcolor{blue!20}4.02 & \cellcolor{blue!22}4.42 & \cellcolor{blue!23}4.65 & 0.90 & 1.21 & 1.42 & \cellcolor{blue!15}2.99 & \cellcolor{blue!17}3.43 & \cellcolor{blue!19}3.70 & 0.97 & 1.32 & 1.49 \\
\midrule
\multicolumn{19}{c}{\textbf{Gemini 2.5 Flash Lite}}\\
\midrule
ID & $K{=}4$ & $K{=}8$ & $K{=}16$ & $K{=}4$ & $K{=}8$ & $K{=}16$ & $K{=}4$ & $K{=}8$ & $K{=}16$ & $K{=}4$ & $K{=}8$ & $K{=}16$ & $K{=}4$ & $K{=}8$ & $K{=}16$ & $K{=}4$ & $K{=}8$ & $K{=}16$\\
\midrule
0 & \cellcolor{blue!1}0.96 & \cellcolor{blue!20}4.05 & \cellcolor{blue!26}5.29 & 1.95 & 1.90 & 2.35 & \cellcolor{blue!5}1.23 & \cellcolor{blue!24}4.74 & \cellcolor{blue!41}8.23 & 1.64 & 1.72 & 1.74 & \cellcolor{blue!5}1.35 & \cellcolor{blue!25}4.90 & \cellcolor{blue!43}8.62 & 1.57 & 1.71 & 1.93 \\
2 & \cellcolor{blue!14}2.54 & \cellcolor{blue!22}4.41 & \cellcolor{blue!50}11.49 & 0.68 & 0.62 & 0.47 & \cellcolor{blue!14}2.76 & \cellcolor{blue!28}5.69 & \cellcolor{blue!50}10.35 & 0.65 & 0.62 & 0.51 & \cellcolor{blue!13}2.38 & \cellcolor{blue!26}5.16 & \cellcolor{blue!48}9.50 & 0.66 & 0.64 & 0.57 \\
4 & \cellcolor{blue!3}1.12 & \cellcolor{blue!1}0.99 & \cellcolor{blue!1}1.00 & 0.40 & 0.51 & 0.58 & \cellcolor{blue!3}1.11 & \cellcolor{blue!5}1.18 & \cellcolor{blue!5}1.19 & 0.40 & 0.51 & 0.58 & \cellcolor{blue!1}0.95 & \cellcolor{blue!3}1.16 & \cellcolor{blue!1}1.00 & 0.40 & 0.51 & 0.58\\
6 & \cellcolor{blue!2}1.06 & \cellcolor{blue!14}2.75 & \cellcolor{blue!50}156.22 & 0.99 & 1.80 & 0.32 & \cellcolor{blue!2}1.10 & \cellcolor{blue!17}3.45 & \cellcolor{blue!50}85.21 & 0.98 & 1.63 & 0.36 & \cellcolor{blue!4}1.27 & \cellcolor{blue!17}3.32 & \cellcolor{blue!50}82.59 & 0.98 & 1.52 & 0.37 \\
\bottomrule
\end{tabular}
}
\vspace{-0.2cm}
\caption{Sampling efficiency gains and RSE (\%) across templates, models, $K$, and $\lambda$ for majority voting. Cell shading for gains is proportional to the gain value, i.e., the higher the darker, capped at gain $\geq 10$.
--- denotes zero failures.
}
\label{tab:gain_error}
\end{table}

We compare against uniform sampling from $P$ with confidence bounds computed via the Clopper-Pearson binomial exact method~\citep{clopper1934}.
To quantify the inference efficiency obtained by importance sampling, we present the \textit{sampling efficiency gain}, calculated by the number of inferences under $P$ divided by that under $Q$ (i.e., $\max(T{\times}N, M)$) to obtain the same CI width.
Across $N$ and evaluation sizes, we report the gains of the Pareto-frontier in the CI width vs. number of inferences trade-off (i.e., dotted line in Figures~\ref{fig:main_curve} and~\ref{fig:curve_and_plots}).
We additionally report \textit{RSE} defined as $\text{RSE}=\frac{\text{SE}}{\hat{\mu}}$ to evaluate the tightness of the confidence bounds.

\paragraph{Failure-prone sampling saves an order of magnitude in inferences.}
Our results demonstrate that the learned distribution $Q$, which concentrates on frequent failure patterns, yields substantially tighter confidence bounds compared to uniform sampling under $P$.
In Figure~\ref{fig:main_curve} and upper row of Figure~\ref{fig:curve_and_plots}, we present CI width under $Q$ (pink) for varying evaluation sizes (dots connected by lines) with different $N$, the number of samples for CEM (different shades).
The Pareto-frontier (dotted line), which connects the minimum number of inferences required to achieve each CI width, decreases much more rapidly under $Q$ as the number of inferences increases than under $P$ (blue).

This indicates that, for a tight confidence bound such as an RSE of $1\%$, importance sampling under $Q$ requires significantly fewer inferences than uniform sampling from $P$.
Specifically, such inference efficiency gains span orders of magnitude across models and templates, e.g., $17.12\times$ for Qwen2.5-Math-7B-Instruct on Template $6$ ($K=16$), $4.57\times$ for gpt-oss-20b-low on Template $8$ ($K=24$), and $156.22\times$ for Gemini 2.5 Flash Lite on Template $6$ ($K=16$).

These inference efficiency gains stem from the tight CIs obtained through importance sampling, as illustrated in Figure~\ref{fig:curve_and_plots} bottom row.
The plots compare the confidence bounds between $P$ and $Q$ across $100$ independent sample sets (sorted by their center).
Across evaluation sizes, we plot the bounds for the sample size corresponding to the bottom-right pink dot on the Pareto-frontier.
Intervals under $Q$ (pink) are much narrower than those under $P$ (blue) for the same sample size.
These $100$ intervals faithfully bound the $\hat{\mu}_P$ (with $\pm 1$ tolerance), confirming that our method meets the nominal $99\%$ confidence level.

We report inference efficiency gains and their RSE across models and templates in Table~\ref{tab:gain_error}.
For each gain, we add darker shading for larger values for visibility.
The results demonstrate that our method achieves consistent efficiency gains, where most exceed $2\times$.
These estimations show substantially low RSE, where most are around just $1\%$, implying that we can estimate LLMs' true error probability with highly accurate estimation across templates and models.

We observe that inference efficiency gains improve as we use larger $K$ for self-consistency decoding~\citep{wang2023selfconsistency}, i.e., shading becomes darker as $K$ increases in gains.
For instance, Qwen2.5-Math-7B-Instruct shows a gain of $3.68\times$ with $K=4$ increasing to $16.41\times$ with $K=16$ on Template 0, sharing the trends across setups.
These results corroborate our empirical observations that systematic failures become more concentrated with higher $K$.
In addition, overall gains and RSE remain consistent across different $\lambda$, which allows practitioners to use our method without careful hyperparameter tuning.

Similarly, efficiency gains are strongly correlated with failure rarity: as models become more saturated on a task, failures concentrate on a smaller subset of inputs, making importance sampling more effective.
As a concrete example, Gemini 2.5 Flash Lite shows higher error rate ($\hat{\mu}_Q$) on Template 4, i.e., $2.59\%$, than others ($0.335\%$-$0.0328\%$) with $K=16$
correspondingly, those high error-rate templates also exhibit lower TV distance (Table~\ref{tab:tv_table}) and smaller efficiency gains (Table~\ref{tab:gain_error}), which is consistent with our finding that stronger failure concentration drives higher importance sampling efficiency.

\section{Reliability Evaluation Between Models on GSM}
\begin{figure}[t]
\centering
\begin{minipage}[t]{\linewidth}
    \centering
    \includegraphics[width=\linewidth]{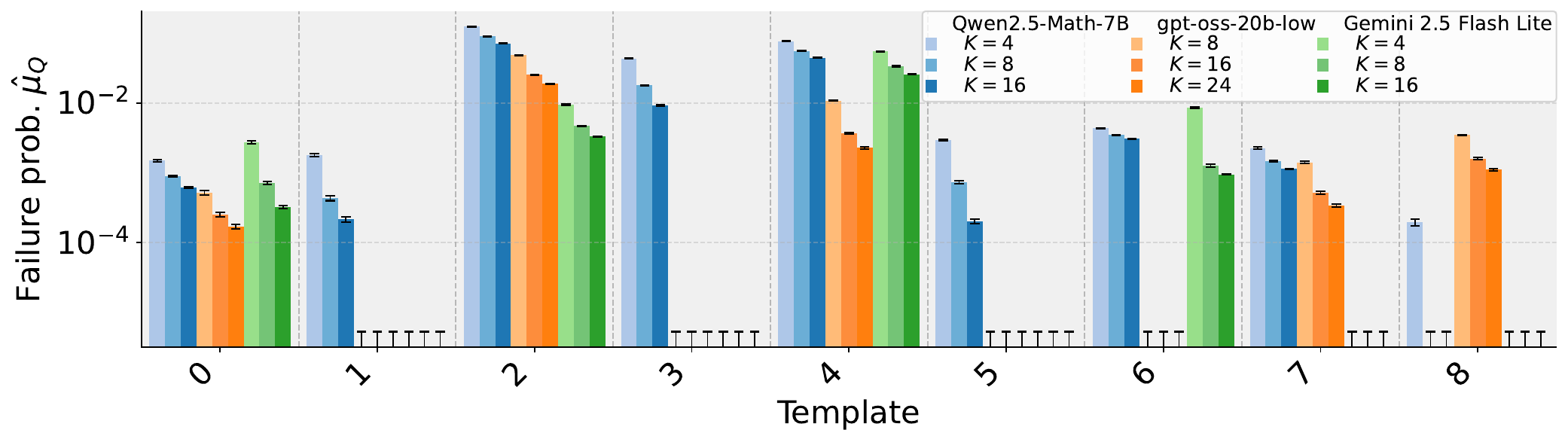}
\end{minipage}
\vspace{-0.5cm}
\caption{Estimated failure probability $\hat{\mu}_Q$ across templates and models at $\lambda = 0.1$ and evaluation size of $1M$, with error bars denoting the half-width of the confidence interval.
For those with zero failures, we report the error bars of exact binomial bounds, i.e., (0, 5.30e-06).
}
\label{fig:eval}
\end{figure}

Finally, Figure~\ref{fig:eval} reports each model's estimated rare failure probability $\hat{\mu}_Q$ across GSM templates, with evaluation size of $M=1$M, presented as a point estimate with confidence bounds. 
While all models achieve saturating performance with error rates well below $1\%$, our evaluation reveals meaningful differences in reliability at finer granularity.
Specifically, gpt-oss-20b-low gives a failure rate of just $0.0254\%$ on Template 0 with $K=16$, which outperforms Qwen2.5-Math-7B-Instruct ($0.0614\%$, $\sim2.5\times$ higher) and Gemini 2.5 Flash Lite ($0.0328\%$, $\sim1.3\times$ higher).
These results highlight that such saturated benchmarks still carry discriminative signal for reliability, and such evaluation encourages the community to pursue models that are not only more capable but more reliably so.
For all template-model pairs that include zero failures, we report the exact binomial bounds, i.e., (0, 0.00053\%).
The full table is provided in Table~\ref{appendix:eval}.

\section{Limitations, Future Work, and Conclusion}
\label{sec:discussion_limitation}

\paragraph{Limitations.}
Our method relies on the assumption that LLM errors concentrate on a small subset of the parameter space, which we validate empirically across the models and templates we study.
When error patterns are diffuse, e.g., when the model is relatively weak for a task, the CEM may not find a substantially better proposal than uniform sampling.
Our method may also be less effective when estimating failure rates without majority voting ($K{=}1$), where stochastic generation outputs produce inconsistent failure patterns that are harder to concentrate on.

In addition, due to computational constraints, our experiments focus on relatively small LLMs.
Evaluating reliability introduces significant overhead: the required number of inferences grows with the evaluation sample size, the majority-vote ensemble size $K$, and the number of independent trials for coverage verification, all of which increase as failures become rarer or as the desired confidence level tightens.
Extending this framework to larger, more capable LLMs therefore remains an important direction for future work.

\paragraph{Lack of reliability benchmarks.}
While nearly all existing benchmarks focus on evaluating how LLMs push the frontier in challenging tasks, to our knowledge, no benchmark specifically tests their reliability on already-saturated benchmarks.
Rather than revisiting these saturated benchmarks, the community tends to curate novel, more challenging ones, leaving a gap in our understanding of model reliability.
In this paper, we take a first step toward addressing this gap by introducing a prototypical framework for evaluating LLM reliability using parameterized GSM templates.
However, reliability evaluation must be extended to broader domains, including safety guardrails, code generation for production systems, and medical or legal question answering, where even rare failures can have serious consequences.
For instance, GPT-5.2-Thinking achieves a 99.9441\% no-violation rate on internal safety benchmarks~\citep{openai2026gpt54benchmarks}, yet the remaining $0.0559\%$ violation rate still corresponds to over 500 failures per million queries, underscoring the importance of rigorous reliability measurement even in near-perfect regimes.

\paragraph{Surgical fixes for systematic failures.}
Identifying model failures that are consistently triggered by certain input prompts could enable more targeted reliability improvements.
Our method directly supports this: the learned proposal distribution $Q$ highlights the failure-inducing parameter values, which practitioners can use as targeted training data or test cases to surgically correct systematic failures via fine-tuning on problematic inputs or adding safeguards rather than retraining broadly.
The ultimate goal is to leverage these systematic failure patterns to push model reliability into the five-nines regime, enabling safe deployment in high-stakes applications.

\paragraph{Conclusion.}
We present an inference-efficient framework for estimating the true error probability of highly accurate LLMs, demonstrated on parameterized GSM problems.
LLM failures are strongly concentrated on a small subset of inputs, as evidenced by high total variation distances from the uniform distribution across all models and templates we study.
Leveraging this finding, we adopt the CEM to learn a failure-prone sampling distribution, achieving up to $156.22\times$ inference savings over uniform sampling for equivalent confidence bounds.
We call on the community to treat reliability as a first-class evaluation criterion alongside capability; our framework provides a practical foundation for rigorous reliability measurement in high-stakes LLM deployments.

\section{Acknowledgements}
We thank Yixuan Xu, Bingqing Chen, and Chen Qiu for insightful discussions. Eungyeup Kim is supported by funding from Bosch Center for Artificial Intelligence.
Vashisth Tiwari is funded by the \href{https://ror.org/05xpvk416}{National Institute of Standards and Technology} under Federal Award ID Number 60NANB24D231 and \href{https://ror.org/05x2bcf33}{Carnegie Mellon University AI Measurement Science and Engineering Center (AIMSEC)}.

\bibliography{colm2026_conference}

@inproceedings{mirzadeh2025gsmsymbolic,
title={{GSM}-Symbolic: Understanding the Limitations of Mathematical Reasoning in Large Language Models},
author={Seyed Iman Mirzadeh and Keivan Alizadeh and Hooman Shahrokhi and Oncel Tuzel and Samy Bengio and Mehrdad Farajtabar},
booktitle={The Thirteenth International Conference on Learning Representations},
year={2025},
url={https://openreview.net/forum?id=AjXkRZIvjB}
}

@article{cobbe2021gsm8k,
  title={Training Verifiers to Solve Math Word Problems},
  author={Cobbe, Karl and Kosaraju, Vineet and Bavarian, Mohammad and Chen, Mark and Jun, Heewoo and Kaiser, Lukasz and Plappert, Matthias and Tworek, Jerry and Hilton, Jacob and Nakano, Reiichiro and Hesse, Christopher and Schulman, John},
  journal={arXiv preprint arXiv:2110.14168},
  year={2021}
}

@inproceedings{
wang2023selfconsistency,
title={Self-Consistency Improves Chain of Thought Reasoning in Language Models},
author={Xuezhi Wang and Jason Wei and Dale Schuurmans and Quoc V Le and Ed H. Chi and Sharan Narang and Aakanksha Chowdhery and Denny Zhou},
booktitle={The Eleventh International Conference on Learning Representations },
year={2023},
url={https://openreview.net/forum?id=1PL1NIMMrw}
}

@article{hesterberg1995weighted,
  title={Weighted average importance sampling and defensive mixture distributions},
  author={Hesterberg, Tim},
  journal={Technometrics},
  volume={37},
  number={2},
  pages={185--194},
  year={1995},
  publisher={Taylor \& Francis}
}

@article{yang2024qwen25math,
  title={Qwen2.5-Math Technical Report: Toward Mathematical Expert Model via Self-Improvement}, 
  author={An Yang and Beichen Zhang and Binyuan Hui and Bofei Gao and Bowen Yu and Chengpeng Li and Dayiheng Liu and Jianhong Tu and Jingren Zhou and Junyang Lin and Keming Lu and Mingfeng Xue and Runji Lin and Tianyu Liu and Xingzhang Ren and Zhenru Zhang},
  journal={arXiv preprint arXiv:2409.12122},
  year={2024}
}

@article{guo2025deepseek,
  title={Deepseek-r1: Incentivizing reasoning capability in llms via reinforcement learning},
  author={Guo, Daya and Yang, Dejian and Zhang, Haowei and Song, Junxiao and Wang, Peiyi and Zhu, Qihao and Xu, Runxin and Zhang, Ruoyu and Ma, Shirong and Bi, Xiao and others},
  journal={arXiv preprint arXiv:2501.12948},
  year={2025}
}

@article{agarwal2025gpt,
  title={gpt-oss-120b \& gpt-oss-20b model card},
  author={Agarwal, Sandhini and Ahmad, Lama and Ai, Jason and Altman, Sam and Applebaum, Andy and Arbus, Edwin and Arora, Rahul K and Bai, Yu and Baker, Bowen and Bao, Haiming and others},
  journal={arXiv preprint arXiv:2508.10925},
  year={2025}
}

@misc{anthropic2026opus4.6,
  author={Anthropic},
  title={Claude Opus 4.6 System Card},
  year={2026},
  url={https://anthropic.com/claude-opus-4-6-system-card}
}

@misc{openai2025gpt52,
  author       = {OpenAI},
  title        = {Introducing GPT-5.2},
  year         = {2025},
  url          = {https://openai.com/index/introducing-gpt-5-2/}
}

@misc{openai2024o1,
  author       = {OpenAI},
  title        = {OpenAI o1 System Card},
  year         = {2024},
  url          = {https://openai.com/index/openai-o1-system-card/}
}

@inproceedings{rein2024gpqa,
title={{GPQA}: A Graduate-Level Google-Proof Q\&A Benchmark},
author={David Rein and Betty Li Hou and Asa Cooper Stickland and Jackson Petty and Richard Yuanzhe Pang and Julien Dirani and Julian Michael and Samuel R. Bowman},
booktitle={First Conference on Language Modeling},
year={2024},
url={https://openreview.net/forum?id=Ti67584b98}
}

@misc{google_gemini3_2025,
  author       = {{Google DeepMind}},
  title        = {Gemini 3 Pro Model Card},
  year         = {2025},
  url          = {https://storage.googleapis.com/deepmind-media/Model-Cards/Gemini-3-Pro-Model-Card.pdf}
}

@article{clopper1934,
 ISSN = {00063444, 14643510},
 URL = {http://www.jstor.org/stable/2331986},
 author = {C. J. Clopper and E. S. Pearson},
 journal = {Biometrika},
 number = {4},
 pages = {404--413},
 publisher = {[Oxford University Press, Biometrika Trust]},
 title = {The Use of Confidence or Fiducial Limits Illustrated in the Case of the Binomial},
 urldate = {2026-04-01},
 volume = {26},
 year = {1934}
}

@article{rubinstein1999cross,
  title={The cross-entropy method for combinatorial and continuous optimization},
  author={Rubinstein, Reuven},
  journal={Methodology and computing in applied probability},
  volume={1},
  number={2},
  pages={127--190},
  year={1999},
  publisher={Springer}
}

@article{de2005tutorial,
  title={A tutorial on the cross-entropy method},
  author={De Boer, Pieter-Tjerk and Kroese, Dirk P and Mannor, Shie and Rubinstein, Reuven Y},
  journal={Annals of operations research},
  volume={134},
  number={1},
  pages={19--67},
  year={2005},
  publisher={Springer}
}

@book{rubinstein2004cross,
  title={The cross-entropy method: a unified approach to combinatorial optimization, Monte-Carlo simulation, and machine learning},
  author={Rubinstein, Reuven Y and Kroese, Dirk P},
  volume={133},
  year={2004},
  publisher={Springer}
}

@article{efron1979Bootstrap,
author = {B. Efron},
title = {{Bootstrap Methods: Another Look at the Jackknife}},
volume = {7},
journal = {The Annals of Statistics},
number = {1},
publisher = {Institute of Mathematical Statistics},
pages = {1 -- 26},
year = {1979},
doi = {10.1214/aos/1176344552},
URL = {https://doi.org/10.1214/aos/1176344552}
}

@misc{openai2026gpt54benchmarks,
  author       = {{OpenAI}},
  title        = {GPT-5.4 Thinking System Card},
  year         = {2026},
  howpublished = {\url{https://deploymentsafety.openai.com/gpt-5-4-thinking/gpt-5-4-thinking.pdf}},
}

@InProceedings{polo2024tinybenchmarks,
  title = 	 {tiny{B}enchmarks: evaluating {LLM}s with fewer examples},
  author =       {Maia Polo, Felipe and Weber, Lucas and Choshen, Leshem and Sun, Yuekai and Xu, Gongjun and Yurochkin, Mikhail},
  booktitle = 	 {Proceedings of the 41st International Conference on Machine Learning},
  pages = 	 {34303--34326},
  year = 	 {2024},
  volume = 	 {235},
  series = 	 {Proceedings of Machine Learning Research},
  month = 	 {21--27 Jul},
  publisher =    {PMLR},
  url = 	 {https://proceedings.mlr.press/v235/maia-polo24a.html},
}

@inproceedings{perlitz2024efficient,
  title={Efficient benchmarking (of language models)},
  author={Perlitz, Yotam and Bandel, Elron and Gera, Ariel and Arviv, Ofir and Ein-Dor, Liat and Shnarch, Eyal and Slonim, Noam and Shmueli-Scheuer, Michal and Choshen, Leshem},
  booktitle={Proceedings of the 2024 Conference of the North American Chapter of the Association for Computational Linguistics: Human Language Technologies (Volume 1: Long Papers)},
  pages={2519--2536},
  year={2024}
}

@inproceedings{vivek2024anchor,
  title={Anchor points: Benchmarking models with much fewer examples},
  author={Vivek, Rajan and Ethayarajh, Kawin and Yang, Diyi and Kiela, Douwe},
  booktitle={Proceedings of the 18th Conference of the European Chapter of the Association for Computational Linguistics (Volume 1: Long Papers)},
  pages={1576--1601},
  year={2024}
}

@article{hofmann2025fluid,
  title={Fluid language model benchmarking},
  author={Hofmann, Valentin and Heineman, David and Magnusson, Ian and Lo, Kyle and Dodge, Jesse and Sap, Maarten and Koh, Pang Wei and Wang, Chun and Hajishirzi, Hannaneh and Smith, Noah A},
  journal={arXiv preprint arXiv:2509.11106},
  year={2025}
}

@misc{metr2026horizon,
    title = {Task-Completion Time Horizons of Frontier AI Models},
    author = {METR},
    howpublished = {\url{https://metr.org/time-horizons/}},
    year = {2026},
    month = {03},
}

@inproceedings{
vendrow2024large,
title={Large Language Model Benchmarks Do Not Test Reliability},
author={Joshua Vendrow and Edward Vendrow and Sara Beery and Aleksander Madry},
booktitle={Neurips Safe Generative AI Workshop 2024},
year={2024},
url={https://openreview.net/forum?id=XSeN6xZtZ9}
}

@misc{wu2026efficientevaluationllmperformance,
      title={Efficient Evaluation of LLM Performance with Statistical Guarantees}, 
      author={Skyler Wu and Yash Nair and Emmanuel J. Candès},
      year={2026},
      eprint={2601.20251},
      archivePrefix={arXiv},
      primaryClass={stat.ML},
      url={https://arxiv.org/abs/2601.20251}, 
}

@inproceedings{
truong2025reliable,
title={Reliable and Efficient Amortized Model-based Evaluation},
author={Sang T. Truong and Yuheng Tu and Percy Liang and Bo Li and Sanmi Koyejo},
booktitle={Forty-second International Conference on Machine Learning},
year={2025},
url={https://openreview.net/forum?id=HDbWrsgkB9}
}

@InProceedings{arief2021deepprob,
  title = 	 {Deep Probabilistic Accelerated Evaluation: A Robust Certifiable Rare-Event Simulation Methodology for Black-Box Safety-Critical Systems},
  author =       {Arief, Mansur and Huang, Zhiyuan and Koushik Senthil Kumar, Guru and Bai, Yuanlu and He, Shengyi and Ding, Wenhao and Lam, Henry and Zhao, Ding},
  booktitle = 	 {Proceedings of The 24th International Conference on Artificial Intelligence and Statistics},
  pages = 	 {595--603},
  year = 	 {2021},
  editor = 	 {Banerjee, Arindam and Fukumizu, Kenji},
  volume = 	 {130},
  series = 	 {Proceedings of Machine Learning Research},
  month = 	 {13--15 Apr},
  publisher =    {PMLR},
  url = 	 {https://proceedings.mlr.press/v130/arief21a.html},
}

@article{bai2022randomforest,
author = {Bai, Yuanlu and Huang, Zhiyuan and Lam, Henry and Zhao, Ding},
title = {Rare-event Simulation for Neural Network and Random Forest Predictors},
year = {2022},
issue_date = {July 2022},
publisher = {Association for Computing Machinery},
address = {New York, NY, USA},
volume = {32},
number = {3},
issn = {1049-3301},
url = {https://doi.org/10.1145/3519385},
doi = {10.1145/3519385},
journal = {ACM Trans. Model. Comput. Simul.},
month = jul,
articleno = {18},
numpages = {33}
}

@misc{maslej2025artificialintelligenceindexreport,
      title={Artificial Intelligence Index Report 2025}, 
      author={Nestor Maslej and Loredana Fattorini and Raymond Perrault and Yolanda Gil and Vanessa Parli and Njenga Kariuki and Emily Capstick and Anka Reuel and Erik Brynjolfsson and John Etchemendy and Katrina Ligett and Terah Lyons and James Manyika and Juan Carlos Niebles and Yoav Shoham and Russell Wald and Toby Walsh and Armin Hamrah and Lapo Santarlasci and Julia Betts Lotufo and Alexandra Rome and Andrew Shi and Sukrut Oak},
      year={2025},
      eprint={2504.07139},
      archivePrefix={arXiv},
      primaryClass={cs.AI},
      url={https://arxiv.org/abs/2504.07139}, 
}

@misc{comanici2025gemini25pushingfrontier,
      title={Gemini 2.5: Pushing the Frontier with Advanced Reasoning, Multimodality, Long Context, and Next Generation Agentic Capabilities}, 
      author={Gheorghe Comanici and Eric Bieber and Mike Schaekermann and Ice Pasupat and Noveen Sachdeva and Inderjit Dhillon and Marcel Blistein and others},
      year={2025},
      eprint={2507.06261},
      archivePrefix={arXiv},
      primaryClass={cs.CL},
      url={https://arxiv.org/abs/2507.06261}, 
}

@inproceedings{
    jimenez2024swebench,
    title={{SWE}-bench: Can Language Models Resolve Real-world Github Issues?},
    author={Carlos E Jimenez and John Yang and Alexander Wettig and Shunyu Yao and Kexin Pei and Ofir Press and Karthik R Narasimhan},
    booktitle={The Twelfth International Conference on Learning Representations},
    year={2024},
    url={https://openreview.net/forum?id=VTF8yNQM66}
}

@misc{merrill2026terminalbenchbenchmarkingagentshard,
      title={Terminal-Bench: Benchmarking Agents on Hard, Realistic Tasks in Command Line Interfaces}, 
      author={Mike A. Merrill and Alexander G. Shaw and Nicholas Carlini and Boxuan Li and Harsh Raj and Ivan Bercovich and Lin Shi and Jeong Yeon Shin and others},
      year={2026},
      eprint={2601.11868},
      archivePrefix={arXiv},
      primaryClass={cs.SE},
      url={https://arxiv.org/abs/2601.11868}, 
}

@inproceedings{yue2023mmmu,
        title={MMMU: A Massive Multi-discipline Multimodal Understanding and Reasoning Benchmark for Expert AGI},
        author={Xiang Yue and Yuansheng Ni and Kai Zhang and Tianyu Zheng and Ruoqi Liu and Ge Zhang and Samuel Stevens and Dongfu Jiang and Weiming Ren and Yuxuan Sun and Cong Wei and Botao Yu and Ruibin Yuan and Renliang Sun and Ming Yin and Boyuan Zheng and Zhenzhu Yang and Yibo Liu and Wenhao Huang and Huan Sun and Yu Su and Wenhu Chen},
        booktitle={Proceedings of CVPR},
        year={2024},
      }

@inproceedings{yue2025mmmu-pro,
  title={MMMU-Pro: A More Robust Multi-discipline Multimodal Understanding Benchmark},
  author={Xiang Yue and Tianyu Zheng and Yuansheng Ni and Yubo Wang and Kai Zhang and Shengbang Tong and Yuxuan Sun and Botao Yu and Ge Zhang and Huan Sun and Yu Su and Wenhu Chen and Graham Neubig},
  booktitle={Proceedings of ACL},
  year={2025}
}

@article{phan2025lastexam,
      title = {A benchmark of expert-level academic questions to assess {AI} capabilities},
      author = {{Center for AI Safety} and {Scale AI} and {HLE Contributors Consortium}},
      journal = {Nature},
      volume = {649},
      pages = {1139--1146},
      year = {2026},
      doi = {10.1038/s41586-025-09962-4},
      eprint = {2501.14249},
      archivePrefix = {arXiv},
      primaryClass = {cs.LG},
      url = {https://arxiv.org/abs/2501.14249}
}

@misc{foundation2026arcagi3newchallengefrontier,
      title={ARC-AGI-3: A New Challenge for Frontier Agentic Intelligence}, 
      author={ARC Prize Foundation},
      year={2026},
      eprint={2603.24621},
      archivePrefix={arXiv},
      primaryClass={cs.AI},
      url={https://arxiv.org/abs/2603.24621}, 
}

@inproceedings{zellers2019hellaswag,
    title={HellaSwag: Can a Machine Really Finish Your Sentence?},
    author={Zellers, Rowan and Holtzman, Ari and Bisk, Yonatan and Farhadi, Ali and Choi, Yejin},
    booktitle ={Proceedings of the 57th Annual Meeting of the Association for Computational Linguistics},
    year={2019}
}

@inproceedings{wang-etal-2018-glue,
    title = "{GLUE}: A Multi-Task Benchmark and Analysis Platform for Natural Language Understanding",
    author = "Wang, Alex  and
      Singh, Amanpreet  and
      Michael, Julian  and
      Hill, Felix  and
      Levy, Omer  and
      Bowman, Samuel",
    editor = "Linzen, Tal  and
      Chrupa{\l}a, Grzegorz  and
      Alishahi, Afra",
    booktitle = "Proceedings of the 2018 {EMNLP} Workshop {B}lackbox{NLP}: Analyzing and Interpreting Neural Networks for {NLP}",
    month = nov,
    year = "2018",
    address = "Brussels, Belgium",
    publisher = "Association for Computational Linguistics",
    url = "https://aclanthology.org/W18-5446/",
    doi = "10.18653/v1/W18-5446",
    pages = "353--355",
}

@article{sakaguchi2019winogrande,
    title={WinoGrande: An Adversarial Winograd Schema Challenge at Scale},
    author={Sakaguchi, Keisuke and Bras, Ronan Le and Bhagavatula, Chandra and Choi, Yejin},
    journal={arXiv preprint arXiv:1907.10641},
    year={2019}
}
\bibliographystyle{colm2026_conference}

\newpage
\appendix
\section{LLM Inference Setup}
\label{appendix:implementation_details}
We describe the implementation details for LLM inference.
We use the same system prompt for all models:
\[
\texttt{Please reason step by step, and put your final answer within $\backslash\backslash$boxed\{\}.}
\]
and parse the final answer within the box.
We follow the recommended decoding hyperparameters specified in each model card: temperature of 0.7 and top-p of 0.8 for Qwen2.5-Math-7B-Instruct, 1.0 and 1.0 for gpt-oss-20b-low, and 1.0 and 0.95 for Gemini 2.5 Flash Lite. For all models, we use a maximum of $32768$ new tokens. If not specified, we use the default values.

\section{Algorithm for calculating confidence bounds}
\begin{algorithm}[ht]
\caption{\textsc{ComputeEstimate}: Importance Sampling with Confidence Bounds}
\begin{algorithmic}[1]
\STATE \textbf{Input:} Failure indicator $f$, distributions $P$ and $Q$, sample size $N$, confidence level $\alpha$
\STATE \textbf{Output:} Estimate $\hat{\mu}$, confidence interval $[\hat{\mu}_{\text{low}}, \hat{\mu}_{\text{high}}]$
\STATE Sample $Z_1,\ldots,Z_N \sim Q$
\STATE Compute importance-weighted estimate: $\hat{\mu} \leftarrow \frac{1}{N}\sum_{i=1}^N f(Z_i) \frac{P(Z_i)}{Q(Z_i)}$
\STATE Compute sample variance: $\hat{\sigma}^2 \leftarrow \frac{1}{N-1}\sum_{i=1}^N \left(f(Z_i)\frac{P(Z_i)}{Q(Z_i)} - \hat{\mu}\right)^2$
\STATE Set $Z_{\alpha/2} \leftarrow$ critical value from standard normal for confidence level $\alpha$
\STATE Compute CI: $[\hat{\mu}_{\text{low}}, \hat{\mu}_{\text{high}}] \leftarrow \left[\hat{\mu} - Z_{\alpha/2}\frac{\hat{\sigma}}{\sqrt{N}}, \hat{\mu} + Z_{\alpha/2}\frac{\hat{\sigma}}{\sqrt{N}}\right]$
\STATE \textbf{Return} $\hat{\mu}$, $[\hat{\mu}_{\text{low}}, \hat{\mu}_{\text{high}}]$
\end{algorithmic}
\label{algorithm:compute_estimate}
\end{algorithm}

Algorithm~\ref{algorithm:compute_estimate} details the \textsc{ComputeEstimate} subroutine called in Algorithm~\ref{algorithm} (Section~\ref{sec:method}).

\section{Model-Template Pairs with Too Few Failures}
\label{appendix:too_rare}
The following model-template pairs have zero observed failures within our $100$K evaluation budget to support reliable estimation (failure counts from the full $100$K cache):
\begin{itemize}
    \item gpt-oss-20b-low: Templates 1, 3, 5, 6 ($K=8$)
    \item Gemini 2.5 Flash Lite: Templates 1, 3, 5, 7, 8 ($K=8$)
\end{itemize}
We expect systematic failure patterns may emerge with a larger evaluation budget, and leave this for future work.

\newpage
\section{Stronger Failure Concentration with Larger $K$}

To supplement Table~\ref{tab:tv_table}, we compare TV distance for each parameter dimension with increasing $K$.
The distance is measured between the normalized failure distribution per parameter dimension and the uniform baseline; we report the three parameter dimensions with the largest distances.
We observe that increasing $K$ leads to larger distances across models and templates.
This implies that with larger $K$, the failure distribution has shorter tails (fewer inconsistent failures) and probability mass concentrates more on the dominant failure-inducing inputs.
This empirically supports the efficiency gains that grow with $K$ in Table~\ref{tab:gain_error}.

\begin{table}[h]
\centering
\scalebox{0.95}{
\begin{tabular}{c|ccc|ccc|ccc}
\toprule
\multicolumn{10}{c}{\textbf{Qwen2.5-Math-7B-Instruct}}\\
\midrule
& \multicolumn{3}{c|}{$K = 4$} & \multicolumn{3}{c|}{$K = 8$} & \multicolumn{3}{c}{$K = 16$}\\
\midrule
Template & Top1 & Top2 & Top3 & Top1 & Top2 & Top3 & Top1 & Top2 & Top3\\
\midrule
0 & 0.822 & 0.705 & 0.400 & 0.884 & 0.749 & 0.433 & 0.899 & 0.749 & 0.509\\
1 & 0.629 & 0.485 & 0.477 & 0.857 & 0.713 & 0.710 & 0.914 & 0.833 & 0.750 \\
2 & 0.424 & 0.213 & 0.069 & 0.495 & 0.234 & 0.022 & 0.538 & 0.249 & 0.019 \\
3 & 0.230 & 0.218 & 0.184 & 0.364 & 0.296 & 0.219 & 0.472 & 0.362 & 0.310 \\
4 & 0.324 & 0.173 & 0.126 & 0.347 & 0.229 & 0.154 & 0.339 & 0.264 & 0.185 \\
5 & 0.752 & 0.446 & 0.266 & 0.799 & 0.599 & 0.559 & 0.939 & 0.866 & 0.799\\
6 & 0.835 & 0.623 & 0.333 & 0.862 & 0.708 & 0.269 & 0.878 & 0.752 & 0.246\\
7 & 0.682 & 0.588 & 0.380 & 0.773 & 0.696 & 0.528 & 0.797 & 0.748 & 0.556\\
8 & 0.850 & 0.750 & 0.524 & --- & --- & --- & --- & --- & --- \\
\midrule
\multicolumn{10}{c}{\textbf{gpt-oss-20b-low}}\\
\midrule
& \multicolumn{3}{c|}{$K = 8$} & \multicolumn{3}{c|}{$K = 16$} & \multicolumn{3}{c}{$K = 24$}\\
\midrule
Template & Top1 & Top2 & Top3 & Top1 & Top2 & Top3 & Top1 & Top2 & Top3\\
\midrule
0 & 0.708 & 0.621 & 0.394 & 0.900 & 0.800 & 0.695 & 0.933 & 0.850 & 0.729 \\
2 & 0.550 & 0.390 & 0.041 & 0.648 & 0.459 & 0.053 & 0.699 & 0.541 & 0.052 \\
4 & 0.419 & 0.292 & 0.223 & 0.552 & 0.396 & 0.335 & 0.631 & 0.508 & 0.385 \\
7 & 0.694 & 0.412 & 0.390 & 0.836 & 0.555 & 0.468 & 0.895 & 0.633 & 0.537 \\
8 & 0.680 & 0.541 & 0.300 & 0.732 & 0.590 & 0.413 & 0.760 & 0.599 & 0.487 \\
\midrule
\multicolumn{10}{c}{\textbf{Gemini 2.5 Flash Lite}}\\
\midrule
& \multicolumn{3}{c|}{$K = 8$} & \multicolumn{3}{c|}{$K = 16$} & \multicolumn{3}{c}{$K = 24$}\\
\midrule
Template & Top1 & Top2 & Top3 & Top1 & Top2 & Top3 & Top1 & Top2 & Top3\\
\midrule
0 & 0.739 & 0.664 & 0.118 & 0.829 & 0.792 & 0.237 & 0.867 & 0.850 & 0.267 \\
2 & 0.696 & 0.691 & 0.032 & 0.848 & 0.750 & 0.062 & 0.907 & 0.750 & 0.080 \\
4 & 0.201 & 0.093 & 0.086 & 0.224 & 0.110 & 0.107 & 0.208 & 0.133 & 0.102 \\
6 & 0.747 & 0.701 & 0.118 & 0.969 & 0.875 & 0.143 & 0.983 & 0.889 & 0.189 \\
\bottomrule
\end{tabular}}
\caption{Top-3 largest TV distances between the failure distribution and the uniform baseline, reported across models, templates, and majority-vote ensemble sizes $K$.
--- denotes model-templat pairs that have zero failures.
}
\label{appendix:tv_table}
\end{table}

\newpage
\section{Failure Histograms for All Parameters}
\begin{figure}[ht]
\begin{minipage}[t]{\linewidth}
    \centering
    \begin{subfigure}[b]{0.32\linewidth}
        \includegraphics[width=\linewidth]{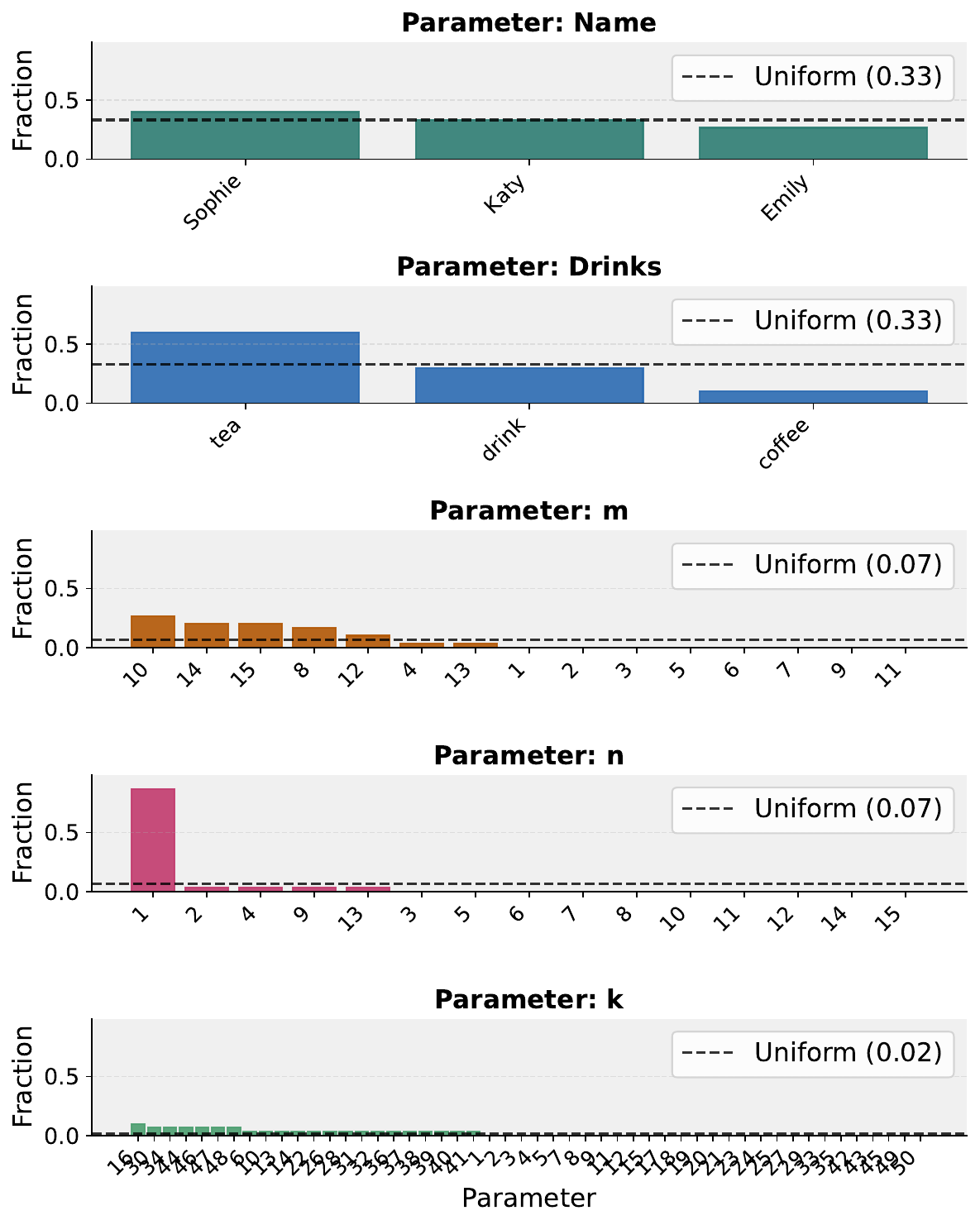}
    \end{subfigure}
    \hfill
    \begin{subfigure}[b]{0.32\linewidth}
        \vspace{-0.3cm}
        \includegraphics[width=\linewidth]{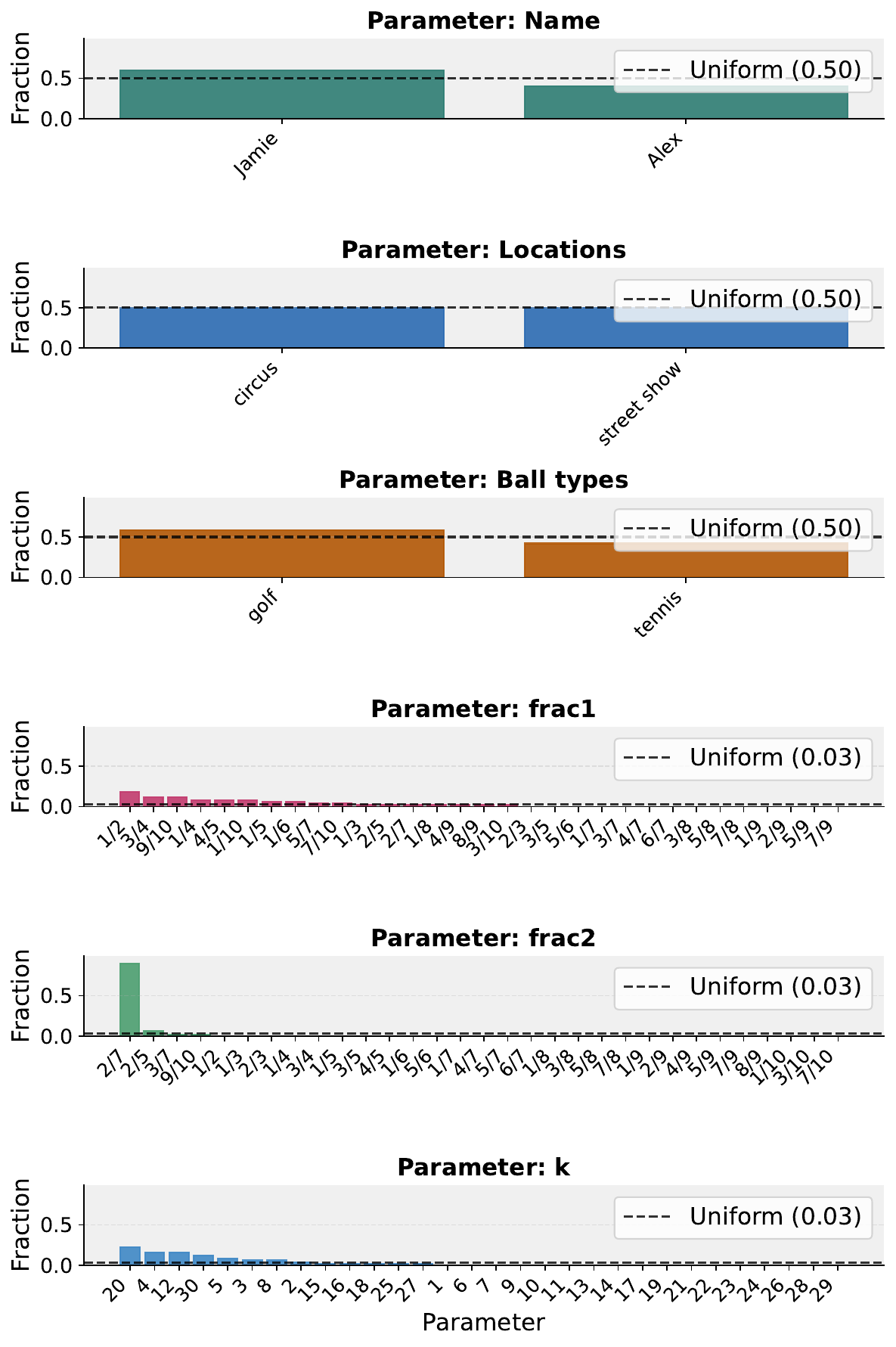}
    \end{subfigure}
    \begin{subfigure}[b]{0.32\linewidth}
        \vspace{-0.3cm}
        \includegraphics[width=\linewidth]{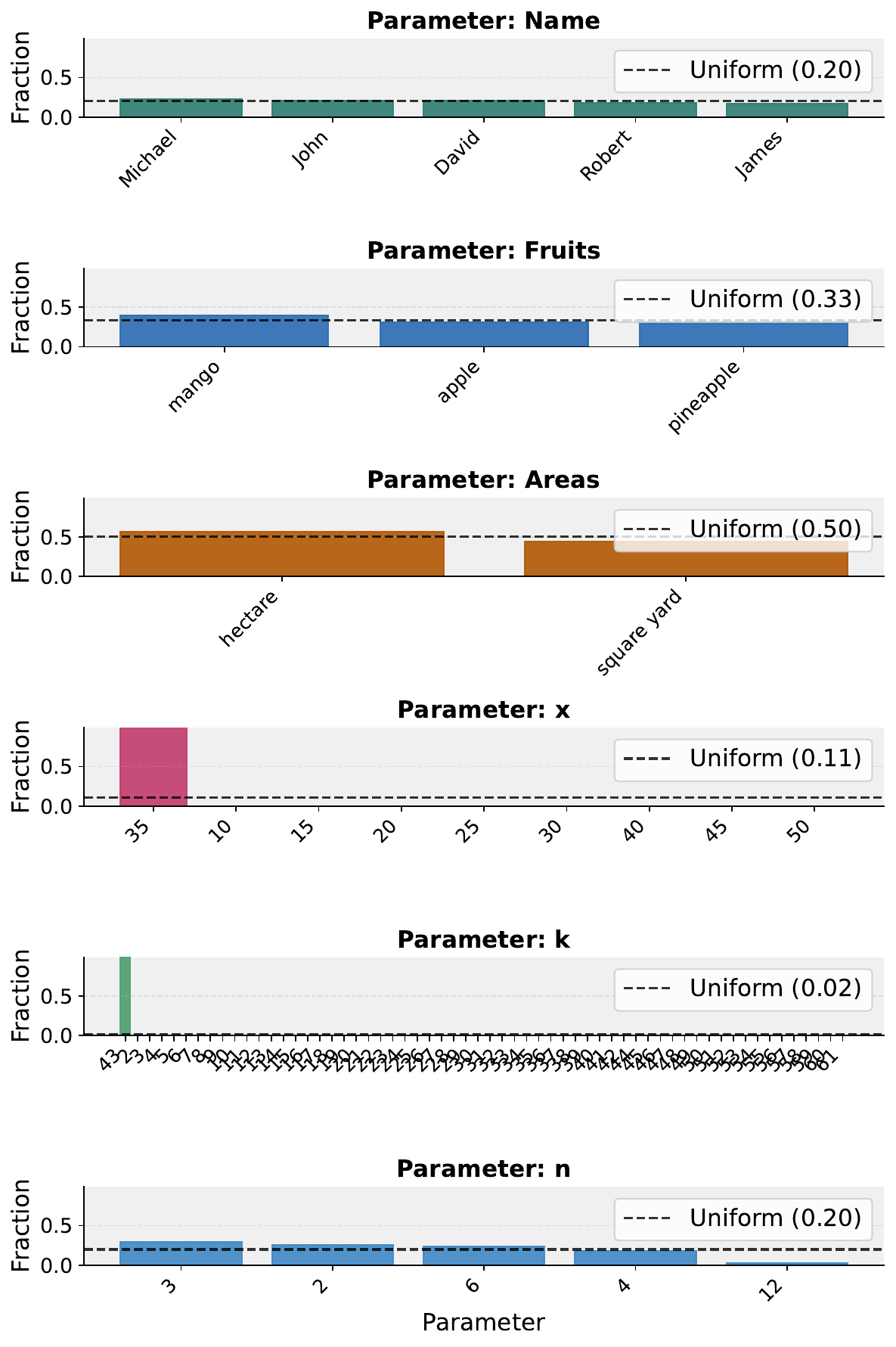}
    \end{subfigure}
    \caption{Parameter-wise failure histograms reveal strong concentration of errors.}
    \label{fig:full_param_hist}
\end{minipage}
\end{figure}

Figure~\ref{fig:full_param_hist} shows parameter-wise failure histograms for all parameters, supplementing Figure~\ref{fig:param_hist} in Section~\ref{sec:observation}.
Parameters such as \texttt{Name}, \texttt{Drinks}, and \texttt{Locations} are roughly uniformly distributed among failures.
In contrast, parameters such as \texttt{n} (first column), \texttt{frac2} (second column), and \texttt{x} (third column) show strong concentration of failures at a few specific values.

\newpage
\section{Table of Failure Probability Evaluation}
This section provides the full table of estimated failure probabilities across models, templates, and $K$ ($\lambda=0.1$), corresponding to Figure~\ref{fig:eval}.

\begin{table}[h]
\centering
\scalebox{0.95}{
\begin{tabular}{c|c|c|c}
\toprule
ID & Estimate ($\hat{\mu}_Q$) $\pm$ Bounds & Estimate ($\hat{\mu}_Q$) $\pm$ Bounds & Estimate ($\hat{\mu}_Q$) $\pm$ Bounds\\
\midrule
\multicolumn{4}{c}{\textbf{Qwen2.5-Math-7B-Instruct}}\\
\midrule
& $K{=}4$ & $K{=}8$ & $K{=}16$\\
\midrule
0 & 1.50e-03 $\pm$ 4.90e-05 & 8.98e-04 $\pm$ 3.03e-05 & 6.14e-04 $\pm$ 1.50e-05 \\
1 & 1.79e-03 $\pm$ 9.20e-05 & 4.34e-04 $\pm$ 3.36e-05 & 2.16e-04 $\pm$ 1.79e-05 \\
2 & 1.25e-01 $\pm$ 6.97e-04 & 9.11e-02 $\pm$ 5.73e-04 & 7.21e-02 $\pm$ 4.98e-04 \\
3 & 4.36e-02 $\pm$ 4.87e-04 & 1.80e-02 $\pm$ 2.95e-04 & 9.29e-03 $\pm$ 1.94e-04 \\
4 & 7.79e-02 $\pm$ 6.22e-04 & 5.62e-02 $\pm$ 5.21e-04 & 4.47e-02 $\pm$ 4.61e-04 \\
5 & 2.95e-03 $\pm$ 9.20e-05 & 7.41e-04 $\pm$ 3.29e-05 & 2.03e-04 $\pm$ 1.62e-05 \\
6 & 4.40e-03 $\pm$ 5.02e-05 & 3.50e-03 $\pm$ 4.16e-05 & 3.08e-03 $\pm$ 3.48e-05 \\
7 & 2.28e-03 $\pm$ 6.37e-05 & 1.46e-03 $\pm$ 3.50e-05 & 1.14e-03 $\pm$ 2.13e-05 \\
8 & 1.96e-04 $\pm$ 2.36e-05 & 0.0, 5.30e-06 & 0.0, 5.30e-06 \\
\midrule
\multicolumn{4}{c}{\textbf{gpt-oss-20b-low}}\\
\midrule
ID & $K{=}8$ & $K{=}16$ & $K{=}24$\\
\midrule
0 & 5.22e-04 $\pm$ 3.40e-05 & 2.54e-04 $\pm$ 1.90e-05 & 1.70e-04 $\pm$ 1.19e-05 \\
2 & 4.87e-02 $\pm$ 3.55e-04 & 2.55e-02 $\pm$ 2.42e-04 & 1.89e-02 $\pm$ 1.98e-04 \\
4 & 1.09e-02 $\pm$ 2.24e-04 & 3.71e-03 $\pm$ 1.15e-04 & 2.29e-03 $\pm$ 7.96e-05 \\
7 & 1.42e-03 $\pm$ 6.18e-05 & 5.17e-04 $\pm$ 2.70e-05 & 3.42e-04 $\pm$ 1.92e-05 \\
8 & 3.49e-03 $\pm$ 7.65e-05 & 1.61e-03 $\pm$ 4.93e-05 & 1.12e-03 $\pm$ 3.86e-05 \\
\midrule
\multicolumn{4}{c}{\textbf{Gemini 2.5 Flash Lite}}\\
\midrule
ID & $K{=}4$ & $K{=}8$ & $K{=}16$\\
\midrule
0 & 2.74e-03 $\pm$ 1.21e-04 & 7.18e-04 $\pm$ 3.63e-05 & 3.28e-04 $\pm$ 1.65e-05 \\
2 & 9.53e-03 $\pm$ 1.66e-04 & 4.73e-03 $\pm$ 7.29e-05 & 3.35e-03 $\pm$ 4.06e-05 \\
4 & 5.54e-02 $\pm$ 5.69e-04 & 3.37e-02 $\pm$ 4.42e-04 & 2.59e-02 $\pm$ 3.87e-04 \\
6 & 8.70e-03 $\pm$ 2.22e-04 & 1.27e-03 $\pm$ 6.26e-05 & 9.53e-04 $\pm$ 1.37e-05 \\
\bottomrule
\end{tabular}
}
\caption{Estimated failure probability across templates, models, and $K$ at $\lambda = 0.1$.
For all missing model-template pairs are those that have zero failures within our evaluation budgets, and they can be estimated via exact binomial bounds of (0, 5.30e-06).
}
\label{appendix:eval}
\end{table}

\clearpage
\section{Parameterization of GSM8K Templates}
\label{appendix:templates}
Motivated by GSM-Symbolic~\citep{mirzadeh2025gsmsymbolic}, we construct parameterized GSM8K templates with extended variables and corrected grammatical flaws. Each template consists of a prompt with placeholders, a variable domain, and a formula determining the answer.

\footnotesize{
\textbf{Template 0}
\begin{itemize}
    \item Prompt:\\ \{$Z_{\text{Name}}$\} saw a \{$Z_{x}$\}-foot \{$Z_{\text{Fish}}$\} with \{$Z_{y}$\} remoras, each \{$Z_{z}$\}-inches long, attached to it. What percentage of the \{$Z_{\text{Fish}}$\}'s body length is the combined length of the remoras?
    \item Variables:\\
$Z_{\text{Name}} \in \{\text{Benny, Alex, Casey, Jordan, Taylor, Morgan, Riley, Sam, Jamie, Avery}\}$\\
$Z_{\text{Fish}} \in \{\text{dolphin, whale, shark}\}$\\
$Z_{k} \in \{2, \ldots, 21\}$\\
$Z_{y} \in \{1, \ldots, 30\}$\\
$Z_{\text{ratio}} \in \{4, 5, 10, 20, 25\}$
    \item Formula:\\
$Z_{y} = Z_{y} \times 12$\\
$Z_{x} = Z_{\text{ratio}} \times Z_{k} \times Z_{y} / 12$\\
$\text{answer}=100/\text{ratio}$
\end{itemize}
\hdashrule{0.96\linewidth}{0.5pt}{3pt 2pt}
\textbf{Template 1}
\begin{itemize}
    \item Prompt:\\ A \{$Z_{\text{weather}}$\} rolls in from the ocean to cover \{$Z_{\text{city}}$\}. It takes \{$Z_{t}$\} minutes to cover every \{$Z_{d}$\} miles of the city. If the city is \{$Z_{y}$\} miles across from the oceanfront to the opposite inland edge, how many minutes will it take for the \{$Z_{\text{weather}}$\} to cover the whole city?
    \item Variables:\\
$Z_{\text{city}} \in \{\text{Harborview, Lakewood, Riverton}\}$\\
$Z_{\text{weather}} \in \{\text{fog bank, marine layer, coastal mist}\}$\\
$Z_{t} \in \{2, \ldots, 71\}$\\
$Z_{d} \in \{2, \ldots, 13\}$\\
$Z_{s} \in \{1, \ldots, 12\}$
    \item Formula:\\
$Z_{y} = Z_{d} \times Z_{s}$\\
$\text{answer}=Z_s\times Z_t$
\end{itemize}
\hdashrule{0.96\linewidth}{0.5pt}{3pt 2pt}
\textbf{Template 2}
\begin{itemize}
    \item Prompt:\\ \{$Z_{\text{name}}$\} is playing a \{$Z_{\text{game}}$\} during \{$Z_{\text{setting}}$\} for a \{$Z_{\text{prize}}$\}. \{$Z_{\text{name}}$\} is rolling a \{$Z_{s}$\}-sided die. By how many percentage points is it more likely that they roll a number greater than \{$Z_{\text{target}}$\} than that they roll two \{$Z_{z}$\} numbers in a row?
    \item Variables:\\
$Z_{\text{name}} \in \{\text{Jerry, Alex, Casey, Jordan, Taylor, Morgan}\}$\\
$Z_{\text{setting}} \in \{\text{game night, casino night, board game meetup}\}$\\
$Z_{\text{game}} \in \{\text{dice game}\}$\\
$Z_{\text{prize}} \in \{\text{sticker, token, pin}\}$\\
$Z_{\text{target}} \in \{2, 4, 6, \ldots, 200\}$\\
$Z_{d} \in \{2, 4, 5, 10, 20, 25, 50, 100\}$\\
$Z_{z} \in \{\text{even, odd}\}$
    \item Formula:\\
$Z_{s} = Z_{\text{target}} \times Z_{d}$\\
$\text{answer} = \frac{Z_{d}-1}{Z_{d}} \cdot 100 - 25$
\end{itemize}
\hdashrule{0.96\linewidth}{0.5pt}{3pt 2pt}
\textbf{Template 3}
\begin{itemize}
    \item Prompt:\\ A \{$Z_{\text{group}}$\} of \{$Z_{n}$\} students has various hobbies. \{$Z_{n1}$\} like to \{$Z_{\text{hobby1}}$\}, \{$Z_{n2}$\} like to play \{$Z_{\text{sport}}$\}, and the rest like to either \{$Z_{\text{hobby3}}$\} or \{$Z_{\text{hobby4}}$\}. How many like to \{$Z_{\text{hobby3}}$\} if the number that like to \{$Z_{\text{hobby4}}$\} is \{$Z_{\text{multi}}$\} times the number that prefer playing \{$Z_{\text{sport}}$\}?
    \item Variables:\\
$Z_{\text{sport}} \in \{\text{basketball, soccer}\}$\\
$Z_{\text{hobby1}} \in \{\text{read, paint}\}$\\
$Z_{\text{hobby3}} \in \{\text{play music, garden}\}$\\
$Z_{\text{hobby4}} \in \{\text{write}\}$\\
$Z_{n1} \in \{2, \ldots, 11\}$\\
$Z_{n2} \in \{2, \ldots, 11\}$\\
$Z_{\text{extra}} \in \{1, \ldots, 10\}$\\
$Z_{\text{multi}} \in \{2, \ldots, 11\}$
    \item Formula:\\
$Z_{n} = Z_{n1} + (Z_{\text{multi}} + 1) \times Z_{n2} + Z_{\text{extra}}$\\
$\text{answer} = Z_{\text{extra}}$
\end{itemize}
\hdashrule{0.96\linewidth}{0.5pt}{3pt 2pt}
\textbf{Template 4}
\begin{itemize}
    \item Prompt:\\ \{$Z_{\text{name}}$\} is popping popcorn for a snack. As the \{$Z_{\text{p}}$\} of kernels heats up, the kernels start popping faster. \{$Z_{n}$\} pop in the first \{$Z_{s}$\} seconds of cooking, then \{$Z_{k2}$\} times that amount in the next \{$Z_{s}$\} seconds. The kernels increase to \{$Z_{k3}$\} times the initial popping rate in the next \{$Z_{s}$\} seconds, but in the final \{$Z_{s}$\} seconds, the popping slows down to \{$Z_{\text{r1}}$\} of the rate as the past \{$Z_{s}$\} seconds. After \{$Z_{\text{name}}$\} takes the \{$Z_{\text{p}}$\} off the heat, a \{$Z_{\text{r2}}$\} of the number of kernels that popped in the final \{$Z_{s}$\} seconds of cooking also pop from the residual heat. How many pieces of popcorn does \{$Z_{\text{name}}$\} have to eat?
    \item Variables:\\
$Z_{\text{name}} \in \{\text{Garrett, Alex, Casey}\}$\\
$Z_{\text{p}} \in \{\text{pan, pot, skillet}\}$\\
$Z_{\text{r1}} \in \{\text{half, quarter}\}$\\
$Z_{\text{r2}} \in \{\text{half, quarter}\}$\\
$Z_{n} \in \{16, 32, 48, \ldots, 160\}$\\
$Z_{s} \in \{2, 3, 4\}$\\
$Z_{k2} \in \{2, \ldots, 11\}$\\
$Z_{k3} \in \{12, \ldots, 21\}$
    \item Formula:\\
$\text{second} = Z_{k2} \cdot Z_n$\\
$\text{third} = Z_{k3} \cdot Z_n$\\
$\text{fourth} = \text{third} / Z_{\text{r1num}}$\\
$\text{residual} = \text{third} / (Z_{\text{r1num}} \cdot Z_{\text{r2num}})$\\
$\text{answer} = Z_n + \text{second} + \text{third} + \text{fourth} + \text{residual}$
\end{itemize}
\hdashrule{0.96\linewidth}{0.5pt}{3pt 2pt}
\textbf{Template 5}
\begin{itemize}
    \item Prompt:\\ \{$Z_{\text{name}}$\} makes \{$Z_{\text{drink}}$\} using teaspoons of sugar and cups of water in the ratio of \{$Z_{m}$\}:\{$Z_{n}$\}. If she used a total of \{$Z_{x}$\} teaspoons of sugar and cups of water, calculate the number of teaspoonfuls of sugar she used.
    \item Variables:\\
$Z_{\text{name}} \in \{\text{Katy, Sophie, Emily}\}$\\
$Z_{\text{drink}} \in \{\text{coffee, tea, drink}\}$\\
$Z_{m} \in \{1, \ldots, 15\}$\\
$Z_{n} \in \{1, \ldots, 15\}$\\
$Z_{k} \in \{1, \ldots, 50\}$
    \item Formula:\\
$Z_{\text{total}} = Z_m + Z_n$\\
$Z_x = Z_{\text{total}} \cdot Z_k$\\
$\text{answer} = \frac{Z_m \cdot Z_x}{Z_{\text{total}}}$
\end{itemize}
\hdashrule{0.96\linewidth}{0.5pt}{3pt 2pt}
\textbf{Template 6}
\begin{itemize}
    \item Prompt:\\ \{$Z_{\text{name}}$\} has \{$Z_{x}$\} \{$Z_{\text{area}}$\}s of \{$Z_{\text{fruit}}$\}s field. There are \{$Z_{k}$\} \{$Z_{\text{fruit}}$\}s per \{$Z_{\text{area}}$\}. \{$Z_{\text{name}}$\} can harvest his \{$Z_{\text{fruit}}$\}s every \{$Z_{n}$\} months. How many \{$Z_{\text{fruit}}$\}s can \{$Z_{\text{name}}$\} harvest within a year?
    \item Variables:\\
$Z_{\text{name}} \in \{\text{John, Michael, David, James, Robert}\}$\\
$Z_{\text{fruit}} \in \{\text{pineapple, apple, mango}\}$\\
$Z_{\text{area}} \in \{\text{hectare, square yard}\}$\\
$Z_{x} \in \{10, 15, 20, \ldots, 50\}$\\
$Z_{k} \in \{2, \ldots, 61\}$\\
$Z_{n} \in \{2, 3, 4, 6, 12\}$
    \item Formula:\\
$Z_{\text{cycles}} = 12 / Z_n$\\
$\text{answer} = Z_x \cdot Z_k \cdot Z_{\text{cycles}}$
\end{itemize}
\hdashrule{0.96\linewidth}{0.5pt}{3pt 2pt}
\textbf{Template 7}
\begin{itemize}
    \item Prompt:\\ \{$Z_{\text{name}}$\} is juggling at a \{$Z_{\text{location}}$\}. \{$Z_{\text{name}}$\} can juggle \{$Z_{n}$\} balls. \{$Z_{\text{frac1}}$\} of the balls are \{$Z_{\text{ball\_type}}$\} balls, and \{$Z_{\text{frac2}}$\} of the \{$Z_{\text{ball\_type}}$\} balls are \{$Z_{\text{color}}$\}. How many \{$Z_{\text{color}}$\} \{$Z_{\text{ball\_type}}$\} balls are there?
    \item Variables:\\
$Z_{\text{name}} \in \{\text{Jamie, Alex}\}$\\
$Z_{\text{location}} \in \{\text{circus, street show}\}$\\
$Z_{\text{ball\_type}} \in \{\text{golf, tennis}\}$\\
$Z_{\text{color}} \in \{\text{blue}\}$\\
$Z_{\text{frac1}}, Z_{\text{frac2}} \in \{1/2, 1/3, 2/3, 1/4, 3/4, 1/5, 2/5, 3/5, 4/5, 1/6, 5/6, 1/7, 2/7, 3/7, \\ 4/7, 5/7, 6/7, 1/8, 3/8, 5/8, 7/8, 1/9, 2/9, 4/9, 5/9, 7/9, 8/9, 1/10, 3/10, 7/10, 9/10\}$\\
$Z_k \in \{1, \ldots, 30\}$
    \item Formula:\\
$Z_n = Z_k \times \text{denom}(Z_{\text{frac1}}) \times \text{denom}(Z_{\text{frac2}})$\\
$\text{answer}=Z_n\times (\text{num}(Z_{\text{frac1}})\times \text{num}(Z_{\text{frac2}})) / (\text{denom}(Z_{\text{frac1}}) \times \text{denom}(Z_{\text{frac2}}))$
\end{itemize}
\hdashrule{0.96\linewidth}{0.5pt}{3pt 2pt}
\textbf{Template 8}
\begin{itemize}
    \item Prompt:\\ \{$Z_{\text{name}}$\} places \{$Z_{\text{obj}}$\}s on some \{$Z_{\text{surface}}$\}s. Each \{$Z_{\text{surface}}$\} can hold \{$Z_{x}$\} \{$Z_{\text{obj}}$\}s. If he has \{$Z_{n}$\} \{$Z_{\text{obj}}$\}s and \{$Z_{k}$\} \{$Z_{\text{surface}}$\}s, how many \{$Z_{\text{obj}}$\}s won't he be able to place on the \{$Z_{\text{surface}}$\}?
    \item Variables:\\
$Z_{\text{name}} \in \{\text{Jaime, John, Mike}\}$\\
$Z_{\text{obj}} \in \{\text{egg, olive, almond}\}$\\
$Z_{\text{surface}} \in \{\text{tray, plate, table}\}$\\
$Z_{x} \in \{2, \ldots, 21\}$\\
$Z_{k} \in \{2, \ldots, 21\}$\\
$Z_{\text{leftover}} \in \{1, \ldots, 10\}$
    \item Formula:\\
$Z_{n} = Z_{k} \cdot Z_{x} + Z_{\text{leftover}}$\\
$\text{answer}=Z_{\text{leftover}}$
\end{itemize}
}

\clearpage
\section{CI Width vs. Number of Inferences}
This section presents CI width versus number of inferences for all three models across varying templates and majority-vote sizes ($K$). Each figure corresponds to a specific model and $K$, with each column representing a different template. Within each subplot, the upper panel shows CI width vs. number of inferences, and the lower panel shows sorted CIs over $100$ independent sample sets at the evaluation size of the bottom-right Pareto-frontier dot. Points with coverage in the range $99\pm1\%$ are shown as solid markers, while points with coverage $\geq 95\%$ are shown as semi-transparent markers. Efficiency gains are computed only for results achieving $99\pm1\%$ coverage.

\begin{figure}[h]
    \centering
    \begin{subfigure}[b]{\linewidth}
        \includegraphics[width=\linewidth]{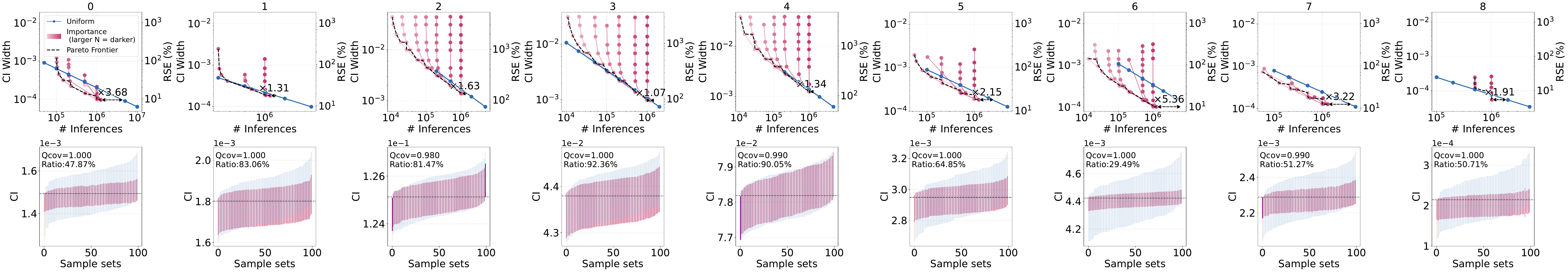}
    \end{subfigure}
    \caption{Qwen2.5-Math-7B-Instruct, $K=4$, $\lambda=0.1$}
    \label{fig:appendix_qwen_4_0.1}
\end{figure}

\begin{figure}[h]
    \centering
    \begin{subfigure}[b]{\linewidth}
        \includegraphics[width=\linewidth]{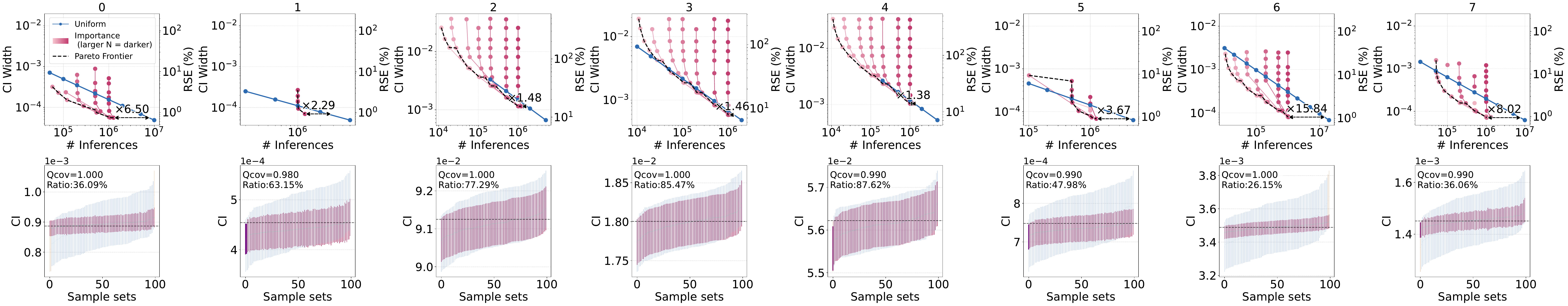}
    \end{subfigure}
    \caption{Qwen2.5-Math-7B-Instruct, $K=8$, $\lambda=0.1$}
    \label{fig:appendix_qwen_8_0.1}
\end{figure}

\begin{figure}[h]
    \centering
    \begin{subfigure}[b]{\linewidth}
        \includegraphics[width=\linewidth]{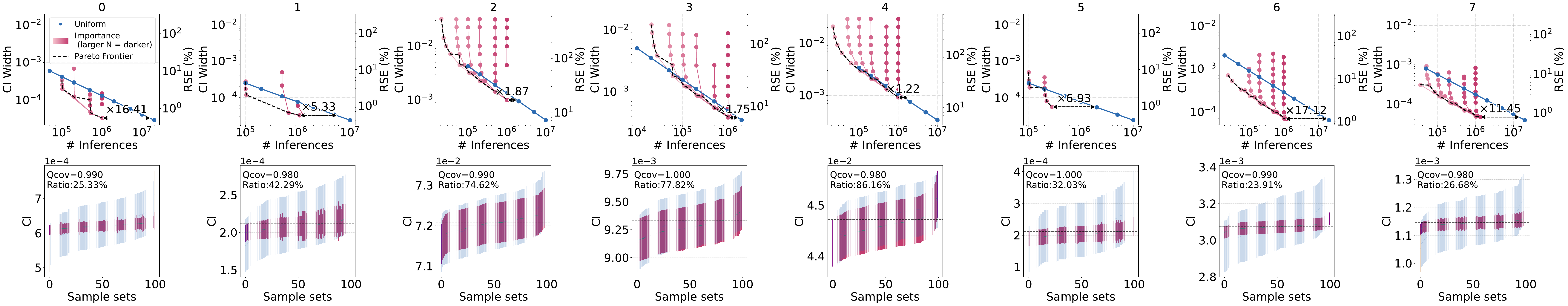}
    \end{subfigure}
    \caption{Qwen2.5-Math-7B-Instruct, $K=16$, $\lambda=0.1$}
    \label{fig:appendix_qwen_16_0.1}
\end{figure}

\begin{figure}[h]
    \centering
    \begin{subfigure}[b]{\linewidth}
        \includegraphics[width=\linewidth]{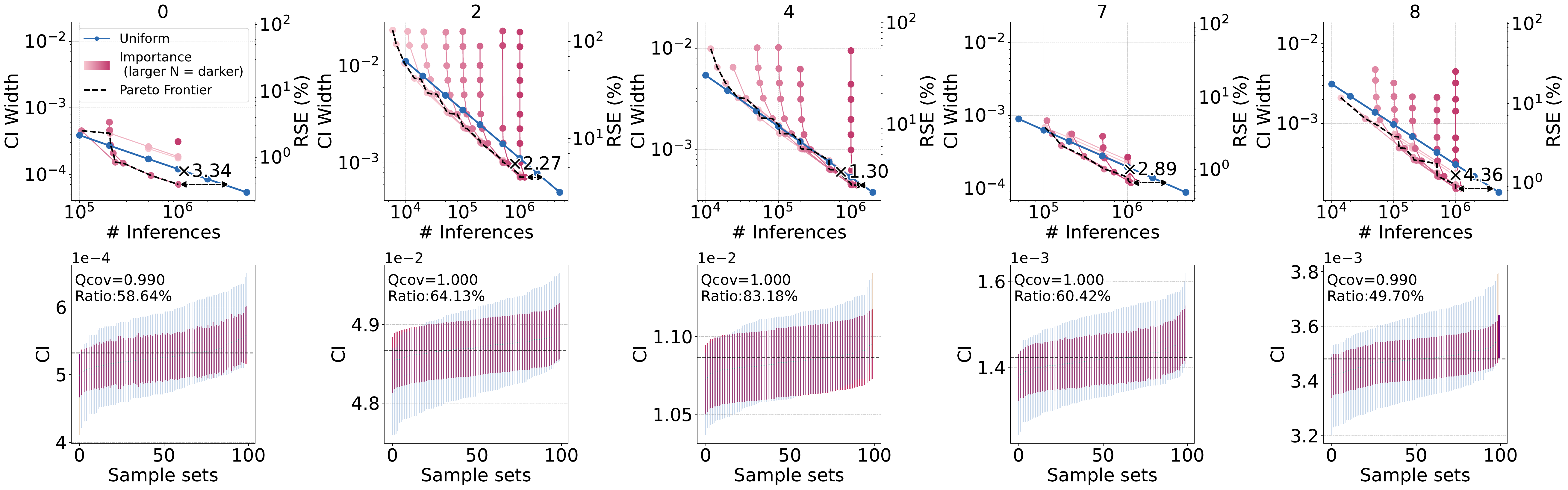}
    \end{subfigure}
    \caption{gpt-oss-20b-low, $K=8$, $\lambda=0.1$}
    \label{fig:appendix_low_8_0.1}
\end{figure}

\begin{figure}[h]
    \centering
    \begin{subfigure}[b]{\linewidth}
        \includegraphics[width=\linewidth]{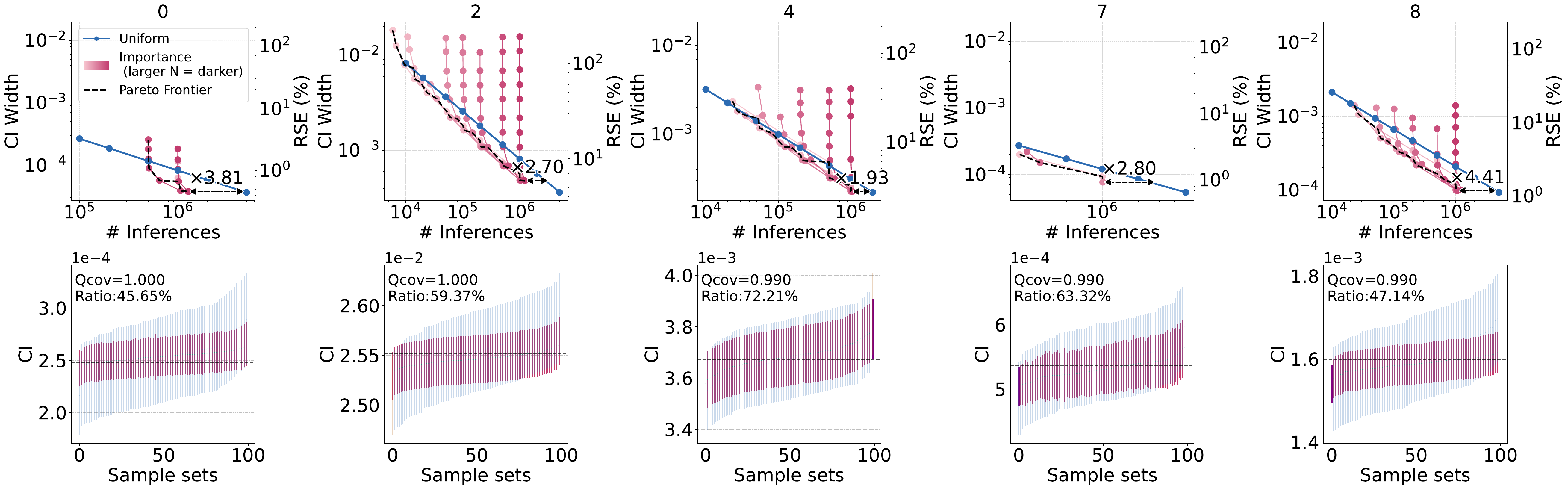}
    \end{subfigure}
    \caption{gpt-oss-20b-low, $K=16$, $\lambda=0.1$}
    \label{fig:appendix_low_16_0.1}
\end{figure}

\begin{figure}[h]
    \centering
    \begin{subfigure}[b]{\linewidth}
        \includegraphics[width=\linewidth]{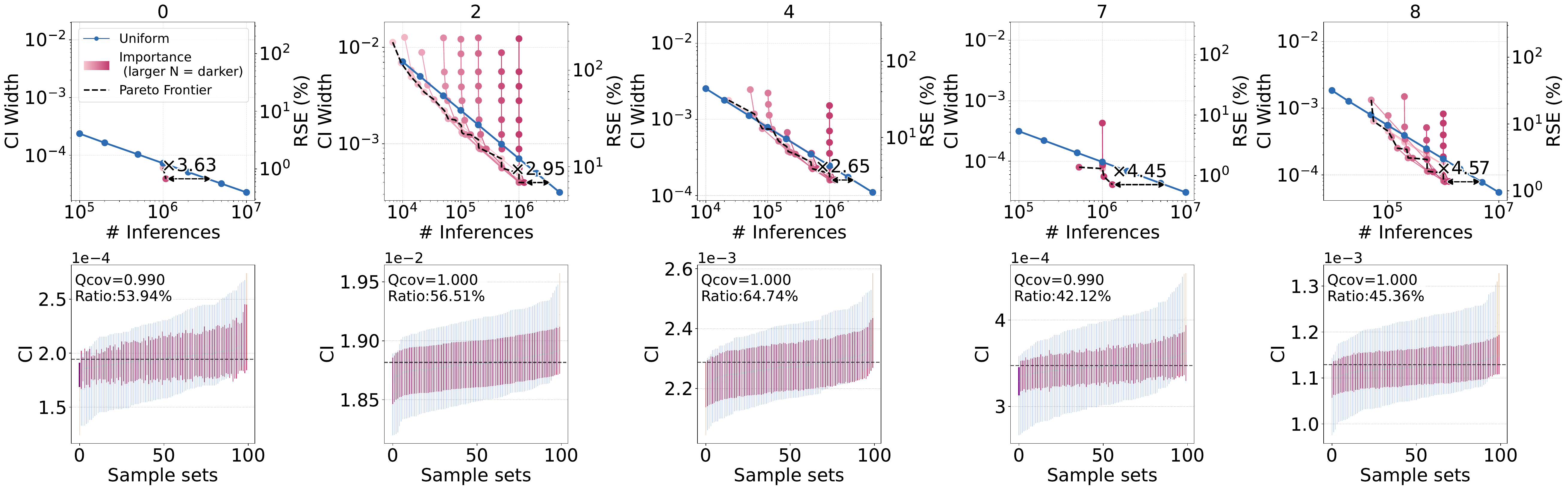}
    \end{subfigure}
    \caption{gpt-oss-20b-low, $K=24$, $\lambda=0.1$}
    \label{fig:appendix_low_24_0.1}
\end{figure}

\clearpage
\begin{figure}[h]
    \centering
    \begin{subfigure}[b]{0.79\linewidth}
        \includegraphics[width=\linewidth]{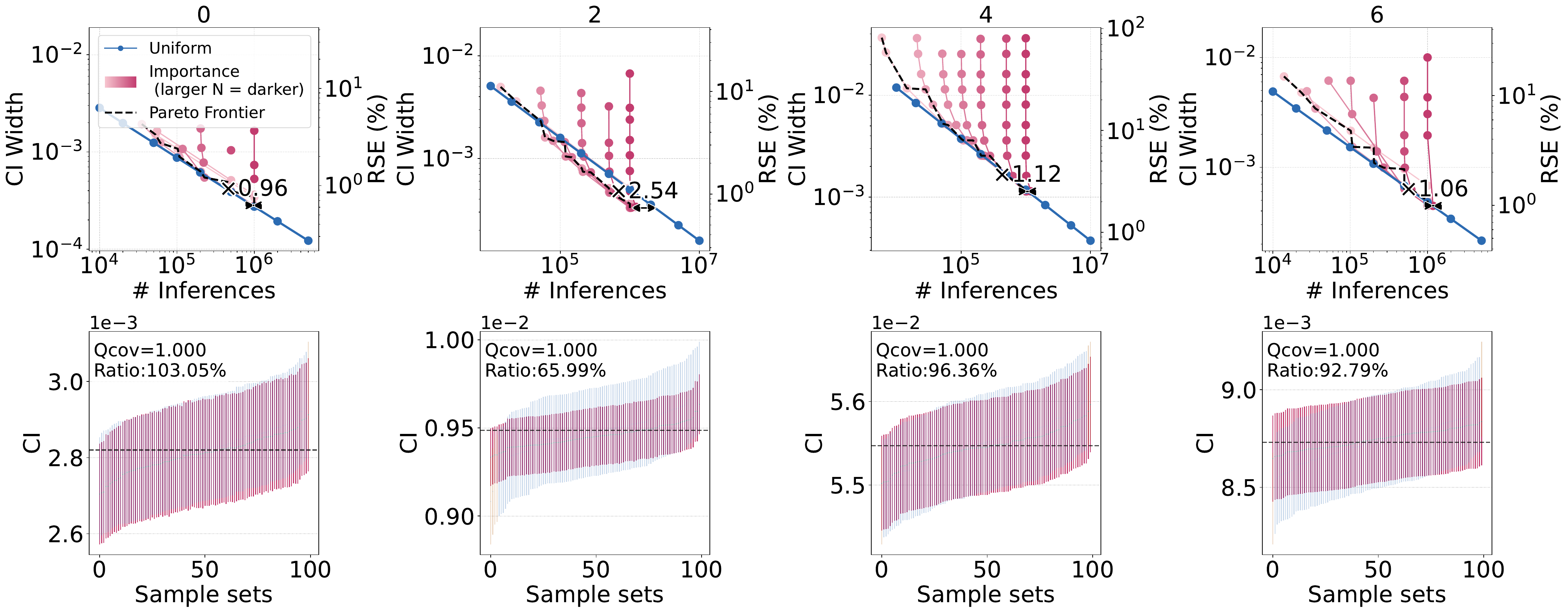}
    \end{subfigure}
    \caption{Gemini 2.5 Flash Lite, $K=4$, $\lambda=0.1$}
    \label{fig:appendix_gemini_4_0.1}
\end{figure}

\begin{figure}[h]
    \centering
    \begin{subfigure}[b]{0.79\linewidth}
        \includegraphics[width=\linewidth]{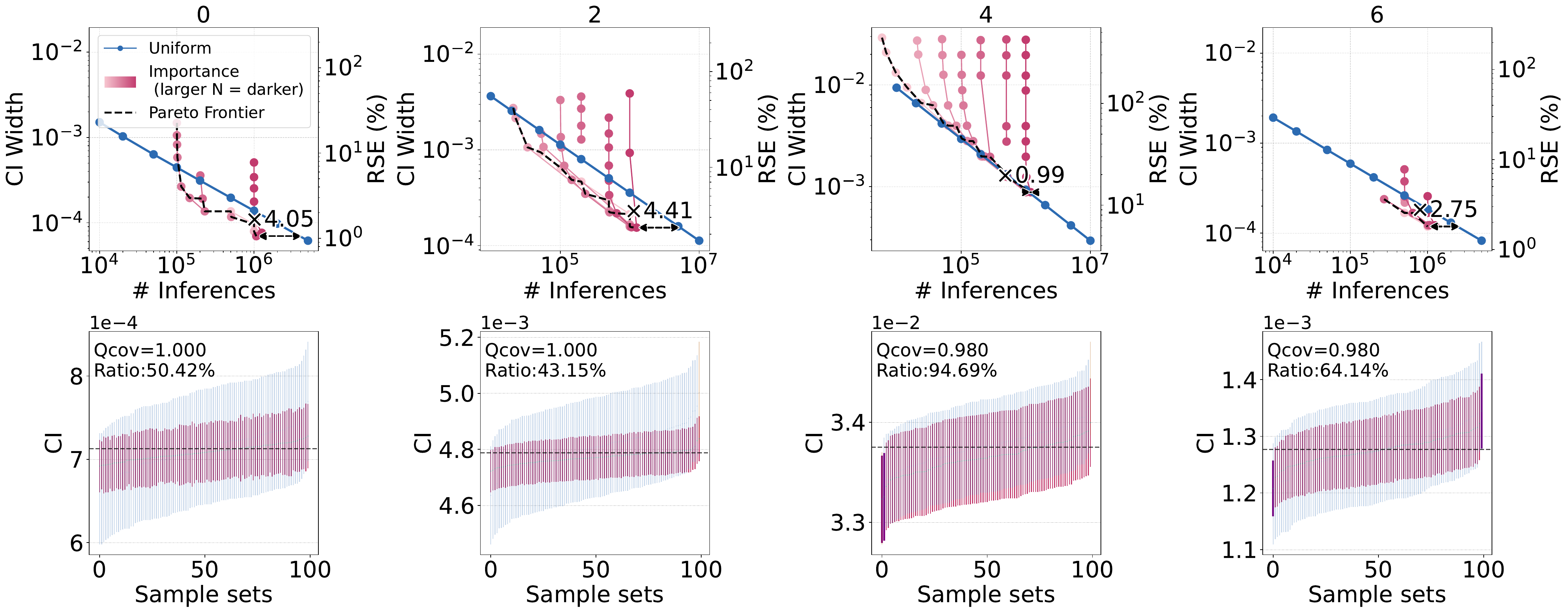}
    \end{subfigure}
    \caption{Gemini 2.5 Flash Lite, $K=8$, $\lambda=0.1$}
    \label{fig:appendix_gemini_8_0.1}
\end{figure}

\begin{figure}[h]
    \centering
    \begin{subfigure}[b]{0.79\linewidth}
        \includegraphics[width=\linewidth]{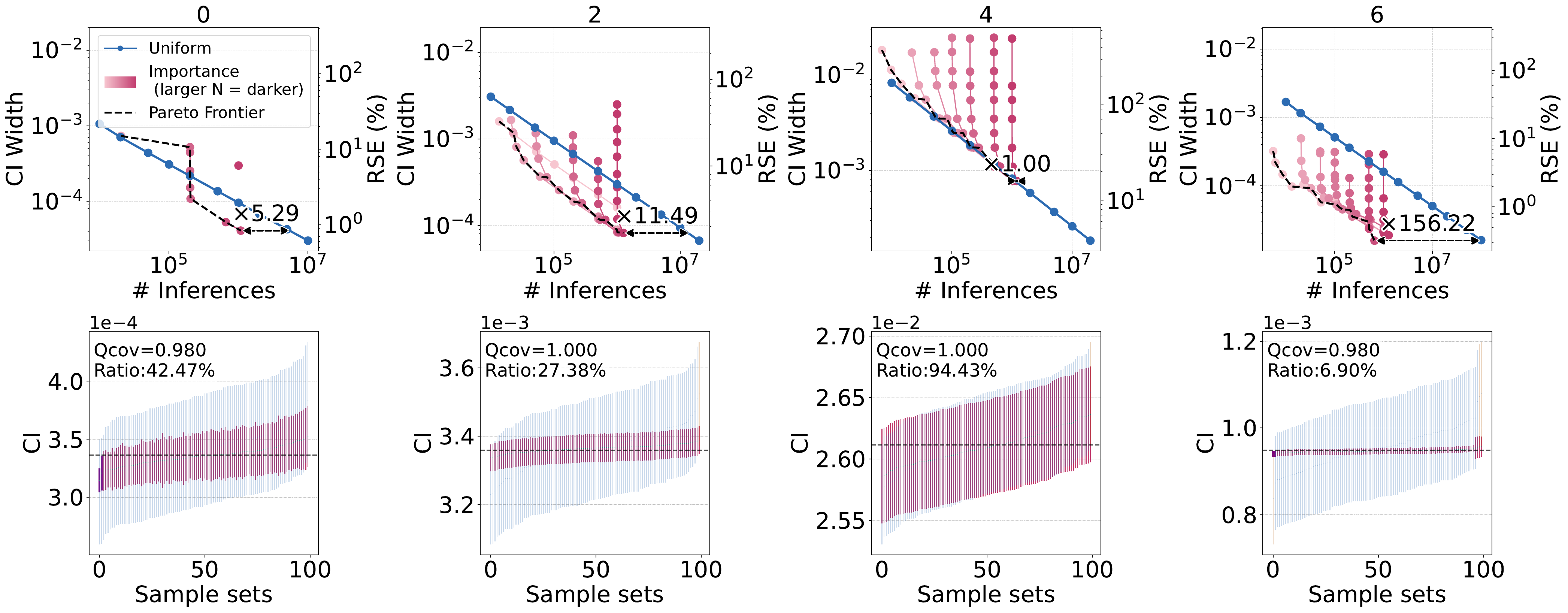}
    \end{subfigure}
    \caption{Gemini 2.5 Flash Lite, $K=16$, $\lambda=0.1$}
    \label{fig:appendix_gemini_16_0.1}
\end{figure}

\clearpage
\section{Additional examples for repetitive failures by LLMs}
\label{appendix:example_fails}
\begin{figure}[b]
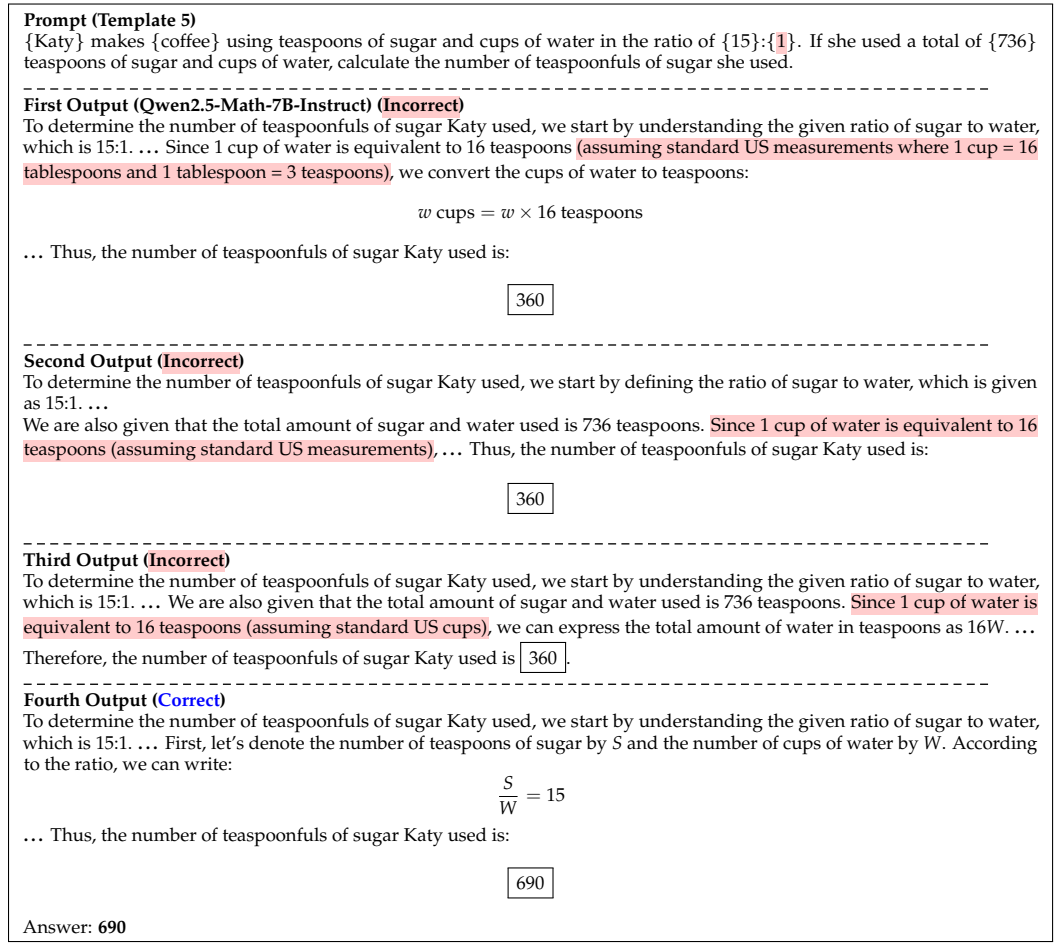

\begin{minipage}[t]{\linewidth}
\fbox{
\parbox{0.96\linewidth}{
\scriptsize
\textbf{Prompt (Template 5)}\\
\{Katy\} makes \{coffee\} using teaspoons of sugar and cups of water in the ratio of \{15\}:\{\hl{1}\}. If she used a total of \{736\} teaspoons of sugar and cups of water, calculate the number of teaspoonfuls of sugar she used.

\hdashrule{0.96\linewidth}{0.5pt}{3pt 2pt}\\
\textbf{First Output (Qwen2.5-Math-7B-Instruct) (\hl{Incorrect})}\\
To determine the number of teaspoonfuls of sugar Katy used, we start by understanding the given ratio of sugar to water, which is 15:1. 
$\textbf{\dots}$
Since 1 cup of water is equivalent to 16 teaspoons \hl{(assuming standard US measurements where 1 cup = 16 tablespoons and 1 tablespoon = 3 teaspoons)}, we convert the cups of water to teaspoons:
\[
w \text{ cups} = w \times 16 \text{ teaspoons}
\]
$\textbf{\dots}$
Thus, the number of teaspoonfuls of sugar Katy used is:
\[
\boxed{360}
\]
\hdashrule{0.96\linewidth}{0.5pt}{3pt 2pt}\\
\textbf{Second Output (\hl{Incorrect})}\\
To determine the number of teaspoonfuls of sugar Katy used, we start by defining the ratio of sugar to water, which is given as 15:1. 
$\textbf{\dots}$

We are also given that the total amount of sugar and water used is 736 teaspoons. \hl{Since 1 cup of water is equivalent to 16 teaspoons (assuming standard US measurements)}, 
$\textbf{\dots}$
Thus, the number of teaspoonfuls of sugar Katy used is:
\[
\boxed{360}
\]
\hdashrule{0.96\linewidth}{0.5pt}{3pt 2pt}\\
\textbf{Third Output (\hl{Incorrect})}\\
To determine the number of teaspoonfuls of sugar Katy used, we start by understanding the given ratio of sugar to water, which is 15:1. 
$\textbf{\dots}$
We are also given that the total amount of sugar and water used is 736 teaspoons. \hl{Since 1 cup of water is equivalent to 16 teaspoons (assuming standard US cups)}, we can express the total amount of water in teaspoons as \( 16W \). 
$\textbf{\dots}$
Therefore, the number of teaspoonfuls of sugar Katy used is \( \boxed{360} \).

\hdashrule{0.96\linewidth}{0.5pt}{3pt 2pt}\\
\textbf{Fourth Output (\textcolor{blue}{Correct})}\\
To determine the number of teaspoonfuls of sugar Katy used, we start by understanding the given ratio of sugar to water, which is 15:1.
$\textbf{\dots}$
First, let's denote the number of teaspoons of sugar by \( S \) and the number of cups of water by \( W \). According to the ratio, we can write:
\[
\frac{S}{W} = 15
\]
$\textbf{\dots}$
Thus, the number of teaspoonfuls of sugar Katy used is:
\[
\boxed{690}
\]
Answer: \textbf{690}
}
}
\end{minipage}
\caption{First four of eight independently generated outputs from Qwen2.5-Math-7B-Instruct evaluated on Template 5. Error type: unjustified unit conversion.}
\label{fig:appendix_example_fails_qwen5}
\end{figure}

\begin{figure}
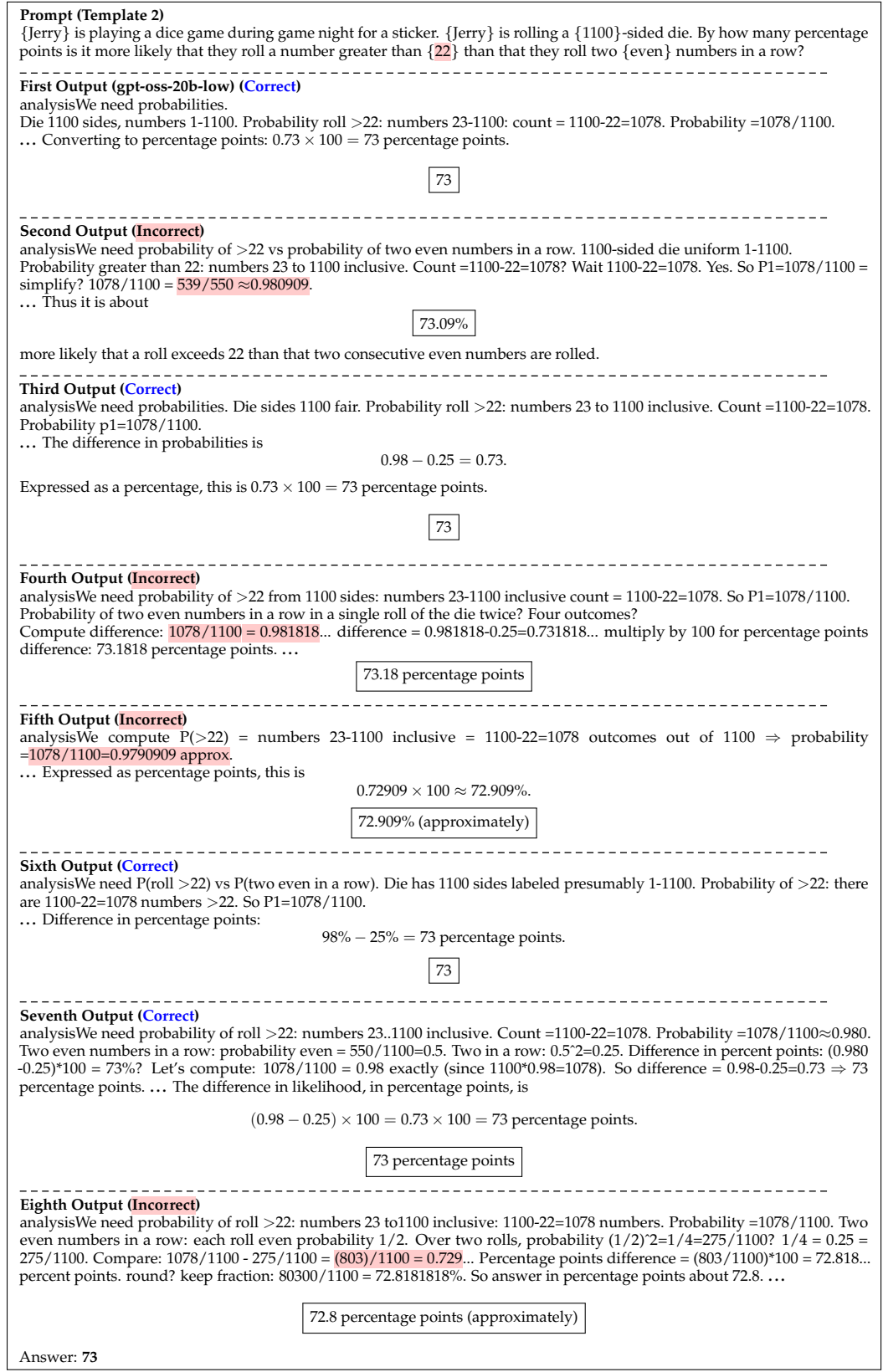

\begin{minipage}[t]{\linewidth}
\fbox{
\parbox{0.96\linewidth}{
\scriptsize
\textbf{Prompt (Template 2)}\\
\{Jerry\} is playing a dice game during game night for a sticker. \{Jerry\} is rolling a \{1100\}-sided die. By how many percentage points is it more likely that they roll a number greater than \{\hl{22}\} than that they roll two \{even\} numbers in a row?

\hdashrule{0.96\linewidth}{0.5pt}{3pt 2pt}\\
\textbf{First Output (gpt-oss-20b-low) (\textcolor{blue}{Correct})}\\
analysisWe need probabilities.

Die 1100 sides, numbers 1-1100. Probability roll $>$22: numbers 23-1100: count = 1100-22=1078. Probability =1078/1100.

$\textbf{\dots}$
Converting to percentage points: \(0.73 \times 100 = 73\) percentage points.
\[
\boxed{73}
\]

\hdashrule{0.96\linewidth}{0.5pt}{3pt 2pt}\\
\textbf{Second Output (\hl{Incorrect})}\\
analysisWe need probability of $>$22 vs probability of two even numbers in a row. 1100-sided die uniform 1-1100.

Probability greater than 22: numbers 23 to 1100 inclusive. Count =1100-22=1078? Wait 1100-22=1078. Yes. So P1=1078/1100 = simplify? 1078/1100 = \hl{539/550 $\approx$0.980909}.

$\textbf{\dots}$
Thus it is about
\[
\boxed{73.09\%}
\]
more likely that a roll exceeds 22 than that two consecutive even numbers are rolled.

\hdashrule{0.96\linewidth}{0.5pt}{3pt 2pt}\\
\textbf{Third Output (\textcolor{blue}{Correct})}\\
analysisWe need probabilities. Die sides 1100 fair. Probability roll $>$22: numbers 23 to 1100 inclusive. Count =1100-22=1078. Probability p1=1078/1100.

$\textbf{\dots}$
The difference in probabilities is
\[
0.98-0.25 = 0.73.
\]
Expressed as a percentage, this is \(0.73 \times 100 = 73\) percentage points.
\[
\boxed{73}
\]

\hdashrule{0.96\linewidth}{0.5pt}{3pt 2pt}\\
\textbf{Fourth Output (\hl{Incorrect})}\\
analysisWe need probability of $>$22 from 1100 sides: numbers 23-1100 inclusive count = 1100-22=1078. So P1=1078/1100.

Probability of two even numbers in a row in a single roll of the die twice? Four outcomes? 

Compute difference: \hl{1078/1100 = 0.981818}... difference = 0.981818-0.25=0.731818... multiply by 100 for percentage points difference: 73.1818 percentage points.
$\textbf{\dots}$
\[
\boxed{73.18\text{ percentage points}}
\]

\hdashrule{0.96\linewidth}{0.5pt}{3pt 2pt}\\
\textbf{Fifth Output (\hl{Incorrect})}\\
analysisWe compute P($>$22) = numbers 23-1100 inclusive = 1100-22=1078 outcomes out of 1100 $\Rightarrow$ probability =\hl{1078/1100=0.9790909 approx}.

$\textbf{\dots}$
Expressed as percentage points, this is
\[
0.72909 \times 100 \approx 72.909\%.
\]
\[
\boxed{72.909\%\text{ (approximately)}}
\]

\hdashrule{0.96\linewidth}{0.5pt}{3pt 2pt}\\
\textbf{Sixth Output (\textcolor{blue}{Correct})}\\
analysisWe need P(roll $>$22) vs P(two even in a row). Die has 1100 sides labeled presumably 1-1100. Probability of $>$22: there are 1100-22=1078 numbers $>$22. So P1=1078/1100.

$\textbf{\dots}$
Difference in percentage points:
\[
98\% - 25\% = 73\text{ percentage points.}
\]
\[
\boxed{73}
\]

\hdashrule{0.96\linewidth}{0.5pt}{3pt 2pt}\\
\textbf{Seventh Output (\textcolor{blue}{Correct})}\\
analysisWe need probability of roll $>$22: numbers 23..1100 inclusive. Count =1100-22=1078. Probability =1078/1100$\approx$0.980. Two even numbers in a row: probability even = 550/1100=0.5. Two in a row: 0.5\^{}2=0.25. Difference in percent points: (0.980 -0.25)*100 = 73\%? Let's compute: 1078/1100 = 0.98 exactly (since 1100*0.98=1078). So difference = 0.98-0.25=0.73 $\Rightarrow$ 73 percentage points.
$\textbf{\dots}$
The difference in likelihood, in percentage points, is
\[
(0.98-0.25)\times 100 = 0.73\times 100 = 73\text{ percentage points.}
\]
\[
\boxed{73\text{ percentage points}}
\]

\hdashrule{0.96\linewidth}{0.5pt}{3pt 2pt}\\
\textbf{Eighth Output (\hl{Incorrect})}\\
analysisWe need probability of roll $>$22: numbers 23 to1100 inclusive: 1100-22=1078 numbers. Probability =1078/1100. Two even numbers in a row: each roll even probability 1/2. Over two rolls, probability (1/2)\^{}2=1/4=275/1100? 1/4 = 0.25 = 275/1100. Compare: 1078/1100 - 275/1100 = \hl{(803)/1100 = 0.729}... Percentage points difference = (803/1100)*100 = 72.818... percent points. round? keep fraction: 80300/1100 = 72.8181818\%. So answer in percentage points about 72.8.
$\textbf{\dots}$
\[
\boxed{72.8\text{ percentage points (approximately)}}
\]
Answer: \textbf{73}
}
}
\end{minipage}
\caption{Eight independently generated outputs from gpt-oss-20b-low evaluated on Template 2. Error type: unnecessary approximation.}
\label{fig:appendix_example_fails_low2}
\end{figure}

\begin{figure}
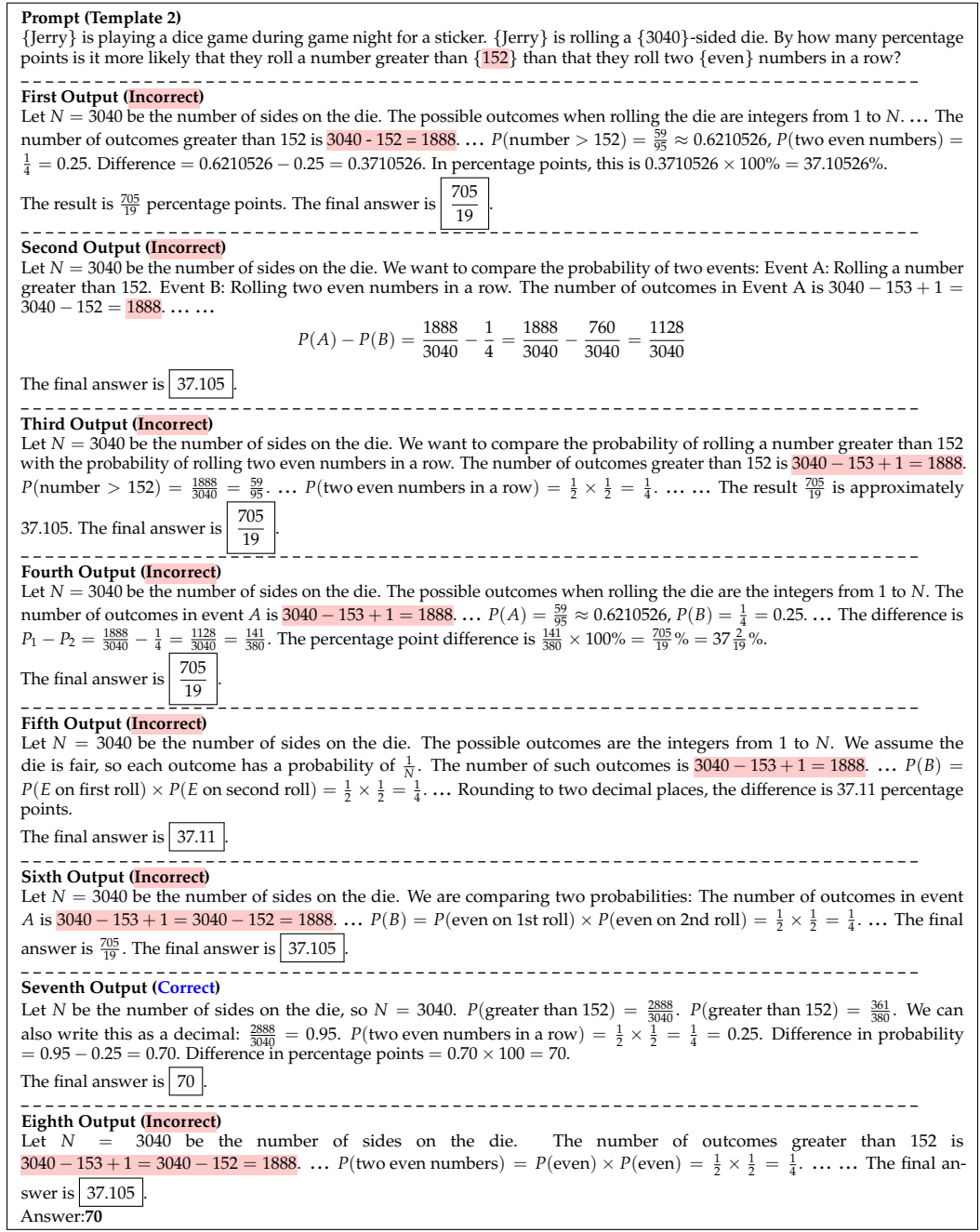

\begin{minipage}[t]{\linewidth}
\fbox{
\parbox{0.96\linewidth}{
\scriptsize
\textbf{Prompt (Template 2)}\\
\{Jerry\} is playing a dice game during game night for a sticker. \{Jerry\} is rolling a \{3040\}-sided die. By how many percentage points is it more likely that they roll a number greater than \{\hl{152}\} than that they roll two \{even\} numbers in a row?

\hdashrule{0.96\linewidth}{0.5pt}{3pt 2pt}\\
\textbf{First Output (\hl{Incorrect})}\\
Let $N = 3040$ be the number of sides on the die. The possible outcomes when rolling the die are integers from 1 to $N$.
$\textbf{\dots}$
The number of outcomes greater than 152 is \hl{3040 - 152 = 1888}.
$\textbf{\dots}$
$P(\text{number} > 152) = \frac{59}{95} \approx 0.6210526$, $P(\text{two even numbers}) = \frac{1}{4} = 0.25$.
Difference $= 0.6210526 - 0.25 = 0.3710526$. In percentage points, this is $0.3710526 \times 100\% = 37.10526\%$.

The result is $\frac{705}{19}$ percentage points. The final answer is $\boxed{\dfrac{705}{19}}$.

\hdashrule{0.96\linewidth}{0.5pt}{3pt 2pt}\\
\textbf{Second Output (\hl{Incorrect})}\\
Let $N = 3040$ be the number of sides on the die. We want to compare the probability of two events: Event A: Rolling a number greater than 152. Event B: Rolling two even numbers in a row.
The number of outcomes in Event A is $3040 - 153 + 1 = 3040 - 152 = \text{\hl{$1888$}}$.
$\textbf{\dots}$
$\textbf{\dots}$
\[
P(A) - P(B) = \frac{1888}{3040} - \frac{1}{4} = \frac{1888}{3040} - \frac{760}{3040} = \frac{1128}{3040}
\]
The final answer is \boxed{37.105}.

\hdashrule{0.96\linewidth}{0.5pt}{3pt 2pt}\\
\textbf{Third Output (\hl{Incorrect})}\\
Let $N = 3040$ be the number of sides on the die. We want to compare the probability of rolling a number greater than 152 with the probability of rolling two even numbers in a row.
The number of outcomes greater than 152 is \hl{$3040 - 153 + 1 = 1888$}.
$P(\text{number} > 152) = \frac{1888}{3040} = \frac{59}{95}$.
$\textbf{\dots}$
$P(\text{two even numbers in a row}) = \frac{1}{2} \times \frac{1}{2} = \frac{1}{4}$.
$\textbf{\dots}$
$\textbf{\dots}$
The result $\frac{705}{19}$ is approximately $37.105$. The final answer is $\boxed{\dfrac{705}{19}}$.

\hdashrule{0.96\linewidth}{0.5pt}{3pt 2pt}\\
\textbf{Fourth Output (\hl{Incorrect})}\\
Let $N = 3040$ be the number of sides on the die. The possible outcomes when rolling the die are the integers from 1 to $N$.
The number of outcomes in event $A$ is \hl{$3040 - 153 + 1 = 1888$}.
$\textbf{\dots}$
$P(A) = \frac{59}{95} \approx 0.6210526$, $P(B) = \frac{1}{4} = 0.25$.
$\textbf{\dots}$
The difference is $P_1 - P_2 = \frac{1888}{3040} - \frac{1}{4} = \frac{1128}{3040} = \frac{141}{380}$.
The percentage point difference is $\frac{141}{380} \times 100\% = \frac{705}{19}\% = 37\frac{2}{19}\%$.

The final answer is $\boxed{\dfrac{705}{19}}$.

\hdashrule{0.96\linewidth}{0.5pt}{3pt 2pt}\\
\textbf{Fifth Output (\hl{Incorrect})}\\
Let $N = 3040$ be the number of sides on the die. The possible outcomes are the integers from 1 to $N$. We assume the die is fair, so each outcome has a probability of $\frac{1}{N}$.
The number of such outcomes is \hl{$3040 - 153 + 1 = 1888$}.
$\textbf{\dots}$
$P(B) = P(E \text{ on first roll}) \times P(E \text{ on second roll}) = \frac{1}{2} \times \frac{1}{2} = \frac{1}{4}$.
$\textbf{\dots}$
Rounding to two decimal places, the difference is $37.11$ percentage points.

The final answer is $\boxed{37.11}$.

\hdashrule{0.96\linewidth}{0.5pt}{3pt 2pt}\\
\textbf{Sixth Output (\hl{Incorrect})}\\
Let $N = 3040$ be the number of sides on the die. We are comparing two probabilities:
The number of outcomes in event $A$ is \hl{$3040 - 153 + 1 = 3040 - 152 = 1888$}.
$\textbf{\dots}$
$P(B) = P(\text{even on 1st roll}) \times P(\text{even on 2nd roll}) = \frac{1}{2} \times \frac{1}{2} = \frac{1}{4}$.
$\textbf{\dots}$
The final answer is $\frac{705}{19}$. The final answer is $\boxed{37.105}$.

\hdashrule{0.96\linewidth}{0.5pt}{3pt 2pt}\\
\textbf{Seventh Output (\textcolor{blue}{Correct})}\\
Let $N$ be the number of sides on the die, so $N = 3040$.
$P(\text{greater than } 152) = \frac{2888}{3040}$.
$P(\text{greater than } 152) = \frac{361}{380}$.
We can also write this as a decimal: $\frac{2888}{3040} = 0.95$.
$P(\text{two even numbers in a row}) = \frac{1}{2} \times \frac{1}{2} = \frac{1}{4} = 0.25$.
Difference in probability $= 0.95 - 0.25 = 0.70$.
Difference in percentage points $= 0.70 \times 100 = 70$.

The final answer is \boxed{70}.

\hdashrule{0.96\linewidth}{0.5pt}{3pt 2pt}\\
\textbf{Eighth Output (\hl{Incorrect})}\\
Let $N = 3040$ be the number of sides on the die.
The number of outcomes greater than 152 is \hl{$3040 - 153 + 1 = 3040 - 152 = 1888$}.
$\textbf{\dots}$
$P(\text{two even numbers}) = P(\text{even}) \times P(\text{even}) = \frac{1}{2} \times \frac{1}{2} = \frac{1}{4}$.
$\textbf{\dots}$
$\textbf{\dots}$
The final answer is $\boxed{37.105}$.
\\
Answer:\textbf{70}

}
}
\end{minipage}
\caption{Eight generated outputs from Gemini 2.5 Flash Lite evaluated on Template 2.
Error type: wrong subtraction.
}
\label{fig:appendix_example_fails_gemini2}
\end{figure}

To supplement the qualitative examples in Figure~\ref{fig:example_fails} (Section~\ref{sec:observation}), we provide models' responses that consistently produce erroneous outputs for certain prompts in Figures~\ref{fig:appendix_example_fails_qwen5} and~\ref{fig:appendix_example_fails_low8}.
The examples illustrate that Qwen2.5-Math-7B-Instruct exhibits repetitive errors for certain ratios of sugar teaspoons to water cups, frequently invoking a spurious ``standard US measurement'' heuristic to convert cups into tablespoons or teaspoons.
For gpt-oss-20b-low on Template 7, the model shows striking failures on the basic arithmetic $84\times(2/7)$, while being generally robust to other fraction values.
We additionally provide dominant failure cases for all model-template pairs and observe various error types.
These include simple arithmetic errors in addition/subtraction (\ref{fig:appendix_example_fails_qwen3},~\ref{fig:appendix_example_fails_gemini2}), multiplication/division (\ref{fig:appendix_example_fails_qwen0},~\ref{fig:appendix_example_fails_qwen6},~\ref{fig:appendix_example_fails_qwen1},~\ref{fig:appendix_example_fails_low7},~\ref{fig:appendix_example_fails_gemini0},~\ref{fig:appendix_example_fails_gemini6}), transcription (\ref{fig:appendix_example_fails_low0}), unnecessary approximation (\ref{fig:appendix_example_fails_low2}), and problem misinterpretation (\ref{fig:appendix_example_fails_low8}).
Failure-inducing parameter values and the onset of erroneous reasoning in the generation traces are highlighted in \hl{red}.

\begin{figure}[b]
\begin{minipage}[t]{\linewidth}
\fbox{
\parbox{0.96\linewidth}{
\scriptsize
\textbf{Prompt (Template 0)}\\
A \{dolphin\} is \{1276\} feet long. \{11\} \{remoras\} attach to it, each measuring \{\hl{348}\} inches in length. What percentage of the \{dolphin\}'s body length is the combined length of the \{remoras\}?

\hdashrule{0.96\linewidth}{0.5pt}{3pt 2pt}\\
\textbf{First Output (Qwen2.5-Math-7B-Instruct) (\hl{Incorrect})}\\
To determine what percentage of the dolphin's body length is the combined length of the remoras, we need to follow these steps:
$\textbf{\dots}$
Now, we perform the division:
\[
\text{\hl{$\frac{319}{1276} \approx 0.2475$}} \]
Then, we multiply by 100 to convert the decimal to a percentage:
\[ 0.2475 \times 100 = 24.75 \]
Therefore, the combined length of the remoras is \(\boxed{24.75}\) percent of the dolphin's body length.

\hdashrule{0.96\linewidth}{0.5pt}{3pt 2pt}\\
\textbf{Second Output (\textcolor{blue}{Correct})}\\
To determine what percentage of the dolphin's body length is the combined length of the remoras, we need to follow these steps:
$\textbf{\dots}$
Perform the division and multiplication:
\[ \frac{319}{1276} \approx 0.25 \]
\[ 0.25 \times 100 = 25 \]
Therefore, the combined length of the remoras is \(\boxed{25}\) percent of the dolphin's body length.

\hdashrule{0.96\linewidth}{0.5pt}{3pt 2pt}\\
\textbf{Third Output (\textcolor{blue}{Correct})}\\
To determine what percentage of the dolphin's body length is the combined length of the remoras, we need to follow these steps:
$\textbf{\dots}$
\[ \left( \frac{319}{1276} \right) \times 100 \approx 25 \text{ percent} \]
Thus, the combined length of the remoras is \(\boxed{25}\) percent of the dolphin's body length.

\hdashrule{0.96\linewidth}{0.5pt}{3pt 2pt}\\
\textbf{Fourth Output (\hl{Incorrect})}\\
To determine what percentage of the dolphin's body length is the combined length of the remoras, we need to follow these steps:
$\textbf{\dots}$
Now, we perform the division and multiplication:
\[ \text{\hl{$\frac{319}{1276} \approx 0.247$}} \]
\[ 0.247 \times 100 = 24.7 \]
Therefore, the combined length of the remoras is approximately 24.7\% of the dolphin's body length. The final answer is:
\[ \boxed{24.7} \]

\hdashrule{0.96\linewidth}{0.5pt}{3pt 2pt}\\
\textbf{Fifth Output (\hl{Incorrect})}\\
To determine what percentage of the dolphin's body length is the combined length of the remoras, we need to follow these steps:
$\textbf{\dots}$
\[ \left( \frac{319 \text{ feet}}{1276 \text{ feet}} \right) \times 100 = \text{\hl{$\frac{31900}{1276} \approx 24.73\%$}} \]
Therefore, the combined length of the remoras is \(\boxed{24.73}\%\) of the dolphin's body length.

\hdashrule{0.96\linewidth}{0.5pt}{3pt 2pt}\\
\textbf{Sixth Output (\textcolor{blue}{Correct})}\\
To determine what percentage of the dolphin's body length is the combined length of the remoras, we need to follow these steps:
$\textbf{\dots}$
\[ \text{Percentage} = \left( \frac{319 \text{ feet}}{1276 \text{ feet}} \right) \times 100 \approx 25\% \]
Therefore, the combined length of the remoras is \(\boxed{25}\%\) of the dolphin's body length.

\hdashrule{0.96\linewidth}{0.5pt}{3pt 2pt}\\
\textbf{Seventh Output (\hl{Incorrect})}\\
To determine what percentage of the dolphin's body length is the combined length of the remoras, we need to follow these steps:
$\textbf{\dots}$
\[ \text{Percentage} = \left( \frac{319 \text{ feet}}{1276 \text{ feet}} \right) \times 100 \approx \text{\hl{$24.72$}}\% \]
Therefore, the combined length of the remoras is \(\boxed{24.72}\%\) of the dolphin's body length.

\hdashrule{0.96\linewidth}{0.5pt}{3pt 2pt}\\
\textbf{Eighth Output (\hl{Incorrect})}\\
To determine what percentage of the dolphin's body length is the combined length of the remoras, we need to follow these steps:
$\textbf{\dots}$
Perform the division and multiplication:
\[ \frac{319}{1276} \approx \text{\hl{$0.247$}} \]
\[ 0.247 \times 100 = 24.7 \]
Therefore, the combined length of the remoras is approximately \(\boxed{24.7}\%\) of the dolphin's body length.\\
Answer:\textbf{25}
}
}
\end{minipage}
\caption{Eight independently generated outputs from Qwen2.5-Math-7B-Instruct evaluated on Template 0. Error type: wrong division.}
\label{fig:appendix_example_fails_qwen0}
\end{figure}

\begin{figure}[b]
\begin{minipage}[t]{\linewidth}
\fbox{
\parbox{0.96\linewidth}{
\scriptsize
\textbf{Prompt (Template 1)}\\
A \{coastal mist\} rolls in from the ocean to cover \{Harborview\}. It takes \{33\} minutes to cover every \{\hl{4}\} miles of the city. If the city is \{28\} miles across from the oceanfront to the opposite inland edge, how many minutes will it take for the \{coastal mist\} to cover the whole city?

\hdashrule{0.96\linewidth}{0.5pt}{3pt 2pt}\\
\textbf{First Output (Qwen2.5-Math-7B-Instruct) (\textcolor{blue}{Correct}})\\
To determine how many minutes it will take for the marine layer to cover the entire city, we start by identifying the rate at which the marine layer rolls in. The marine layer covers 4 miles in 33 minutes. We need to find out how many minutes it will take to cover 28 miles.
$\textbf{\dots}$
\[
8.25 \times 28 = 8.25 \times (30 - 2) = 8.25 \times 30 - 8.25 \times 2
\]
\[
8.25 \times 30 = 247.5
\]
\[
8.25 \times 2 = 16.5
\]
\[
247.5 - 16.5 = 231
\]
Therefore, it will take the marine layer \(\boxed{231}\) minutes to cover the whole city.

\hdashrule{0.96\linewidth}{0.5pt}{3pt 2pt}\\
\textbf{Second Output (\hl{Incorrect})}\\
To determine how many minutes it will take for the marine layer to cover the entire city, we start by identifying the rate at which the marine layer rolls in. The marine layer covers 4 miles in 33 minutes. We need to find out how many minutes it will take to cover 28 miles.
$\textbf{\dots}$
\[
\text{Time to cover 28 miles} = 8.25 \text{ minutes per mile} \times 28 \text{ miles} = \text{\hl{$229.5$}} \text{ minutes}
\]
Therefore, it will take the marine layer \(\boxed{229.5}\) minutes to cover the whole city.

\hdashrule{0.96\linewidth}{0.5pt}{3pt 2pt}\\
\textbf{Third Output (\textcolor{blue}{Correct})}\\
To determine how many minutes it will take for the marine layer to cover the entire city, we need to follow these steps:
$\textbf{\dots}$
Since it takes 33 minutes to cover each 4-mile segment, we multiply the number of segments by the time per segment:
\[
\text{Total time} = 7 \text{ segments} \times 33 \text{ minutes/segment} = 231 \text{ minutes}
\]
Therefore, it will take \(\boxed{231}\) minutes for the marine layer to cover the whole city.

\hdashrule{0.96\linewidth}{0.5pt}{3pt 2pt}\\
\textbf{Fourth Output (\hl{Incorrect})}\\
To determine how many minutes it will take for the marine layer to cover the entire city, we start by identifying the rate at which the marine layer rolls in. The marine layer covers 4 miles in 33 minutes. We need to find out how long it will take to cover 28 miles.
$\textbf{\dots}$
\[
8.25 \text{ minutes per mile} \times 28 \text{ miles} = \text{\hl{$229.5$}} \text{ minutes}
\]
Therefore, it will take the marine layer \(\boxed{229.5}\) minutes to cover the whole city.

\hdashrule{0.96\linewidth}{0.5pt}{3pt 2pt}\\
\textbf{Fifth Output (\hl{Incorrect})}\\
To determine how many minutes it will take for the marine layer to cover the entire city, we start by finding the rate at which the marine layer rolls in. The marine layer covers 4 miles in 33 minutes. We need to find out how many minutes it takes to cover 28 miles.
$\textbf{\dots}$
\[
8.25 \text{ minutes per mile} \times 28 \text{ miles} = \text{\hl{$229.5$}} \text{ minutes}
\]
Therefore, it will take the marine layer \(\boxed{229.5}\) minutes to cover the whole city.

\hdashrule{0.96\linewidth}{0.5pt}{3pt 2pt}\\
\textbf{Sixth Output (\hl{Incorrect})}\\
To determine how many minutes it will take for the marine layer to cover the entire city, we start by identifying the rate at which the marine layer rolls in. The marine layer covers 4 miles in 33 minutes. We need to find out how many minutes it will take to cover 28 miles.
$\textbf{\dots}$
\[
8.25 \text{ minutes per mile} \times 28 \text{ miles} = \text{\hl{$229.5$}} \text{ minutes}
\]
Therefore, it will take the marine layer \(\boxed{229.5}\) minutes to cover the whole city.

\hdashrule{0.96\linewidth}{0.5pt}{3pt 2pt}\\
\textbf{Seventh Output (\textcolor{blue}{Correct})}\\
To determine how many minutes it will take for the marine layer to cover the entire city, we start by finding the rate at which the marine layer covers the city. The problem states that it takes 33 minutes to cover every 4 miles.
$\textbf{\dots}$
Since it takes 33 minutes to cover each 4-mile segment, we multiply the number of segments by the time it takes to cover each segment:
\[
\text{Total time} = 7 \text{ segments} \times 33 \text{ minutes/segment} = 231 \text{ minutes}
\]
Thus, the total time it will take for the marine layer to cover the whole city is:
\[
\boxed{231}
\]

\hdashrule{0.96\linewidth}{0.5pt}{3pt 2pt}\\
\textbf{Eighth Output (\hl{Incorrect})}\\
To determine how many minutes it will take for the marine layer to cover the entire city, we start by identifying the rate at which the marine layer rolls in. The marine layer covers 4 miles in 33 minutes. We need to find out how many minutes it will take to cover 28 miles.
$\textbf{\dots}$
\[
8.25 \text{ minutes per mile} \times 28 \text{ miles} = \text{\hl{$229.5$}} \text{ minutes}
\]
Therefore, it will take the marine layer \(\boxed{229.5}\) minutes to cover the whole city.\\
Answer:\textbf{231}
}
}
\end{minipage}
\caption{Eight independently generated outputs from Qwen2.5-Math-7B-Instruct evaluated on Template 1. Error type: wrong multiplication.}
\label{fig:appendix_example_fails_qwen1}
\end{figure}

\begin{figure}[b]
\begin{minipage}[t]{\linewidth}
\fbox{
\parbox{0.96\linewidth}{
\scriptsize
\textbf{Prompt (Template 3)}\\
A group of \{71\} students has various hobbies. \{4\} like to read, \{6\} like to play basketball, and the rest like to either play music or write. How many like to play music if the number that like to write is \{\hl{10}\} times the number that prefer playing basketball?

\hdashrule{0.96\linewidth}{0.5pt}{3pt 2pt}\\
\textbf{First Output (Qwen2.5-Math-7B-Instruct) (\hl{Incorrect})}\\
To determine how many students like to play music, we need to follow these steps:
$\textbf{\dots}$
\[
\text{Total} = \text{\hl{4 + 6 + 60 = $64$}}
\]
\[
\text{Number of students who like to play music} = 71 - 64 = 7
\]
Thus, the number of students who like to play music is \(\boxed{7}\).

\hdashrule{0.96\linewidth}{0.5pt}{3pt 2pt}\\
\textbf{Second Output (\hl{Incorrect})}\\
To determine how many students like to play music, we start by identifying the number of students who like to write. According to the problem, the number of students who like to write is 10 times the number of students who like to play basketball.
$\textbf{\dots}$
\[
\text{\hl{71 - 4 - 60 = $7$}}
\]
Thus, the number of students who like to play music is \(\boxed{7}\).

\hdashrule{0.96\linewidth}{0.5pt}{3pt 2pt}\\
\textbf{Third Output (\textcolor{blue}{Correct})}\\
To determine how many students like to play music, we start by identifying the number of students who like to write. 
$\textbf{\dots}$
Next, let \( m \) represent the number of students who like to play music. Then:
\[
4 + 6 + m + 60 = 71
\]
\[
70 + m = 71
\]
\[
m = 1
\]
Therefore, the number of students who like to play music is \(\boxed{1}\).

\hdashrule{0.96\linewidth}{0.5pt}{3pt 2pt}\\
\textbf{Fourth Output (\hl{Incorrect})}\\
To determine how many students like to play music, we need to follow these steps:
$\textbf{\dots}$
\[
\text{Total} = \text{\hl{4 + 6 + 60 = $66$}}
\]
\[
\text{Number of students who like to play music} = 71 - 66 = 5
\]
Thus, the number of students who like to play music is \(\boxed{5}\).

\hdashrule{0.96\linewidth}{0.5pt}{3pt 2pt}\\
\textbf{Fifth Output (\hl{Incorrect})}\\
To determine how many students like to play music, we start by identifying the number of students who like to write. 
$\textbf{\dots}$
\[
\text{\hl{4 + 6 + 60 = $66$}}
\]
Since the total number of students is 71, the number of students who like to play music is:
\[
71 - 66 = 5
\]
Thus, the number of students who like to play music is:
\[
\boxed{5}
\]

\hdashrule{0.96\linewidth}{0.5pt}{3pt 2pt}\\
\textbf{Sixth Output (\hl{Incorrect})}\\
To determine how many students like to play music, we need to follow these steps:
$\textbf{\dots}$
\[
\text{Total} = \text{\hl{4 + 6 + 60 = $60$}}
\]
\[
\text{Number of students who like to play music} = 71 - 60 = 11
\]
Thus, the number of students who like to play music is \(\boxed{11}\).

\hdashrule{0.96\linewidth}{0.5pt}{3pt 2pt}\\
\textbf{Seventh Output (\textcolor{blue}{Correct})}\\
To determine how many students like to play music, we start by identifying the number of students who like to write. 
$\textbf{\dots}$
\[
71 - (4 + 6 + 60) = 71 - 70 = 1
\]
Thus, the number of students who like to play music is \(\boxed{1}\).

\hdashrule{0.96\linewidth}{0.5pt}{3pt 2pt}\\
\textbf{Eighth Output (\hl{Incorrect})}\\
To determine how many students like to play music, we need to follow these steps:
$\textbf{\dots}$
\[
\text{Total} = 4 + 6 + 60 = \text{\hl{$66$}}
\]
\[
\text{Number of students who like to play music} = 71 - 66 = \text{\hl{$5$}}
\]
Thus, the number of students who like to play music is \(\boxed{5}\).\\
Answer:\textbf{1}
}
}
\end{minipage}
\caption{Eight independently generated outputs from Qwen2.5-Math-7B-Instruct evaluated on Template 3. Error type: wrong addition/subtraction.}
\label{fig:appendix_example_fails_qwen3}
\end{figure}

\begin{figure}
\begin{minipage}[t]{\linewidth}
\fbox{
\parbox{0.96\linewidth}{
\scriptsize
\textbf{Prompt (Template 6)}\\
\{John\} has \{15\} \{hectares\} of \{pineapples\} field. There are \{\hl{47}\} \{pineapples\} per \{hectare\}. \{John\} can harvest his \{pineapples\} every \{2\} months. How many \{pineapples\} can \{John\} harvest within a year?

\hdashrule{0.96\linewidth}{0.5pt}{3pt 2pt}\\
\textbf{First Output (Qwen2.5-Math-7B-Instruct) (\hl{Incorrect})}\\
To determine how many pineapples John can harvest within a year, we need to follow these steps:
$\textbf{\dots}$
\[
\text{Total number of pineapples} = 15 \text{ hectares} \times 47 \text{ pineapples/hectare} = \text{\hl{$655$}} \text{ pineapples}
\]
Therefore, the total number of pineapples John can harvest within a year is \(\boxed{3930}\).

\hdashrule{0.96\linewidth}{0.5pt}{3pt 2pt}\\
\textbf{Second Output (\hl{Incorrect})}\\
To determine how many pineapples John can harvest within a year, we need to follow these steps:
$\textbf{\dots}$
\[
\text{Total number of pineapples} = 15 \text{ hectares} \times 47 \text{ pineapples/hectare} = \text{\hl{$695$}} \text{ pineapples}
\]
Therefore, the total number of pineapples John can harvest within a year is \(\boxed{4170}\).

\hdashrule{0.96\linewidth}{0.5pt}{3pt 2pt}\\
\textbf{Third Output (\hl{Incorrect})}\\
To determine how many pineapples John can harvest within a year, we need to follow these steps:
$\textbf{\dots}$
\[
15 \text{ hectares} \times 47 \text{ pineapples/hectare} = \text{\hl{$655$}} \text{ pineapples}
\]
Therefore, the number of pineapples John can harvest within a year is \(\boxed{3930}\).

\hdashrule{0.96\linewidth}{0.5pt}{3pt 2pt}\\
\textbf{Fourth Output (\hl{Incorrect})}\\
To determine how many pineapples John can harvest within a year, we need to follow these steps:
$\textbf{\dots}$
\[
15 \text{ hectares} \times 47 \text{ pineapples/hectare} = \text{\hl{$655$}} \text{ pineapples}
\]
Therefore, the total number of pineapples John can harvest within a year is:
\[
\boxed{3930}
\]
\hdashrule{0.96\linewidth}{0.5pt}{3pt 2pt}\\
\textbf{Fifth Output (\hl{Incorrect})}\\
To determine how many pineapples John can harvest within a year, we need to follow these steps:
$\textbf{\dots}$
\[
\text{Total number of pineapples} = 15 \text{ hectares} \times 47 \text{ pineapples/hectare} = \text{\hl{$695$}} \text{ pineapples}
\]
Therefore, the number of pineapples John can harvest within a year is \(\boxed{4170}\).

\hdashrule{0.96\linewidth}{0.5pt}{3pt 2pt}\\
\textbf{Sixth Output (\hl{Incorrect})}\\
To determine how many pineapples John can harvest within a year, we need to follow these steps:
$\textbf{\dots}$
\[
\text{Total number of pineapples} = 15 \text{ hectares} \times 47 \text{ pineapples/hectare} = \text{\hl{$655$}} \text{ pineapples}
\]
Therefore, the total number of pineapples John can harvest within a year is \(\boxed{3930}\).

\hdashrule{0.96\linewidth}{0.5pt}{3pt 2pt}\\
\textbf{Seventh Output (\textcolor{blue}{Correct})}\\
To determine how many pineapples John can harvest within a year, we need to follow these steps:\
$\textbf{\dots}$
\[
15 \text{ hectares} \times 47 \text{ pineapples/hectare} = 705 \text{ pineapples}
\]
Therefore, the number of pineapples John can harvest within a year is \(\boxed{4230}\).\\
Answer: \textbf{4230}
}
}
\end{minipage}
\caption{Seven of eight independently generated outputs from Qwen2.5-Math-7B-Instruct evaluated on Template 6. Error type: wrong multiplication.}
\label{fig:appendix_example_fails_qwen6}
\end{figure}

\begin{figure}
\begin{minipage}[t]{\linewidth}
\fbox{
\parbox{0.96\linewidth}{
\scriptsize
\textbf{Prompt (Template 0)}\\
\{Benny\} saw a \{1920\}-foot whale with \{12\} remoras, each \{\hl{192}\} inches long, attached to it. What percentage of the whale's body length is the combined length of the remoras?

\hdashrule{0.96\linewidth}{0.5pt}{3pt 2pt}\\
\textbf{First Output (Qwen2.5-Math-7B-Instruct) (\hl{Incorrect})}\\
analysisWe need combined length of remoras: 12 * 192 inches. Convert to feet: 192 inches = 16 feet. So each 16 ft, 12*16 = 192 ft. \hl{Whale length 192 ft}. Percentage = (192/192)*100 = 100
$\textbf{\dots}$
The whale's body length is given as \(\text{\hl{$192$}}\) feet. Thus, the combined remora length equals the whale's length:
\[
\frac{192\text{ ft}}{192\text{ ft}} = 1 \quad\Longrightarrow\quad \text{\hl{$100\%$}}
\]
\[
\boxed{100\%}
\]

\hdashrule{0.96\linewidth}{0.5pt}{3pt 2pt}\\
\textbf{Second Output (\textcolor{blue}{Correct})}\\
analysisCompute remoras total length: 12 * 192 inches = 2304 inches. Convert whale length: 1920 ft = 1920*12=23040 inches. Percentage = 2304/23040=0.1 =10\%.
$\textbf{\dots}$
Thus the percentage is:
\[
\frac{2304}{23040} \times 100\% = 10\%
\]
\[
\boxed{10\%}
\]

\hdashrule{0.96\linewidth}{0.5pt}{3pt 2pt}\\
\textbf{Third Output (\textcolor{blue}{Correct})}\\
analysisNeed remora combined length: 12 * 192 inches = 2304 inches = 192 inches? Wait 192 inches = 16 ft. 12*16=192 ft. Whale length 1920 ft. Percentage = 192/1920=0.10=10\%.
$\textbf{\dots}$
As a percentage,
\[
0.10 \times 100\% = 10\%
\]
\[
\boxed{10\%}
\]

\hdashrule{0.96\linewidth}{0.5pt}{3pt 2pt}\\
\textbf{Fourth Output (\hl{Incorrect})}\\
analysisConvert remora length to feet: 192 inches /12=16 ft. 12 remoras $\rightarrow$192 ft. \hl{Whale 192 ft}. Percentage =192/192*100=100\%.

$\textbf{\dots}$
\[
\boxed{100\%}
\]

\hdashrule{0.96\linewidth}{0.5pt}{3pt 2pt}\\
\textbf{Fifth Output (\hl{Incorrect})}\\
analysisWe need combined length of remoras: 12 * 192 inches. Convert to feet: 192 inches = 16 feet (since 12 inches =1 ft, 192/12=16). So each remora 16 ft. 12*16=192 ft. \hl{Whale length 192 ft}. So percentage = (192/192)*100 =100\%. Box.
$\textbf{\dots}$
\[
\text{Combined remora length} = 12\times192\text{ in} = 12\times16\text{ ft} = 192\text{ ft}
\]
Whale length \(= 192\text{ ft}.\)
\[
\text{Percentage} = \frac{192}{192}\times100\% = 100\%
\]
\[
\boxed{100\%}
\]

\hdashrule{0.96\linewidth}{0.5pt}{3pt 2pt}\\
\textbf{Sixth Output (\textcolor{blue}{Correct})}\\
analysisCompute total remora length: 12 * 192 inches. Convert to feet: 12*192 = 2304 inches. 12 feet? Wait 2304/12 = 192 feet. Whale length 1920 feet. Percentage = 192 / 1920 *100 = 10\%.
$\textbf{\dots}$
The whale is 1920~ft long. The percentage of the whale's length that the remoras occupy is:
\[
\frac{192 \text{ ft}}{1920 \text{ ft}}\times 100\% = 10\%
\]
\[
\boxed{10\%}
\]

\hdashrule{0.96\linewidth}{0.5pt}{3pt 2pt}\\
\textbf{Seventh Output (\hl{Incorrect})}\\
analysisConvert remora length to feet:192 in=16 ft. Each 16 ft. 12 remoras total 12*16=192 ft. \hl{Whale length 192 ft}. Percentage =192/192*100=100\%.
$\textbf{\dots}$
The whale is also \(192\)~ft long, so
\[
\frac{192\text{~ft}}{192\text{~ft}}\times 100\% = 100\%
\]
\[
\boxed{100\%}
\]

\hdashrule{0.96\linewidth}{0.5pt}{3pt 2pt}\\
\textbf{Eighth Output (\hl{Incorrect})}\\
analysisCompute remora total length: 12*192 inches. Convert to feet: inches/12. 192in=16ft. 12*16=192ft. \hl{Whale length 192ft}. Ratio 192/192=1 $\rightarrow$ 100\%.
$\textbf{\dots}$
The whale is \(192\) feet long, so the combined remora length equals the whale's length.
\[
\frac{192\text{ ft}}{192\text{ ft}} = 1 \;\Longrightarrow\; 100\%
\]
\[
\boxed{100\%}
\]
Answer: \textbf{10}
}
}
\end{minipage}
\caption{Eight independently generated outputs from gpt-oss-20b-low evaluated on Template 0. Error type: transcription.}
\label{fig:appendix_example_fails_low0}
\end{figure}

\begin{figure}
\begin{minipage}[t]{\linewidth}
\fbox{
\parbox{0.96\linewidth}{
\scriptsize
\textbf{Prompt (Template 7)}\\
\{Jamie\} is juggling at a circus. \{Jamie\} can juggle \{928\} balls. \{3/4\} of the balls are golf balls, and \{\hl{7/8}\} of the golf balls are blue. How many blue golf balls are there?

\hdashrule{0.96\linewidth}{0.5pt}{3pt 2pt}\\
\textbf{First Output (gpt-oss-20b-low) (\textcolor{blue}{Correct})}\\
To determine the number of blue golf balls Jamie has, we need to follow these steps:
$\textbf{\dots}$
\[
696 \times \frac{7}{8} = 696 \div 8 \times 7 = 87 \times 7 = 609
\]
So, Jamie has 609 blue golf balls. Therefore, the number of blue golf balls is \(\boxed{609}\).

\hdashrule{0.96\linewidth}{0.5pt}{3pt 2pt}\\
\textbf{Second Output (\hl{Incorrect})}\\
To determine the number of blue golf balls Jamie has, we need to follow these steps:
$\textbf{\dots}$
\[
696 \times \frac{7}{8} = \text{\hl{$696 \times 0.875 = 612$}}
\]
So, Jamie has 612 blue golf balls. The final answer is:
\[
\boxed{612}
\]

\hdashrule{0.96\linewidth}{0.5pt}{3pt 2pt}\\
\textbf{Third Output (\hl{Incorrect})}\\
To determine the number of blue golf balls Jamie is juggling, we need to follow these steps:
$\textbf{\dots}$
\[
696 \times \frac{7}{8} = \text{\hl{$696 \times 0.875 = 612$}}
\]
So, Jamie has 612 blue golf balls. Therefore, the number of blue golf balls is \(\boxed{612}\).

\hdashrule{0.96\linewidth}{0.5pt}{3pt 2pt}\\
\textbf{Fourth Output (\hl{Incorrect})}\\
To determine the number of blue golf balls Jamie is juggling, we need to follow these steps:
$\textbf{\dots}$
\[
696 \times \frac{7}{8} = \text{\hl{$696 \times 0.875 = 612$}}
\]
So, Jamie has 612 blue golf balls. The number of blue golf balls is \(\boxed{612}\).

\hdashrule{0.96\linewidth}{0.5pt}{3pt 2pt}\\
\textbf{Fifth Output (\textcolor{blue}{Correct})}\\
To determine the number of blue golf balls Jamie is juggling, we need to follow these steps:
$\textbf{\dots}$
\[
696 \times \frac{7}{8} = 696 \div 8 \times 7 = 87 \times 7 = 609
\]
So, Jamie has 609 blue golf balls. The final answer is:
\[
\boxed{609}
\]

\hdashrule{0.96\linewidth}{0.5pt}{3pt 2pt}\\
\textbf{Sixth Output (\hl{Incorrect})}\\
To determine the number of blue golf balls Jamie has, we need to follow these steps:
$\textbf{\dots}$
\[
696 \times \frac{7}{8} = \text{\hl{$696 \times 0.875 = 612$}}
\]
Therefore, Jamie has 612 blue golf balls. The final answer is:
\[
\boxed{612}
\]

\hdashrule{0.96\linewidth}{0.5pt}{3pt 2pt}\\
\textbf{Seventh Output (\textcolor{blue}{Correct})}\\
To determine the number of blue golf balls Jamie is juggling, we need to follow these steps:
$\textbf{\dots}$
\[
7 \times 87 = 609
\]
So, there are 609 blue golf balls. Therefore, the number of blue golf balls is \(\boxed{609}\).\\
Answer: \textbf{609}
}
}
\end{minipage}
\caption{Seven of eight independently generated outputs from gpt-oss-20b-low evaluated on Template 7. Error type: wrong multiplication.}
\label{fig:appendix_example_fails_low7}
\end{figure}

\begin{figure}
\begin{minipage}[t]{\linewidth}
\fbox{
\parbox{0.96\linewidth}{
\scriptsize
\textbf{Prompt (Template 8)}\\
\{Jaime\} places eggs on some tables. Each table can hold \{\hl{7}\} eggs. If he has \{57\} eggs and \{7\} tables, how many eggs won't he be able to place on the table?

\hdashrule{0.96\linewidth}{0.5pt}{3pt 2pt}\\
\textbf{First Output (gpt-oss-20b-low) (\hl{Incorrect})}\\
analysis\hl{We compute 57 mod 7. 7*8=56, remainder 1}. So answer 1 egg.
\[
\boxed{1}
\]

\hdashrule{0.96\linewidth}{0.5pt}{3pt 2pt}\\
\textbf{Second Output (\hl{Incorrect})}\\
analysis\hl{We need remainder 57 mod 7. 7*8=56, remainder1}. So 1 egg can't be placed.
\[
\boxed{1}
\]

\hdashrule{0.96\linewidth}{0.5pt}{3pt 2pt}\\
\textbf{Third Output (\textcolor{blue}{Correct})}\\
analysisWe need to see: 7 tables each hold 7 eggs $\Rightarrow$ max capacity 49? Wait 7 tables *7 eggs =49. He has 57 eggs. So leftover =57-49=8. So answer 8.
1. \textbf{Capacity of the tables}:
   Each table holds \(7\) eggs. With \(7\) tables, the total capacity is
   \[ 7 \text{ tables} \times 7 \text{ eggs/table} = 49 \text{ eggs}. \]
2. \textbf{Eggs he has}: Jaime has \(57\) eggs.
3. \textbf{Eggs that cannot be placed}: Subtract the capacity from the total eggs:
\[
57 - 49 = 8
\]
\[
\boxed{8}
\]

\hdashrule{0.96\linewidth}{0.5pt}{3pt 2pt}\\
\textbf{Fourth Output (\hl{Incorrect})}\\
analysis\hl{Compute 57 mod 7. 7*8=56 remainder 1}.
\[
\boxed{1}
\]

\hdashrule{0.96\linewidth}{0.5pt}{3pt 2pt}\\
\textbf{Fifth Output (\hl{Incorrect})}\\
analysis\hl{We need remainder of 57 divided by 7. 7*8=56, remainder 1}. So 1 egg won't fit.
The number of eggs that cannot be placed is the remainder when 57 is divided by 7.
\[ 57\div 7=8\text{ with a remainder of }1. \]
So \(\boxed{1}\) egg cannot be placed on the tables.

\hdashrule{0.96\linewidth}{0.5pt}{3pt 2pt}\\
\textbf{Sixth Output (\textcolor{blue}{Correct})}\\
analysisWe need 57 eggs, each table holds 7, 7 tables. Max capacity 7*7 =49. Eggs unused =57-49=8. So answer 8.
\[
\boxed{8}
\]

\hdashrule{0.96\linewidth}{0.5pt}{3pt 2pt}\\
\textbf{Seventh Output (\hl{Incorrect})}\\
analysis\hl{We compute 57 mod 7. 7*8=56, remainder 1}. So 1 egg left.
To find the number of eggs that cannot be placed, divide the total eggs by the capacity of each table and look at the remainder:
\[ 57 \div 7 = 8 \text{ with a remainder of } 1 \]
So after using 8 tables for 56 eggs, there is \textbf{1 egg} that cannot be placed.
\[
\boxed{1}
\]

\hdashrule{0.96\linewidth}{0.5pt}{3pt 2pt}\\
\textbf{Eighth Output (\hl{Incorrect})}\\
analysis\hl{Compute 57 mod 7. 7*8=56 remainder 1}. So 1 egg cannot be placed.
\[
\boxed{1}
\]
\\
Answer: \textbf{8}
}
}
\end{minipage}
\caption{Eight generated outputs from gpt-oss-20b-low evaluated on Template 8. Error type: problem misinterpretation.}
\label{fig:appendix_example_fails_low8}
\end{figure}

\begin{figure}
\begin{minipage}[t]{\linewidth}
\fbox{
\parbox{0.96\linewidth}{
\scriptsize
\textbf{Prompt (Template 0)}\\
A dolphin is \{2090\} feet long. \{19\} remoras, each \{264\} inches long, attach to it. What percentage of the dolphin's body length is the combined length of the remoras?

\hdashrule{0.96\linewidth}{0.5pt}{3pt 2pt}\\
\textbf{First Output (Qwen2.5-Math-7B-Instruct) (\hl{Incorrect})}\\
Here's how to solve this problem step-by-step:
$\textbf{\dots}$

\textbf{Combined length of remoras in inches:}
\begin{itemize}
    \item \hl{19 remoras * 264 inches/remora = 4960 + 792 = 4960 + 800 - 8 = 5760 - 8 = 5752} inches
\end{itemize}

\textbf{2. Calculate the combined length of the remoras as a percentage of the dolphin's length.}
$\textbf{\dots}$

\textbf{Answer:}
The combined length of the remoras is approximately 22.93\% of the dolphin's body length.

\hdashrule{0.96\linewidth}{0.5pt}{3pt 2pt}\\
\textbf{Second Output (\hl{Incorrect})}\\
Here's how to solve this problem step-by-step:
$\textbf{\dots}$

\textbf{2. Calculate the combined length of the remoras.}
\begin{itemize}
    \item Number of remoras = 19
    \item Length of each remora = 264 inches
    \item Combined length of remoras = \hl{19 remoras * 264 inches/remora = 4956 inches}
\end{itemize}
$\textbf{\dots}$

The final answer is $\boxed{19.8\%}$.

\hdashrule{0.96\linewidth}{0.5pt}{3pt 2pt}\\
\textbf{Third Output (\hl{Incorrect})}\\
Here's how to solve this problem step-by-step:

\textbf{2. Calculate the combined length of the remoras.}
\begin{itemize}
    \item Combined remora length = \hl{19 remoras * 264 inches/remora = 4996 inches}
\end{itemize}

The final answer is \boxed{19.92\%}.

\hdashrule{0.96\linewidth}{0.5pt}{3pt 2pt}\\
\textbf{Fourth Output (\hl{Incorrect})}\\
Here's how to solve this problem step-by-step:

\textbf{1. Convert all measurements to the same unit.}
It's easiest to work with inches since the remoras are already measured in inches.
$\textbf{\dots}$

\textbf{2. Calculate the combined length of the remoras.}
\begin{itemize}
    \item There are 19 remoras.
    \item \hl{19 remoras * 264 inches/remora = 4968 inches}
\end{itemize}
$\textbf{\dots}$

The final answer is \boxed{19.81\%}.

\hdashrule{0.96\linewidth}{0.5pt}{3pt 2pt}\\
\textbf{Fifth Output (\textcolor{blue}{Correct})}\\
Here's how to solve this problem:
$\textbf{\dots}$

\textbf{5. Calculate the Combined Length of the Remoras:}
\[
\text{Combined remora length} = 19 \times 264 \text{ inches} = 5016 \text{ inches}
\]
$\textbf{\dots}$

\textbf{9. Final Answer:} The combined length of the remoras is 20\% of the dolphin's body length. The final answer is $\boxed{20\%}$.

\hdashrule{0.96\linewidth}{0.5pt}{3pt 2pt}\\
\textbf{Sixth Output (\hl{Incorrect})}\\
Here's how to solve this problem:
$\textbf{\dots}$

\textbf{2. Calculate the combined length of the remoras.}
\begin{itemize}
    \item Number of remoras: 19
    \item Length of each remora: 264 inches
    \item Combined remora length: \hl{19 remoras * 264 inches/remora = 4996 inches}
\end{itemize}
$\textbf{\dots}$

The final answer is \boxed{19.9\%}.

\hdashrule{0.96\linewidth}{0.5pt}{3pt 2pt}\\
\textbf{Seventh Output (\hl{Incorrect})}\\
Here's how to solve this problem step-by-step:
$\textbf{\dots}$

\textbf{2. Calculate the combined length of the remoras.}
\begin{itemize}
    \item There are 19 remoras, and each is 264 inches long.
    \item Combined remora length = \hl{19 remoras * 264 inches/remora = 4916} inches
\end{itemize}
$\textbf{\dots}$

The final answer is \boxed{19.6\%}.

\hdashrule{0.96\linewidth}{0.5pt}{3pt 2pt}\\
\textbf{Eighth Output (\hl{Incorrect})}\\
Here's how to solve this problem step-by-step:
$\textbf{\dots}$

\textbf{2. Calculate the combined length of the remoras.}
There are 19 remoras, and each is 264 inches long.
\begin{itemize}
    \item Combined remora length = \hl{19 remoras * 264 inches/remora = 4916 inches}
\end{itemize}
$\textbf{\dots}$

The final answer is \boxed{19.6\%}.
\\
Answer: \textbf{20}
}
}
\end{minipage}
\caption{Eight generated outputs from Gemini 2.5 Flash Lite evaluated on Template 0. Error type: wrong multiplication.}
\label{fig:appendix_example_fails_gemini0}
\end{figure}

\begin{figure}
\begin{minipage}[t]{\linewidth}
\fbox{
\parbox{0.96\linewidth}{
\scriptsize
\textbf{Prompt (Template 6)}\\
\{John\} has \{\hl{35}\} \{square yard\}s of \{pineapple\}s field. There are \{\hl{43}\} \{pineapple\}s per \{square yard\}. \{John\} can harvest his \{pineapple\}s every \{2\} months. How many \{pineapple\}s can \{John\} harvest within a year?"

\hdashrule{0.96\linewidth}{0.5pt}{3pt 2pt}\\
\textbf{First Output (Gemini 2.5 Flash Lite) (\textcolor{blue}{Correct})}\\
Here's how to solve the problem step-by-step:\\

1.  \textbf{Calculate the total number of pineapples:}\\
    *   John has 35 square yards of field.\\
    *   There are 43 pineapples per square yard.\\
    *   Total pineapples = 35 square yards * 43 pineapples/square yard = 1505 pineapples.\\

$\textbf{\dots}$

The final answer is $\boxed{9030}$.

\hdashrule{0.96\linewidth}{0.5pt}{3pt 2pt}\\
\textbf{Second Output (\hl{Incorrect})}\\
Here's how to solve the problem step-by-step:\\

1.  \textbf{Calculate the total number of pineapples:}\\
    *   John has 35 square yards of field.\\
    *   There are 43 pineapples per square yard.\\
    *   Total pineapples = 35 square yards * 43 pineapples/square yard\\
    *   Total pineapples = \hl{1455} pineapples\\

$\textbf{\dots}$

The final answer is $\boxed{8730}$.

\hdashrule{0.96\linewidth}{0.5pt}{3pt 2pt}\\
\textbf{Third Output (\hl{Incorrect})}\\
Here's how to solve the problem step-by-step:\\

1.  \textbf{Calculate the total number of pineapples John has:}\\
    *   He has 35 square yards.\\
    *   There are 43 pineapples per square yard.\\
    *   Total pineapples = 35 square yards * 43 pineapples/square yard\\

    *   \hl{35 * 43 = 1495} pineapples\\

$\textbf{\dots}$

The final answer is $\boxed{8970}$.

\hdashrule{0.96\linewidth}{0.5pt}{3pt 2pt}\\
\textbf{Fourth Output (\hl{Incorrect})}\\
Here's how to solve the problem step-by-step:\\

1.  \textbf{Calculate the total number of pineapples:}\\
    *   John has 35 square yards of field.\\
    *   There are 43 pineapples per square yard.\\
    *   Total pineapples = \hl{35 square yards * 43 pineapples/square yard = 1495 pineapples.}\\

$\textbf{\dots}$

The final answer is $\boxed{8970}$.

\hdashrule{0.96\linewidth}{0.5pt}{3pt 2pt}\\
\textbf{Fifth Output (\hl{Incorrect})}\\

Here's how to solve the problem step-by-step:\\

1.  \textbf{Calculate the total number of pineapples:}\\
    *   John has 35 square yards of field.\\
    *   There are 43 pineapples per square yard.\\
    *   Total pineapples = \hl{35 square yards * 43 pineapples/square yard = 1495 pineapples.}\\

$\textbf{\dots}$

The final answer is $\boxed{8970}$.\\
Answer: \textbf{9030}
}
}
\end{minipage}
\caption{First five of eight independently generated outputs from Gemini 2.5 Flash Lite evaluated on Template 6. Error type: wrong multiplication.}
\label{fig:appendix_example_fails_gemini6}
\end{figure}

\end{document}